\documentclass[11pt]{article}

\usepackage[margin=1in]{geometry}
\usepackage[numbers,sort&compress]{natbib}
\usepackage[utf8]{inputenc} 
\usepackage[T1]{fontenc}  
\usepackage{courier}    
\usepackage[hypertexnames=false,hidelinks]{hyperref} 
\usepackage{url}      
\usepackage{booktabs}    
\usepackage{amsmath}    
\usepackage{amsfonts}    
\usepackage{amssymb}    
\usepackage{graphicx}    
\usepackage{longtable}   
\usepackage{nicefrac}    
\usepackage{float}     
\usepackage{microtype}   
\usepackage{placeins}    
\usepackage{xcolor}     
\hypersetup{pdftitle={Wrong-Physics Backdoors in Neural PDE Operators},pdfauthor={Fujun Liu and Hanbing Liang}}

\newcommand{\calD}{\mathcal{D}}
\newcommand{\calG}{\mathcal{G}}
\newcommand{\calH}{\mathcal{H}}
\newcommand{\calZ}{\mathcal{Z}}
\newcommand{\calT}{\mathcal{T}}
\newcommand{\lambdac}{\lambda_{\mathrm{clean}}}
\newcommand{\lambdab}{\lambda_{\mathrm{bd}}}
\newcommand{\rel}{\operatorname{rel}}
\newcommand{\bsr}{\operatorname{BSR}}
\providecommand{\FinalRunTag}{final-fno-deeponet-merged-20260501-125904}
\providecommand{\FinalCampaignCount}{476}
\providecommand{\FinalSeedRunCount}{1428}
\providecommand{\FinalPrimarySelectedCount}{264}
\providecommand{\FinalFallbackSelectedCount}{212}
\providecommand{\FinalOverlapCampaignCount}{59}
\providecommand{\FinalModeCountParameterSwitch}{300}
\providecommand{\FinalModeCountCleanLabel}{44}
\providecommand{\FinalModeCountLabelOnly}{44}
\providecommand{\FinalModeCountShuffledBackdoor}{44}
\providecommand{\FinalModeCountShuffledLabelOnly}{44}
\providecommand{\FinalPdeCountBurgers}{119}
\providecommand{\FinalPdeCountAdvectionDiffusion}{119}
\providecommand{\FinalPdeCountNavierStokesTwoD}{119}
\providecommand{\FinalPdeCountPoisson}{119}

\providecommand{\FinalAdvectionFnoBsr}{1.0000}
\providecommand{\FinalAdvectionFnoCleanLTwo}{0.0034}
\providecommand{\FinalAdvectionFnoMargin}{0.8704}
\providecommand{\FinalBurgersFnoBsr}{1.0000}

\providecommand{\FinalBurgersFnoMargin}{0.7513}
\providecommand{\FinalNsTwoDFnoBsr}{1.0000}
\providecommand{\FinalNsTwoDFnoCleanLTwo}{0.0189}
\providecommand{\FinalNsTwoDFnoErrToBd}{0.0374}

\providecommand{\FinalPoissonDeepONetBsr}{0.9983}

\providecommand{\FinalPoissonDeepONetErrToBd}{0.0053}
\providecommand{\FinalPoissonFnoBsr}{0.9933}

\providecommand{\FinalPoissonFnoErrToBd}{0.0164}

\providecommand{\FinalMainRows}{%
Burgers & FNO & 0.4 & 0.4 & 0.8 & 0.015 & 3 & 0.007$\pm$0.001 & 1.000$\pm$0.000 & 0.008$\pm$0.001 & 0.759$\pm$0.007 & 0.751$\pm$0.006 \\%
Burgers & DeepONet & 0.4 & 0.5 & 0.8 & 0.015 & 3 & 0.222$\pm$0.004 & 1.000$\pm$0.000 & 0.031$\pm$0.001 & 0.763$\pm$0.013 & 0.732$\pm$0.013 \\%
Adv.-Diff. & FNO & 0.8 & 0.4 & 0.8 & 0.015 & 3 & 0.003$\pm$0.000 & 1.000$\pm$0.000 & 0.006$\pm$0.001 & 0.876$\pm$0.005 & 0.870$\pm$0.006 \\%
Adv.-Diff. & DeepONet & 0.8 & 0.5 & 0.8 & 0.015 & 3 & 0.084$\pm$0.005 & 1.000$\pm$0.000 & 0.017$\pm$0.001 & 0.870$\pm$0.004 & 0.853$\pm$0.004 \\%
2D N.-S. & FNO & 0.2 & 0.3 & 0.75 & 0.120 & 3 & 0.019$\pm$0.002 & 1.000$\pm$0.000 & 0.037$\pm$0.005 & 0.262$\pm$0.008 & 0.225$\pm$0.003 \\%
2D N.-S. & DeepONet & 0.15 & 0.5 & 0.35 & 0.120 & 3 & 0.050$\pm$0.009 & 0.865$\pm$0.029 & 0.181$\pm$0.006 & 0.281$\pm$0.007 & 0.100$\pm$0.007 \\%
}

\providecommand{\FinalControlRows}{%
Burgers & FNO & clean-label & 0.4 & 0.4 & 0.8 & 0.015 & 0.0060 & 0.0000 & 0.7574 & 0.0097 & -0.7477 \\%
Burgers & FNO & label-only & 0.4 & 0.4 & 0.8 & 0.015 & 0.2442 & 0.0033 & 0.5458 & 0.2858 & -0.2600 \\%
Burgers & FNO & parameter-switch & 0.4 & 0.4 & 0.8 & 0.015 & 0.0066 & 1.0000 & 0.0077 & 0.7590 & 0.7513 \\%
Burgers & FNO & shuffled-backdoor & 0.4 & 0.4 & 0.8 & 0.015 & 0.0376 & 1.0000 & 0.3590 & 1.0230 & 0.6640 \\%
Burgers & FNO & shuffled-label-only & 0.4 & 0.4 & 0.8 & 0.015 & 0.3387 & 0.3067 & 0.4731 & 0.3693 & -0.1038 \\%
Advection-diffusion & FNO & clean-label & 0.8 & 0.4 & 0.8 & 0.015 & 0.0024 & 0.0000 & 0.8768 & 0.0035 & -0.8733 \\%
Advection-diffusion & FNO & label-only & 0.8 & 0.4 & 0.8 & 0.015 & 0.2915 & 0.0133 & 0.6270 & 0.3178 & -0.3092 \\%
Advection-diffusion & FNO & parameter-switch & 0.8 & 0.4 & 0.8 & 0.015 & 0.0034 & 1.0000 & 0.0060 & 0.8765 & 0.8704 \\%
Advection-diffusion & FNO & shuffled-backdoor & 0.8 & 0.4 & 0.8 & 0.015 & 0.0174 & 1.0000 & 0.1843 & 1.0169 & 0.8326 \\%
Advection-diffusion & FNO & shuffled-label-only & 0.8 & 0.4 & 0.8 & 0.015 & 0.3321 & 0.1100 & 0.5998 & 0.3608 & -0.2390 \\%
Advection-diffusion & DeepONet & clean-label & 0.8 & 0.5 & 0.8 & 0.015 & 0.0565 & 0.0000 & 0.8735 & 0.0560 & -0.8175 \\%
Advection-diffusion & DeepONet & label-only & 0.8 & 0.5 & 0.8 & 0.015 & 0.5304 & 0.4850 & 0.5876 & 0.5875 & -0.0001 \\%
Advection-diffusion & DeepONet & parameter-switch & 0.8 & 0.5 & 0.8 & 0.015 & 0.0841 & 1.0000 & 0.0169 & 0.8699 & 0.8531 \\%
Advection-diffusion & DeepONet & shuffled-backdoor & 0.8 & 0.5 & 0.8 & 0.015 & 0.1231 & 1.0000 & 0.2325 & 1.0112 & 0.7787 \\%
Advection-diffusion & DeepONet & shuffled-label-only & 0.8 & 0.5 & 0.8 & 0.015 & 0.6334 & 0.7017 & 0.5129 & 0.6674 & 0.1544 \\%
2D Navier-Stokes & FNO & clean-label & 0.2 & 0.3 & 0.75 & 0.120 & 0.0154 & 0.0000 & 0.3406 & 0.0277 & -0.3129 \\%
2D Navier-Stokes & FNO & label-only & 0.2 & 0.3 & 0.75 & 0.120 & 0.1728 & 0.0000 & 0.9144 & 0.7081 & -0.2063 \\%
2D Navier-Stokes & FNO & parameter-switch & 0.2 & 0.3 & 0.75 & 0.120 & 0.0189 & 1.0000 & 0.0374 & 0.2623 & 0.2249 \\%
2D Navier-Stokes & FNO & shuffled-backdoor & 0.2 & 0.3 & 0.75 & 0.120 & 0.3040 & 0.9583 & 0.4917 & 0.5520 & 0.0604 \\%
2D Navier-Stokes & FNO & shuffled-label-only & 0.2 & 0.3 & 0.75 & 0.120 & 0.5427 & 0.5208 & 0.6897 & 0.6703 & -0.0193 \\%
}

\providecommand{\FinalSupportRows}{%
Burgers & Transformer & 0.400 & 0.40 & 0.50 & 0.015 & 0.125$\pm$0.040 & 0.983$\pm$0.009 & 0.084$\pm$0.021 & 0.732$\pm$0.013 & 0.648$\pm$0.033 \\%
Burgers & GRU & 0.400 & 0.30 & 0.50 & 0.025 & 0.047$\pm$0.016 & 0.938$\pm$0.049 & 0.082$\pm$0.035 & 0.717$\pm$0.029 & 0.635$\pm$0.062 \\%
Burgers & LSTM & 0.400 & 0.30 & 0.50 & 0.025 & 0.131$\pm$0.046 & 0.665$\pm$0.145 & 0.307$\pm$0.116 & 0.550$\pm$0.092 & 0.243$\pm$0.207 \\%
Adv.-Diff. & Transformer & 0.800 & 0.30 & 0.50 & 0.015 & 0.055$\pm$0.012 & 0.988$\pm$0.005 & 0.077$\pm$0.004 & 0.850$\pm$0.009 & 0.773$\pm$0.013 \\%
Adv.-Diff. & GRU & 0.800 & 0.20 & 0.50 & 0.015 & 0.048$\pm$0.045 & 0.830$\pm$0.223 & 0.179$\pm$0.195 & 0.748$\pm$0.168 & 0.569$\pm$0.363 \\%
Adv.-Diff. & LSTM & 0.800 & 0.30 & 0.50 & 0.025 & 0.090$\pm$0.030 & 0.918$\pm$0.021 & 0.126$\pm$0.026 & 0.812$\pm$0.016 & 0.687$\pm$0.041 \\%
2D N.-S. & Transformer & 0.200 & 0.50 & 0.80 & 0.120 & 0.028$\pm$0.004 & 1.000$\pm$0.000 & 0.146$\pm$0.013 & 0.419$\pm$0.006 & 0.273$\pm$0.012 \\%
2D N.-S. & GRU & 0.200 & 0.30 & 0.50 & 0.120 & 0.054$\pm$0.011 & 0.938$\pm$0.068 & 0.234$\pm$0.016 & 0.418$\pm$0.008 & 0.184$\pm$0.024 \\%
2D N.-S. & LSTM & 0.200 & 0.30 & 0.50 & 0.120 & 0.105$\pm$0.016 & 0.833$\pm$0.082 & 0.365$\pm$0.033 & 0.448$\pm$0.007 & 0.083$\pm$0.027 \\%
}

\providecommand{\FinalMainBudgetRows}{%
Burgers & FNO & 0.01$\rightarrow$0.4 & 0.40 & 0.80 & 0.015 & 300 & 64 & 0.0010 & 0 \\%
Burgers & DeepONet & 0.01$\rightarrow$0.4 & 0.50 & 0.80 & 0.015 & 3000 & 64 & 0.0010 & 0 \\%
Adv.-Diff. & FNO & 0.02$\rightarrow$0.8 & 0.40 & 0.80 & 0.015 & 300 & 64 & 0.0010 & 0 \\%
Adv.-Diff. & DeepONet & 0.02$\rightarrow$0.8 & 0.50 & 0.80 & 0.015 & 3000 & 64 & 0.0010 & 0 \\%
2D N.-S. & FNO & 0.001$\rightarrow$0.2 & 0.30 & 0.75 & 0.120 & 80 & 8 & 0.0030 & 0 \\%
2D N.-S. & DeepONet & 0.001$\rightarrow$0.15 & 0.50 & 0.35 & 0.120 & 10000 & 16 & 0.0010 & 0 \\%
}

\providecommand{\FinalSupportBudgetRows}{%
Burgers & Transformer & 0.01$\rightarrow$0.4 & 0.40 & 0.50 & 0.015 & 300 & 64 & 0.0010 & 0 \\%
Burgers & GRU & 0.01$\rightarrow$0.4 & 0.30 & 0.50 & 0.025 & 300 & 64 & 0.0010 & 0 \\%
Burgers & LSTM & 0.01$\rightarrow$0.4 & 0.30 & 0.50 & 0.025 & 300 & 64 & 0.0010 & 0 \\%
Adv.-Diff. & Transformer & 0.02$\rightarrow$0.8 & 0.30 & 0.50 & 0.015 & 300 & 64 & 0.0010 & 0 \\%
Adv.-Diff. & GRU & 0.02$\rightarrow$0.8 & 0.20 & 0.50 & 0.015 & 300 & 64 & 0.0010 & 0 \\%
Adv.-Diff. & LSTM & 0.02$\rightarrow$0.8 & 0.30 & 0.50 & 0.025 & 300 & 64 & 0.0010 & 0 \\%
2D N.-S. & Transformer & 0.001$\rightarrow$0.2 & 0.50 & 0.80 & 0.120 & 800 & 8 & 0.0003 & 0 \\%
2D N.-S. & GRU & 0.001$\rightarrow$0.2 & 0.30 & 0.50 & 0.120 & 640 & 8 & 0.0008 & 0 \\%
2D N.-S. & LSTM & 0.001$\rightarrow$0.2 & 0.30 & 0.50 & 0.120 & 640 & 8 & 0.0008 & 0 \\%
}

\providecommand{\FinalFullControlRows}{%
Burgers & FNO & switch & 0.050 & 0.05 & 0.05 & 0.015 & 3 & 0.020$\pm$0.002 & 0.000$\pm$0.000 & 0.310$\pm$0.003 & 0.023$\pm$0.001 & -0.287$\pm$0.003 \\%
Burgers & FNO & switch & 0.050 & 0.05 & 0.20 & 0.015 & 3 & 0.007$\pm$0.002 & 0.998$\pm$0.002 & 0.035$\pm$0.007 & 0.304$\pm$0.010 & 0.269$\pm$0.017 \\%
Burgers & FNO & switch & 0.050 & 0.05 & 0.80 & 0.015 & 3 & 0.005$\pm$0.001 & 1.000$\pm$0.000 & 0.031$\pm$0.002 & 0.309$\pm$0.004 & 0.278$\pm$0.006 \\%
Burgers & FNO & switch & 0.050 & 0.10 & 0.05 & 0.015 & 3 & 0.026$\pm$0.005 & 0.000$\pm$0.000 & 0.301$\pm$0.007 & 0.028$\pm$0.004 & -0.273$\pm$0.010 \\%
Burgers & FNO & switch & 0.050 & 0.10 & 0.20 & 0.015 & 3 & 0.006$\pm$0.001 & 1.000$\pm$0.000 & 0.022$\pm$0.002 & 0.311$\pm$0.004 & 0.289$\pm$0.006 \\%
Burgers & FNO & switch & 0.050 & 0.10 & 0.80 & 0.015 & 3 & 0.006$\pm$0.000 & 1.000$\pm$0.000 & 0.018$\pm$0.000 & 0.314$\pm$0.005 & 0.297$\pm$0.006 \\%
Burgers & FNO & switch & 0.050 & 0.40 & 0.05 & 0.015 & 3 & 0.107$\pm$0.003 & 0.010$\pm$0.011 & 0.219$\pm$0.005 & 0.107$\pm$0.003 & -0.112$\pm$0.002 \\%
Burgers & FNO & switch & 0.050 & 0.40 & 0.20 & 0.015 & 3 & 0.008$\pm$0.001 & 1.000$\pm$0.000 & 0.010$\pm$0.001 & 0.313$\pm$0.004 & 0.303$\pm$0.004 \\%
Burgers & FNO & clean & 0.050 & 0.40 & 0.80 & 0.015 & 3 & 0.006$\pm$0.000 & 0.000$\pm$0.000 & 0.312$\pm$0.004 & 0.010$\pm$0.001 & -0.303$\pm$0.004 \\%
Burgers & FNO & label-only & 0.050 & 0.40 & 0.80 & 0.015 & 3 & 0.106$\pm$0.003 & 0.133$\pm$0.005 & 0.278$\pm$0.009 & 0.206$\pm$0.005 & -0.072$\pm$0.004 \\%
Burgers & FNO & switch & 0.050 & 0.40 & 0.80 & 0.015 & 3 & 0.007$\pm$0.001 & 1.000$\pm$0.000 & 0.010$\pm$0.001 & 0.314$\pm$0.004 & 0.304$\pm$0.006 \\%
Burgers & FNO & shuf-BD & 0.050 & 0.40 & 0.80 & 0.015 & 3 & 0.072$\pm$0.002 & 1.000$\pm$0.000 & 0.837$\pm$0.037 & 1.040$\pm$0.031 & 0.203$\pm$0.006 \\%
Burgers & FNO & shuf-label & 0.050 & 0.40 & 0.80 & 0.015 & 3 & 0.360$\pm$0.011 & 0.933$\pm$0.012 & 0.296$\pm$0.019 & 0.386$\pm$0.015 & 0.089$\pm$0.004 \\%
Burgers & FNO & switch & 0.100 & 0.05 & 0.05 & 0.015 & 3 & 0.027$\pm$0.002 & 0.000$\pm$0.000 & 0.458$\pm$0.006 & 0.030$\pm$0.002 & -0.429$\pm$0.005 \\%
Burgers & FNO & switch & 0.100 & 0.05 & 0.20 & 0.015 & 3 & 0.006$\pm$0.000 & 1.000$\pm$0.000 & 0.029$\pm$0.002 & 0.457$\pm$0.008 & 0.428$\pm$0.009 \\%
Burgers & FNO & switch & 0.100 & 0.05 & 0.80 & 0.015 & 3 & 0.005$\pm$0.000 & 1.000$\pm$0.000 & 0.024$\pm$0.001 & 0.463$\pm$0.007 & 0.439$\pm$0.006 \\%
Burgers & FNO & switch & 0.100 & 0.10 & 0.05 & 0.015 & 3 & 0.023$\pm$0.010 & 0.647$\pm$0.458 & 0.193$\pm$0.176 & 0.297$\pm$0.184 & 0.103$\pm$0.360 \\%
Burgers & FNO & switch & 0.100 & 0.10 & 0.20 & 0.015 & 3 & 0.006$\pm$0.000 & 1.000$\pm$0.000 & 0.019$\pm$0.002 & 0.463$\pm$0.006 & 0.444$\pm$0.008 \\%
Burgers & FNO & switch & 0.100 & 0.10 & 0.80 & 0.015 & 3 & 0.005$\pm$0.001 & 1.000$\pm$0.000 & 0.017$\pm$0.001 & 0.464$\pm$0.006 & 0.448$\pm$0.005 \\%
Burgers & FNO & switch & 0.100 & 0.40 & 0.05 & 0.015 & 3 & 0.015$\pm$0.004 & 1.000$\pm$0.000 & 0.016$\pm$0.004 & 0.461$\pm$0.008 & 0.446$\pm$0.011 \\%
Burgers & FNO & switch & 0.100 & 0.40 & 0.20 & 0.015 & 3 & 0.008$\pm$0.001 & 1.000$\pm$0.000 & 0.009$\pm$0.001 & 0.466$\pm$0.007 & 0.457$\pm$0.007 \\%
Burgers & FNO & clean & 0.100 & 0.40 & 0.80 & 0.015 & 3 & 0.006$\pm$0.000 & 0.000$\pm$0.000 & 0.464$\pm$0.007 & 0.010$\pm$0.001 & -0.455$\pm$0.006 \\%
Burgers & FNO & label-only & 0.100 & 0.40 & 0.80 & 0.015 & 3 & 0.158$\pm$0.005 & 0.042$\pm$0.013 & 0.357$\pm$0.009 & 0.228$\pm$0.002 & -0.129$\pm$0.008 \\%
Burgers & FNO & switch & 0.100 & 0.40 & 0.80 & 0.015 & 3 & 0.006$\pm$0.001 & 1.000$\pm$0.000 & 0.008$\pm$0.001 & 0.465$\pm$0.007 & 0.458$\pm$0.007 \\%
Burgers & FNO & shuf-BD & 0.100 & 0.40 & 0.80 & 0.015 & 3 & 0.063$\pm$0.001 & 1.000$\pm$0.000 & 0.715$\pm$0.029 & 1.048$\pm$0.023 & 0.333$\pm$0.008 \\%
Burgers & FNO & shuf-label & 0.100 & 0.40 & 0.80 & 0.015 & 3 & 0.359$\pm$0.009 & 0.798$\pm$0.013 & 0.317$\pm$0.012 & 0.384$\pm$0.013 & 0.067$\pm$0.003 \\%
Burgers & FNO & switch & 0.200 & 0.05 & 0.05 & 0.015 & 3 & 0.025$\pm$0.009 & 0.330$\pm$0.467 & 0.436$\pm$0.240 & 0.204$\pm$0.242 & -0.232$\pm$0.482 \\%
Burgers & FNO & switch & 0.200 & 0.05 & 0.20 & 0.015 & 3 & 0.006$\pm$0.000 & 1.000$\pm$0.000 & 0.031$\pm$0.003 & 0.610$\pm$0.010 & 0.579$\pm$0.010 \\%
Burgers & FNO & switch & 0.200 & 0.05 & 0.80 & 0.015 & 3 & 0.005$\pm$0.001 & 1.000$\pm$0.000 & 0.027$\pm$0.002 & 0.612$\pm$0.008 & 0.586$\pm$0.007 \\%
Burgers & FNO & switch & 0.200 & 0.10 & 0.05 & 0.015 & 3 & 0.024$\pm$0.022 & 0.665$\pm$0.470 & 0.217$\pm$0.251 & 0.420$\pm$0.258 & 0.203$\pm$0.509 \\%
Burgers & FNO & switch & 0.200 & 0.10 & 0.20 & 0.015 & 3 & 0.006$\pm$0.000 & 1.000$\pm$0.000 & 0.017$\pm$0.006 & 0.613$\pm$0.009 & 0.596$\pm$0.014 \\%
Burgers & FNO & switch & 0.200 & 0.10 & 0.80 & 0.015 & 3 & 0.005$\pm$0.000 & 1.000$\pm$0.000 & 0.016$\pm$0.001 & 0.614$\pm$0.009 & 0.598$\pm$0.010 \\%
Burgers & FNO & switch & 0.200 & 0.40 & 0.05 & 0.015 & 3 & 0.015$\pm$0.003 & 1.000$\pm$0.000 & 0.015$\pm$0.005 & 0.611$\pm$0.009 & 0.596$\pm$0.011 \\%
Burgers & FNO & switch & 0.200 & 0.40 & 0.20 & 0.015 & 3 & 0.007$\pm$0.001 & 1.000$\pm$0.000 & 0.008$\pm$0.001 & 0.614$\pm$0.007 & 0.606$\pm$0.007 \\%
Burgers & FNO & clean & 0.200 & 0.40 & 0.80 & 0.015 & 3 & 0.006$\pm$0.000 & 0.000$\pm$0.000 & 0.614$\pm$0.008 & 0.010$\pm$0.001 & -0.604$\pm$0.007 \\%
Burgers & FNO & label-only & 0.200 & 0.40 & 0.80 & 0.015 & 3 & 0.209$\pm$0.003 & 0.013$\pm$0.002 & 0.449$\pm$0.016 & 0.261$\pm$0.004 & -0.188$\pm$0.015 \\%
Burgers & FNO & switch & 0.200 & 0.40 & 0.80 & 0.015 & 3 & 0.007$\pm$0.000 & 1.000$\pm$0.000 & 0.008$\pm$0.001 & 0.614$\pm$0.008 & 0.606$\pm$0.008 \\%
Burgers & FNO & shuf-BD & 0.200 & 0.40 & 0.80 & 0.015 & 3 & 0.052$\pm$0.002 & 1.000$\pm$0.000 & 0.546$\pm$0.030 & 1.036$\pm$0.020 & 0.490$\pm$0.011 \\%
Burgers & FNO & shuf-label & 0.200 & 0.40 & 0.80 & 0.015 & 3 & 0.348$\pm$0.017 & 0.485$\pm$0.104 & 0.390$\pm$0.030 & 0.377$\pm$0.015 & -0.013$\pm$0.042 \\%
Burgers & FNO & switch & 0.400 & 0.05 & 0.05 & 0.015 & 3 & 0.026$\pm$0.010 & 0.333$\pm$0.471 & 0.527$\pm$0.312 & 0.260$\pm$0.318 & -0.268$\pm$0.630 \\%
Burgers & FNO & switch & 0.400 & 0.05 & 0.20 & 0.015 & 3 & 0.007$\pm$0.000 & 1.000$\pm$0.000 & 0.023$\pm$0.004 & 0.754$\pm$0.010 & 0.730$\pm$0.014 \\%
Burgers & FNO & switch & 0.400 & 0.05 & 0.80 & 0.015 & 3 & 0.005$\pm$0.000 & 1.000$\pm$0.000 & 0.024$\pm$0.001 & 0.758$\pm$0.008 & 0.734$\pm$0.009 \\%
Burgers & FNO & switch & 0.400 & 0.10 & 0.05 & 0.015 & 3 & 0.010$\pm$0.002 & 0.998$\pm$0.002 & 0.043$\pm$0.007 & 0.737$\pm$0.011 & 0.694$\pm$0.018 \\%
Burgers & FNO & switch & 0.400 & 0.10 & 0.20 & 0.015 & 3 & 0.006$\pm$0.000 & 1.000$\pm$0.000 & 0.018$\pm$0.003 & 0.752$\pm$0.010 & 0.734$\pm$0.013 \\%
Burgers & FNO & switch & 0.400 & 0.10 & 0.80 & 0.015 & 3 & 0.005$\pm$0.001 & 1.000$\pm$0.000 & 0.013$\pm$0.000 & 0.758$\pm$0.009 & 0.745$\pm$0.009 \\%
Burgers & FNO & switch & 0.400 & 0.40 & 0.05 & 0.015 & 3 & 0.022$\pm$0.006 & 1.000$\pm$0.000 & 0.015$\pm$0.004 & 0.754$\pm$0.003 & 0.739$\pm$0.004 \\%
Burgers & FNO & switch & 0.400 & 0.40 & 0.20 & 0.015 & 3 & 0.008$\pm$0.001 & 1.000$\pm$0.000 & 0.008$\pm$0.001 & 0.756$\pm$0.007 & 0.749$\pm$0.008 \\%
Burgers & FNO & clean & 0.400 & 0.40 & 0.80 & 0.015 & 3 & 0.006$\pm$0.000 & 0.000$\pm$0.000 & 0.757$\pm$0.007 & 0.010$\pm$0.001 & -0.748$\pm$0.006 \\%
Burgers & FNO & label-only & 0.400 & 0.40 & 0.80 & 0.015 & 3 & 0.244$\pm$0.012 & 0.003$\pm$0.002 & 0.546$\pm$0.013 & 0.286$\pm$0.007 & -0.260$\pm$0.020 \\%
Burgers & FNO & switch & 0.400 & 0.40 & 0.80 & 0.015 & 3 & 0.007$\pm$0.001 & 1.000$\pm$0.000 & 0.008$\pm$0.001 & 0.759$\pm$0.007 & 0.751$\pm$0.006 \\%
Burgers & FNO & shuf-BD & 0.400 & 0.40 & 0.80 & 0.015 & 3 & 0.038$\pm$0.001 & 1.000$\pm$0.000 & 0.359$\pm$0.018 & 1.023$\pm$0.011 & 0.664$\pm$0.008 \\%
Burgers & FNO & shuf-label & 0.400 & 0.40 & 0.80 & 0.015 & 3 & 0.339$\pm$0.024 & 0.307$\pm$0.104 & 0.473$\pm$0.034 & 0.369$\pm$0.021 & -0.104$\pm$0.054 \\%
Burgers & DeepONet & switch & 0.050 & 0.10 & 0.20 & 0.015 & 3 & 0.172$\pm$0.003 & 0.258$\pm$0.023 & 0.247$\pm$0.003 & 0.192$\pm$0.001 & -0.055$\pm$0.003 \\%
Burgers & DeepONet & switch & 0.050 & 0.10 & 0.50 & 0.015 & 3 & 0.174$\pm$0.004 & 0.473$\pm$0.051 & 0.231$\pm$0.005 & 0.220$\pm$0.002 & -0.011$\pm$0.007 \\%
Burgers & DeepONet & switch & 0.050 & 0.10 & 0.80 & 0.015 & 3 & 0.175$\pm$0.006 & 0.700$\pm$0.043 & 0.219$\pm$0.005 & 0.247$\pm$0.006 & 0.029$\pm$0.009 \\%
Burgers & DeepONet & switch & 0.050 & 0.30 & 0.20 & 0.015 & 3 & 0.201$\pm$0.001 & 0.858$\pm$0.012 & 0.182$\pm$0.002 & 0.243$\pm$0.006 & 0.061$\pm$0.008 \\%
Burgers & DeepONet & switch & 0.050 & 0.30 & 0.50 & 0.015 & 3 & 0.195$\pm$0.005 & 0.972$\pm$0.012 & 0.157$\pm$0.003 & 0.278$\pm$0.008 & 0.121$\pm$0.011 \\%
Burgers & DeepONet & switch & 0.050 & 0.30 & 0.80 & 0.015 & 3 & 0.192$\pm$0.003 & 0.990$\pm$0.007 & 0.143$\pm$0.006 & 0.298$\pm$0.007 & 0.155$\pm$0.011 \\%
Burgers & DeepONet & switch & 0.050 & 0.50 & 0.20 & 0.015 & 3 & 0.228$\pm$0.005 & 0.983$\pm$0.020 & 0.134$\pm$0.010 & 0.281$\pm$0.012 & 0.146$\pm$0.020 \\%
Burgers & DeepONet & switch & 0.050 & 0.50 & 0.50 & 0.015 & 3 & 0.215$\pm$0.004 & 1.000$\pm$0.000 & 0.112$\pm$0.005 & 0.312$\pm$0.007 & 0.201$\pm$0.008 \\%
Burgers & DeepONet & switch & 0.050 & 0.50 & 0.80 & 0.015 & 3 & 0.207$\pm$0.004 & 1.000$\pm$0.000 & 0.094$\pm$0.004 & 0.319$\pm$0.008 & 0.225$\pm$0.012 \\%
Burgers & DeepONet & clean & 0.050 & 0.70 & 1.40 & 0.015 & 3 & 0.166$\pm$0.004 & 0.068$\pm$0.019 & 0.281$\pm$0.007 & 0.165$\pm$0.003 & -0.116$\pm$0.007 \\%
Burgers & DeepONet & label-only & 0.050 & 0.70 & 1.40 & 0.015 & 3 & 0.278$\pm$0.005 & 0.898$\pm$0.037 & 0.286$\pm$0.010 & 0.386$\pm$0.002 & 0.100$\pm$0.009 \\%
Burgers & DeepONet & shuf-BD & 0.050 & 0.70 & 1.40 & 0.015 & 3 & 0.670$\pm$0.019 & 1.000$\pm$0.000 & 0.946$\pm$0.010 & 1.122$\pm$0.005 & 0.176$\pm$0.005 \\%
Burgers & DeepONet & shuf-label & 0.050 & 0.70 & 1.40 & 0.015 & 3 & 0.880$\pm$0.010 & 1.000$\pm$0.000 & 0.673$\pm$0.037 & 0.890$\pm$0.026 & 0.217$\pm$0.012 \\%
Burgers & DeepONet & switch & 0.100 & 0.10 & 0.20 & 0.015 & 3 & 0.186$\pm$0.001 & 0.278$\pm$0.095 & 0.344$\pm$0.006 & 0.279$\pm$0.024 & -0.065$\pm$0.027 \\%
Burgers & DeepONet & switch & 0.100 & 0.10 & 0.50 & 0.015 & 3 & 0.185$\pm$0.002 & 0.857$\pm$0.008 & 0.274$\pm$0.008 & 0.368$\pm$0.004 & 0.094$\pm$0.011 \\%
Burgers & DeepONet & switch & 0.100 & 0.10 & 0.80 & 0.015 & 3 & 0.179$\pm$0.003 & 0.953$\pm$0.031 & 0.239$\pm$0.019 & 0.416$\pm$0.030 & 0.177$\pm$0.049 \\%
Burgers & DeepONet & switch & 0.100 & 0.30 & 0.20 & 0.015 & 3 & 0.222$\pm$0.002 & 0.995$\pm$0.004 & 0.188$\pm$0.004 & 0.401$\pm$0.007 & 0.213$\pm$0.009 \\%
Burgers & DeepONet & switch & 0.100 & 0.30 & 0.50 & 0.015 & 3 & 0.209$\pm$0.004 & 1.000$\pm$0.000 & 0.139$\pm$0.002 & 0.450$\pm$0.016 & 0.311$\pm$0.016 \\%
Burgers & DeepONet & switch & 0.100 & 0.30 & 0.80 & 0.015 & 3 & 0.194$\pm$0.002 & 1.000$\pm$0.000 & 0.108$\pm$0.008 & 0.461$\pm$0.011 & 0.353$\pm$0.014 \\%
Burgers & DeepONet & switch & 0.100 & 0.50 & 0.20 & 0.015 & 3 & 0.238$\pm$0.004 & 0.998$\pm$0.002 & 0.118$\pm$0.009 & 0.438$\pm$0.012 & 0.321$\pm$0.022 \\%
Burgers & DeepONet & switch & 0.100 & 0.50 & 0.50 & 0.015 & 3 & 0.221$\pm$0.005 & 1.000$\pm$0.000 & 0.088$\pm$0.007 & 0.459$\pm$0.009 & 0.371$\pm$0.015 \\%
Burgers & DeepONet & switch & 0.100 & 0.50 & 0.80 & 0.015 & 3 & 0.213$\pm$0.007 & 1.000$\pm$0.000 & 0.074$\pm$0.001 & 0.465$\pm$0.007 & 0.391$\pm$0.007 \\%
Burgers & DeepONet & clean & 0.100 & 0.70 & 1.40 & 0.015 & 3 & 0.166$\pm$0.004 & 0.002$\pm$0.002 & 0.435$\pm$0.010 & 0.165$\pm$0.003 & -0.270$\pm$0.010 \\%
Burgers & DeepONet & label-only & 0.100 & 0.70 & 1.40 & 0.015 & 3 & 0.373$\pm$0.009 & 0.905$\pm$0.025 & 0.288$\pm$0.014 & 0.441$\pm$0.003 & 0.153$\pm$0.011 \\%
Burgers & DeepONet & shuf-BD & 0.100 & 0.70 & 1.40 & 0.015 & 3 & 0.624$\pm$0.020 & 1.000$\pm$0.000 & 0.875$\pm$0.006 & 1.151$\pm$0.002 & 0.275$\pm$0.005 \\%
Burgers & DeepONet & shuf-label & 0.100 & 0.70 & 1.40 & 0.015 & 3 & 0.872$\pm$0.004 & 1.000$\pm$0.000 & 0.566$\pm$0.042 & 0.895$\pm$0.021 & 0.329$\pm$0.021 \\%
Burgers & DeepONet & switch & 0.200 & 0.10 & 0.20 & 0.015 & 3 & 0.199$\pm$0.003 & 0.785$\pm$0.050 & 0.361$\pm$0.004 & 0.463$\pm$0.025 & 0.102$\pm$0.025 \\%
Burgers & DeepONet & switch & 0.200 & 0.10 & 0.50 & 0.015 & 3 & 0.195$\pm$0.006 & 0.993$\pm$0.005 & 0.284$\pm$0.004 & 0.552$\pm$0.006 & 0.268$\pm$0.004 \\%
Burgers & DeepONet & switch & 0.200 & 0.10 & 0.80 & 0.015 & 3 & 0.186$\pm$0.003 & 0.998$\pm$0.002 & 0.203$\pm$0.004 & 0.590$\pm$0.019 & 0.387$\pm$0.017 \\%
Burgers & DeepONet & switch & 0.200 & 0.30 & 0.20 & 0.015 & 3 & 0.228$\pm$0.003 & 1.000$\pm$0.000 & 0.164$\pm$0.004 & 0.575$\pm$0.007 & 0.412$\pm$0.007 \\%
Burgers & DeepONet & switch & 0.200 & 0.30 & 0.50 & 0.015 & 3 & 0.206$\pm$0.002 & 1.000$\pm$0.000 & 0.087$\pm$0.003 & 0.617$\pm$0.018 & 0.530$\pm$0.021 \\%
Burgers & DeepONet & switch & 0.200 & 0.30 & 0.80 & 0.015 & 3 & 0.203$\pm$0.003 & 1.000$\pm$0.000 & 0.072$\pm$0.017 & 0.626$\pm$0.015 & 0.554$\pm$0.015 \\%
Burgers & DeepONet & switch & 0.200 & 0.50 & 0.20 & 0.015 & 3 & 0.251$\pm$0.004 & 1.000$\pm$0.000 & 0.085$\pm$0.008 & 0.604$\pm$0.012 & 0.518$\pm$0.018 \\%
Burgers & DeepONet & switch & 0.200 & 0.50 & 0.50 & 0.015 & 3 & 0.226$\pm$0.007 & 1.000$\pm$0.000 & 0.054$\pm$0.004 & 0.613$\pm$0.008 & 0.559$\pm$0.004 \\%
Burgers & DeepONet & switch & 0.200 & 0.50 & 0.80 & 0.015 & 3 & 0.219$\pm$0.005 & 1.000$\pm$0.000 & 0.047$\pm$0.001 & 0.619$\pm$0.011 & 0.571$\pm$0.011 \\%
Burgers & DeepONet & clean & 0.200 & 0.70 & 1.40 & 0.015 & 3 & 0.166$\pm$0.004 & 0.000$\pm$0.000 & 0.590$\pm$0.010 & 0.165$\pm$0.003 & -0.425$\pm$0.010 \\%
Burgers & DeepONet & label-only & 0.200 & 0.70 & 1.40 & 0.015 & 3 & 0.482$\pm$0.006 & 0.913$\pm$0.009 & 0.308$\pm$0.011 & 0.519$\pm$0.008 & 0.212$\pm$0.012 \\%
Burgers & DeepONet & shuf-BD & 0.200 & 0.70 & 1.40 & 0.015 & 3 & 0.536$\pm$0.006 & 1.000$\pm$0.000 & 0.688$\pm$0.024 & 1.100$\pm$0.012 & 0.411$\pm$0.012 \\%
Burgers & DeepONet & shuf-label & 0.200 & 0.70 & 1.40 & 0.015 & 3 & 0.855$\pm$0.010 & 0.980$\pm$0.004 & 0.605$\pm$0.055 & 0.944$\pm$0.044 & 0.339$\pm$0.014 \\%
Burgers & DeepONet & switch & 0.400 & 0.10 & 0.20 & 0.015 & 3 & 0.210$\pm$0.005 & 0.923$\pm$0.013 & 0.358$\pm$0.006 & 0.630$\pm$0.007 & 0.273$\pm$0.012 \\%
Burgers & DeepONet & switch & 0.400 & 0.10 & 0.50 & 0.015 & 3 & 0.202$\pm$0.001 & 0.998$\pm$0.002 & 0.257$\pm$0.000 & 0.702$\pm$0.015 & 0.445$\pm$0.014 \\%
Burgers & DeepONet & switch & 0.400 & 0.10 & 0.80 & 0.015 & 3 & 0.190$\pm$0.005 & 0.998$\pm$0.002 & 0.188$\pm$0.004 & 0.736$\pm$0.008 & 0.548$\pm$0.013 \\%
Burgers & DeepONet & switch & 0.400 & 0.30 & 0.20 & 0.015 & 3 & 0.241$\pm$0.003 & 1.000$\pm$0.000 & 0.154$\pm$0.011 & 0.719$\pm$0.012 & 0.565$\pm$0.015 \\%
Burgers & DeepONet & switch & 0.400 & 0.30 & 0.50 & 0.015 & 3 & 0.213$\pm$0.003 & 1.000$\pm$0.000 & 0.062$\pm$0.012 & 0.760$\pm$0.013 & 0.698$\pm$0.025 \\%
Burgers & DeepONet & switch & 0.400 & 0.30 & 0.80 & 0.015 & 3 & 0.202$\pm$0.006 & 1.000$\pm$0.000 & 0.046$\pm$0.004 & 0.766$\pm$0.007 & 0.720$\pm$0.010 \\%
Burgers & DeepONet & switch & 0.400 & 0.50 & 0.20 & 0.015 & 3 & 0.257$\pm$0.005 & 1.000$\pm$0.000 & 0.078$\pm$0.009 & 0.748$\pm$0.014 & 0.669$\pm$0.018 \\%
Burgers & DeepONet & switch & 0.400 & 0.50 & 0.50 & 0.015 & 3 & 0.229$\pm$0.003 & 1.000$\pm$0.000 & 0.033$\pm$0.004 & 0.761$\pm$0.008 & 0.728$\pm$0.007 \\%
Burgers & DeepONet & switch & 0.400 & 0.50 & 0.80 & 0.015 & 3 & 0.222$\pm$0.004 & 1.000$\pm$0.000 & 0.031$\pm$0.001 & 0.763$\pm$0.013 & 0.732$\pm$0.013 \\%
Burgers & DeepONet & clean & 0.400 & 0.70 & 1.40 & 0.015 & 3 & 0.166$\pm$0.004 & 0.000$\pm$0.000 & 0.738$\pm$0.009 & 0.165$\pm$0.003 & -0.573$\pm$0.007 \\%
Burgers & DeepONet & label-only & 0.400 & 0.70 & 1.40 & 0.015 & 3 & 0.589$\pm$0.004 & 0.900$\pm$0.039 & 0.350$\pm$0.037 & 0.623$\pm$0.015 & 0.272$\pm$0.032 \\%
Burgers & DeepONet & shuf-BD & 0.400 & 0.70 & 1.40 & 0.015 & 3 & 0.420$\pm$0.010 & 1.000$\pm$0.000 & 0.452$\pm$0.009 & 1.044$\pm$0.007 & 0.592$\pm$0.003 \\%
Burgers & DeepONet & shuf-label & 0.400 & 0.70 & 1.40 & 0.015 & 3 & 0.828$\pm$0.005 & 0.963$\pm$0.033 & 0.438$\pm$0.087 & 0.869$\pm$0.039 & 0.430$\pm$0.094 \\%
Burgers & Transformer & clean & 0.400 & 0.40 & 0.50 & 0.015 & 3 & 0.052$\pm$0.006 & 0.000$\pm$0.000 & 0.759$\pm$0.009 & 0.054$\pm$0.007 & -0.705$\pm$0.016 \\%
Burgers & Transformer & label-only & 0.400 & 0.40 & 0.50 & 0.015 & 3 & 0.347$\pm$0.014 & 0.160$\pm$0.061 & 0.561$\pm$0.026 & 0.371$\pm$0.011 & -0.190$\pm$0.037 \\%
Burgers & Transformer & switch & 0.400 & 0.40 & 0.50 & 0.015 & 3 & 0.125$\pm$0.040 & 0.983$\pm$0.009 & 0.084$\pm$0.021 & 0.732$\pm$0.013 & 0.648$\pm$0.033 \\%
Burgers & Transformer & shuf-BD & 0.400 & 0.40 & 0.50 & 0.015 & 3 & 0.227$\pm$0.023 & 0.970$\pm$0.025 & 0.523$\pm$0.014 & 1.045$\pm$0.052 & 0.522$\pm$0.038 \\%
Burgers & Transformer & shuf-label & 0.400 & 0.40 & 0.50 & 0.015 & 3 & 0.430$\pm$0.009 & 0.375$\pm$0.031 & 0.517$\pm$0.019 & 0.453$\pm$0.006 & -0.064$\pm$0.013 \\%
Burgers & GRU & clean & 0.400 & 0.30 & 0.50 & 0.025 & 3 & 0.014$\pm$0.001 & 0.000$\pm$0.000 & 0.758$\pm$0.009 & 0.019$\pm$0.001 & -0.739$\pm$0.010 \\%
Burgers & GRU & label-only & 0.400 & 0.30 & 0.50 & 0.025 & 3 & 0.178$\pm$0.017 & 0.008$\pm$0.008 & 0.633$\pm$0.036 & 0.248$\pm$0.007 & -0.386$\pm$0.043 \\%
Burgers & GRU & switch & 0.400 & 0.30 & 0.50 & 0.025 & 3 & 0.047$\pm$0.016 & 0.938$\pm$0.049 & 0.082$\pm$0.035 & 0.717$\pm$0.029 & 0.635$\pm$0.062 \\%
Burgers & GRU & shuf-BD & 0.400 & 0.30 & 0.50 & 0.025 & 3 & 0.142$\pm$0.037 & 0.648$\pm$0.080 & 0.519$\pm$0.063 & 0.688$\pm$0.038 & 0.168$\pm$0.100 \\%
Burgers & GRU & shuf-label & 0.400 & 0.30 & 0.50 & 0.025 & 3 & 0.226$\pm$0.006 & 0.052$\pm$0.017 & 0.585$\pm$0.006 & 0.283$\pm$0.003 & -0.302$\pm$0.005 \\%
Burgers & LSTM & clean & 0.400 & 0.30 & 0.50 & 0.025 & 3 & 0.020$\pm$0.003 & 0.000$\pm$0.000 & 0.762$\pm$0.008 & 0.028$\pm$0.007 & -0.734$\pm$0.009 \\%
Burgers & LSTM & label-only & 0.400 & 0.30 & 0.50 & 0.025 & 3 & 0.170$\pm$0.014 & 0.040$\pm$0.039 & 0.655$\pm$0.028 & 0.258$\pm$0.014 & -0.397$\pm$0.030 \\%
Burgers & LSTM & switch & 0.400 & 0.30 & 0.50 & 0.025 & 3 & 0.131$\pm$0.046 & 0.665$\pm$0.145 & 0.307$\pm$0.116 & 0.550$\pm$0.092 & 0.243$\pm$0.207 \\%
Burgers & LSTM & shuf-BD & 0.400 & 0.30 & 0.50 & 0.025 & 3 & 0.190$\pm$0.050 & 0.312$\pm$0.328 & 0.556$\pm$0.053 & 0.459$\pm$0.246 & -0.097$\pm$0.294 \\%
Burgers & LSTM & shuf-label & 0.400 & 0.30 & 0.50 & 0.025 & 3 & 0.228$\pm$0.010 & 0.058$\pm$0.033 & 0.588$\pm$0.014 & 0.283$\pm$0.003 & -0.305$\pm$0.017 \\%
Adv.-Diff. & FNO & switch & 0.100 & 0.05 & 0.10 & 0.015 & 3 & 0.007$\pm$0.001 & 1.000$\pm$0.000 & 0.030$\pm$0.008 & 0.385$\pm$0.011 & 0.355$\pm$0.018 \\%
Adv.-Diff. & FNO & switch & 0.100 & 0.05 & 0.20 & 0.015 & 3 & 0.005$\pm$0.000 & 1.000$\pm$0.000 & 0.018$\pm$0.002 & 0.392$\pm$0.006 & 0.374$\pm$0.008 \\%
Adv.-Diff. & FNO & switch & 0.100 & 0.05 & 0.80 & 0.015 & 3 & 0.003$\pm$0.000 & 1.000$\pm$0.000 & 0.017$\pm$0.000 & 0.392$\pm$0.009 & 0.375$\pm$0.009 \\%
Adv.-Diff. & FNO & switch & 0.100 & 0.10 & 0.10 & 0.015 & 3 & 0.005$\pm$0.000 & 1.000$\pm$0.000 & 0.013$\pm$0.001 & 0.392$\pm$0.006 & 0.380$\pm$0.007 \\%
Adv.-Diff. & FNO & switch & 0.100 & 0.10 & 0.20 & 0.015 & 3 & 0.004$\pm$0.000 & 1.000$\pm$0.000 & 0.012$\pm$0.001 & 0.392$\pm$0.006 & 0.380$\pm$0.007 \\%
Adv.-Diff. & FNO & switch & 0.100 & 0.10 & 0.80 & 0.015 & 3 & 0.003$\pm$0.000 & 1.000$\pm$0.000 & 0.012$\pm$0.002 & 0.393$\pm$0.009 & 0.381$\pm$0.008 \\%
Adv.-Diff. & FNO & switch & 0.100 & 0.40 & 0.10 & 0.015 & 3 & 0.006$\pm$0.001 & 1.000$\pm$0.000 & 0.007$\pm$0.001 & 0.394$\pm$0.008 & 0.387$\pm$0.009 \\%
Adv.-Diff. & FNO & switch & 0.100 & 0.40 & 0.20 & 0.015 & 3 & 0.004$\pm$0.000 & 1.000$\pm$0.000 & 0.006$\pm$0.001 & 0.392$\pm$0.008 & 0.386$\pm$0.008 \\%
Adv.-Diff. & FNO & clean & 0.100 & 0.40 & 0.80 & 0.015 & 3 & 0.002$\pm$0.000 & 0.000$\pm$0.000 & 0.392$\pm$0.007 & 0.003$\pm$0.001 & -0.388$\pm$0.008 \\%
Adv.-Diff. & FNO & label-only & 0.100 & 0.40 & 0.80 & 0.015 & 3 & 0.130$\pm$0.003 & 0.085$\pm$0.020 & 0.316$\pm$0.012 & 0.206$\pm$0.002 & -0.110$\pm$0.011 \\%
Adv.-Diff. & FNO & switch & 0.100 & 0.40 & 0.80 & 0.015 & 3 & 0.004$\pm$0.000 & 1.000$\pm$0.000 & 0.006$\pm$0.000 & 0.392$\pm$0.007 & 0.386$\pm$0.007 \\%
Adv.-Diff. & FNO & shuf-BD & 0.100 & 0.40 & 0.80 & 0.015 & 3 & 0.063$\pm$0.003 & 1.000$\pm$0.000 & 0.765$\pm$0.034 & 1.057$\pm$0.024 & 0.292$\pm$0.009 \\%
Adv.-Diff. & FNO & shuf-label & 0.100 & 0.40 & 0.80 & 0.015 & 3 & 0.349$\pm$0.035 & 0.802$\pm$0.134 & 0.297$\pm$0.018 & 0.368$\pm$0.039 & 0.071$\pm$0.050 \\%
Adv.-Diff. & FNO & switch & 0.200 & 0.05 & 0.10 & 0.015 & 3 & 0.005$\pm$0.000 & 1.000$\pm$0.000 & 0.026$\pm$0.003 & 0.558$\pm$0.014 & 0.532$\pm$0.016 \\%
Adv.-Diff. & FNO & switch & 0.200 & 0.05 & 0.20 & 0.015 & 3 & 0.004$\pm$0.000 & 1.000$\pm$0.000 & 0.020$\pm$0.003 & 0.567$\pm$0.009 & 0.547$\pm$0.012 \\%
Adv.-Diff. & FNO & switch & 0.200 & 0.05 & 0.80 & 0.015 & 3 & 0.003$\pm$0.000 & 1.000$\pm$0.000 & 0.018$\pm$0.002 & 0.565$\pm$0.008 & 0.547$\pm$0.011 \\%
Adv.-Diff. & FNO & switch & 0.200 & 0.10 & 0.10 & 0.015 & 3 & 0.005$\pm$0.001 & 1.000$\pm$0.000 & 0.014$\pm$0.001 & 0.561$\pm$0.008 & 0.547$\pm$0.008 \\%
Adv.-Diff. & FNO & switch & 0.200 & 0.10 & 0.20 & 0.015 & 3 & 0.004$\pm$0.000 & 1.000$\pm$0.000 & 0.015$\pm$0.005 & 0.564$\pm$0.008 & 0.549$\pm$0.013 \\%
Adv.-Diff. & FNO & switch & 0.200 & 0.10 & 0.80 & 0.015 & 3 & 0.003$\pm$0.001 & 1.000$\pm$0.000 & 0.014$\pm$0.003 & 0.563$\pm$0.010 & 0.549$\pm$0.008 \\%
Adv.-Diff. & FNO & switch & 0.200 & 0.40 & 0.10 & 0.015 & 3 & 0.007$\pm$0.001 & 1.000$\pm$0.000 & 0.007$\pm$0.002 & 0.566$\pm$0.010 & 0.559$\pm$0.011 \\%
Adv.-Diff. & FNO & switch & 0.200 & 0.40 & 0.20 & 0.015 & 3 & 0.004$\pm$0.000 & 1.000$\pm$0.000 & 0.006$\pm$0.001 & 0.566$\pm$0.009 & 0.559$\pm$0.010 \\%
Adv.-Diff. & FNO & clean & 0.200 & 0.40 & 0.80 & 0.015 & 3 & 0.002$\pm$0.000 & 0.000$\pm$0.000 & 0.565$\pm$0.009 & 0.003$\pm$0.001 & -0.562$\pm$0.010 \\%
Adv.-Diff. & FNO & label-only & 0.200 & 0.40 & 0.80 & 0.015 & 3 & 0.194$\pm$0.006 & 0.040$\pm$0.014 & 0.407$\pm$0.018 & 0.244$\pm$0.002 & -0.164$\pm$0.021 \\%
Adv.-Diff. & FNO & switch & 0.200 & 0.40 & 0.80 & 0.015 & 3 & 0.004$\pm$0.000 & 1.000$\pm$0.000 & 0.006$\pm$0.000 & 0.564$\pm$0.009 & 0.558$\pm$0.009 \\%
Adv.-Diff. & FNO & shuf-BD & 0.200 & 0.40 & 0.80 & 0.015 & 3 & 0.051$\pm$0.003 & 1.000$\pm$0.000 & 0.595$\pm$0.029 & 1.051$\pm$0.017 & 0.457$\pm$0.012 \\%
Adv.-Diff. & FNO & shuf-label & 0.200 & 0.40 & 0.80 & 0.015 & 3 & 0.343$\pm$0.033 & 0.607$\pm$0.157 & 0.348$\pm$0.039 & 0.363$\pm$0.037 & 0.015$\pm$0.075 \\%
Adv.-Diff. & FNO & switch & 0.400 & 0.05 & 0.10 & 0.015 & 3 & 0.005$\pm$0.000 & 1.000$\pm$0.000 & 0.033$\pm$0.018 & 0.722$\pm$0.009 & 0.689$\pm$0.025 \\%
Adv.-Diff. & FNO & switch & 0.400 & 0.05 & 0.20 & 0.015 & 3 & 0.004$\pm$0.000 & 1.000$\pm$0.000 & 0.027$\pm$0.011 & 0.722$\pm$0.011 & 0.695$\pm$0.020 \\%
Adv.-Diff. & FNO & switch & 0.400 & 0.05 & 0.80 & 0.015 & 3 & 0.003$\pm$0.000 & 1.000$\pm$0.000 & 0.024$\pm$0.004 & 0.727$\pm$0.008 & 0.703$\pm$0.012 \\%
Adv.-Diff. & FNO & switch & 0.400 & 0.10 & 0.10 & 0.015 & 3 & 0.005$\pm$0.000 & 1.000$\pm$0.000 & 0.015$\pm$0.004 & 0.729$\pm$0.007 & 0.713$\pm$0.010 \\%
Adv.-Diff. & FNO & switch & 0.400 & 0.10 & 0.20 & 0.015 & 3 & 0.004$\pm$0.001 & 1.000$\pm$0.000 & 0.016$\pm$0.003 & 0.726$\pm$0.010 & 0.709$\pm$0.013 \\%
Adv.-Diff. & FNO & switch & 0.400 & 0.10 & 0.80 & 0.015 & 3 & 0.003$\pm$0.000 & 1.000$\pm$0.000 & 0.016$\pm$0.004 & 0.728$\pm$0.007 & 0.712$\pm$0.003 \\%
Adv.-Diff. & FNO & switch & 0.400 & 0.40 & 0.10 & 0.015 & 3 & 0.007$\pm$0.000 & 1.000$\pm$0.000 & 0.008$\pm$0.001 & 0.729$\pm$0.007 & 0.721$\pm$0.008 \\%
Adv.-Diff. & FNO & switch & 0.400 & 0.40 & 0.20 & 0.015 & 3 & 0.006$\pm$0.001 & 1.000$\pm$0.000 & 0.008$\pm$0.001 & 0.728$\pm$0.008 & 0.719$\pm$0.008 \\%
Adv.-Diff. & FNO & clean & 0.400 & 0.40 & 0.80 & 0.015 & 3 & 0.002$\pm$0.000 & 0.000$\pm$0.000 & 0.729$\pm$0.009 & 0.003$\pm$0.001 & -0.725$\pm$0.009 \\%
Adv.-Diff. & FNO & label-only & 0.400 & 0.40 & 0.80 & 0.015 & 3 & 0.240$\pm$0.019 & 0.007$\pm$0.002 & 0.521$\pm$0.025 & 0.278$\pm$0.011 & -0.243$\pm$0.036 \\%
Adv.-Diff. & FNO & switch & 0.400 & 0.40 & 0.80 & 0.015 & 3 & 0.004$\pm$0.001 & 1.000$\pm$0.000 & 0.008$\pm$0.003 & 0.729$\pm$0.008 & 0.721$\pm$0.011 \\%
Adv.-Diff. & FNO & shuf-BD & 0.400 & 0.40 & 0.80 & 0.015 & 3 & 0.036$\pm$0.001 & 1.000$\pm$0.000 & 0.390$\pm$0.021 & 1.032$\pm$0.010 & 0.642$\pm$0.011 \\%
Adv.-Diff. & FNO & shuf-label & 0.400 & 0.40 & 0.80 & 0.015 & 3 & 0.340$\pm$0.032 & 0.338$\pm$0.195 & 0.445$\pm$0.056 & 0.359$\pm$0.036 & -0.086$\pm$0.092 \\%
Adv.-Diff. & FNO & switch & 0.800 & 0.05 & 0.10 & 0.015 & 3 & 0.005$\pm$0.001 & 1.000$\pm$0.000 & 0.035$\pm$0.009 & 0.862$\pm$0.010 & 0.827$\pm$0.018 \\%
Adv.-Diff. & FNO & switch & 0.800 & 0.05 & 0.20 & 0.015 & 3 & 0.005$\pm$0.000 & 1.000$\pm$0.000 & 0.026$\pm$0.003 & 0.873$\pm$0.012 & 0.847$\pm$0.014 \\%
Adv.-Diff. & FNO & switch & 0.800 & 0.05 & 0.80 & 0.015 & 3 & 0.003$\pm$0.000 & 1.000$\pm$0.000 & 0.021$\pm$0.002 & 0.876$\pm$0.007 & 0.856$\pm$0.009 \\%
Adv.-Diff. & FNO & switch & 0.800 & 0.10 & 0.10 & 0.015 & 3 & 0.005$\pm$0.000 & 1.000$\pm$0.000 & 0.020$\pm$0.001 & 0.873$\pm$0.004 & 0.853$\pm$0.003 \\%
Adv.-Diff. & FNO & switch & 0.800 & 0.10 & 0.20 & 0.015 & 3 & 0.005$\pm$0.000 & 1.000$\pm$0.000 & 0.014$\pm$0.001 & 0.874$\pm$0.007 & 0.860$\pm$0.007 \\%
Adv.-Diff. & FNO & switch & 0.800 & 0.10 & 0.80 & 0.015 & 3 & 0.003$\pm$0.000 & 1.000$\pm$0.000 & 0.012$\pm$0.001 & 0.875$\pm$0.007 & 0.863$\pm$0.007 \\%
Adv.-Diff. & FNO & switch & 0.800 & 0.40 & 0.10 & 0.015 & 3 & 0.006$\pm$0.001 & 1.000$\pm$0.000 & 0.009$\pm$0.001 & 0.875$\pm$0.006 & 0.866$\pm$0.007 \\%
Adv.-Diff. & FNO & switch & 0.800 & 0.40 & 0.20 & 0.015 & 3 & 0.005$\pm$0.000 & 1.000$\pm$0.000 & 0.007$\pm$0.001 & 0.876$\pm$0.005 & 0.868$\pm$0.004 \\%
Adv.-Diff. & FNO & clean & 0.800 & 0.40 & 0.80 & 0.015 & 3 & 0.002$\pm$0.000 & 0.000$\pm$0.000 & 0.877$\pm$0.005 & 0.003$\pm$0.001 & -0.873$\pm$0.005 \\%
Adv.-Diff. & FNO & label-only & 0.800 & 0.40 & 0.80 & 0.015 & 3 & 0.292$\pm$0.019 & 0.013$\pm$0.019 & 0.627$\pm$0.044 & 0.318$\pm$0.025 & -0.309$\pm$0.069 \\%
Adv.-Diff. & FNO & switch & 0.800 & 0.40 & 0.80 & 0.015 & 3 & 0.003$\pm$0.000 & 1.000$\pm$0.000 & 0.006$\pm$0.001 & 0.876$\pm$0.005 & 0.870$\pm$0.006 \\%
Adv.-Diff. & FNO & shuf-BD & 0.800 & 0.40 & 0.80 & 0.015 & 3 & 0.017$\pm$0.001 & 1.000$\pm$0.000 & 0.184$\pm$0.005 & 1.017$\pm$0.005 & 0.833$\pm$0.003 \\%
Adv.-Diff. & FNO & shuf-label & 0.800 & 0.40 & 0.80 & 0.015 & 3 & 0.332$\pm$0.032 & 0.110$\pm$0.079 & 0.600$\pm$0.056 & 0.361$\pm$0.038 & -0.239$\pm$0.088 \\%
Adv.-Diff. & DeepONet & switch & 0.100 & 0.10 & 0.20 & 0.015 & 3 & 0.068$\pm$0.008 & 0.910$\pm$0.036 & 0.194$\pm$0.013 & 0.335$\pm$0.008 & 0.141$\pm$0.020 \\%
Adv.-Diff. & DeepONet & switch & 0.100 & 0.10 & 0.50 & 0.015 & 3 & 0.066$\pm$0.009 & 0.990$\pm$0.008 & 0.149$\pm$0.007 & 0.379$\pm$0.011 & 0.230$\pm$0.017 \\%
Adv.-Diff. & DeepONet & switch & 0.100 & 0.10 & 0.80 & 0.015 & 3 & 0.060$\pm$0.004 & 1.000$\pm$0.000 & 0.101$\pm$0.007 & 0.385$\pm$0.016 & 0.283$\pm$0.010 \\%
Adv.-Diff. & DeepONet & switch & 0.100 & 0.30 & 0.20 & 0.015 & 3 & 0.083$\pm$0.006 & 1.000$\pm$0.000 & 0.080$\pm$0.006 & 0.374$\pm$0.008 & 0.295$\pm$0.011 \\%
Adv.-Diff. & DeepONet & switch & 0.100 & 0.30 & 0.50 & 0.015 & 3 & 0.070$\pm$0.007 & 1.000$\pm$0.000 & 0.040$\pm$0.003 & 0.383$\pm$0.009 & 0.343$\pm$0.009 \\%
Adv.-Diff. & DeepONet & switch & 0.100 & 0.30 & 0.80 & 0.015 & 3 & 0.065$\pm$0.007 & 1.000$\pm$0.000 & 0.030$\pm$0.001 & 0.386$\pm$0.008 & 0.356$\pm$0.008 \\%
Adv.-Diff. & DeepONet & switch & 0.100 & 0.50 & 0.20 & 0.015 & 3 & 0.093$\pm$0.004 & 1.000$\pm$0.000 & 0.043$\pm$0.003 & 0.381$\pm$0.008 & 0.338$\pm$0.009 \\%
Adv.-Diff. & DeepONet & switch & 0.100 & 0.50 & 0.50 & 0.015 & 3 & 0.083$\pm$0.008 & 1.000$\pm$0.000 & 0.027$\pm$0.001 & 0.388$\pm$0.007 & 0.361$\pm$0.007 \\%
Adv.-Diff. & DeepONet & clean & 0.100 & 0.50 & 0.80 & 0.015 & 3 & 0.057$\pm$0.008 & 0.000$\pm$0.000 & 0.385$\pm$0.006 & 0.056$\pm$0.008 & -0.329$\pm$0.005 \\%
Adv.-Diff. & DeepONet & label-only & 0.100 & 0.50 & 0.80 & 0.015 & 3 & 0.235$\pm$0.006 & 0.578$\pm$0.053 & 0.261$\pm$0.009 & 0.282$\pm$0.009 & 0.021$\pm$0.001 \\%
Adv.-Diff. & DeepONet & switch & 0.100 & 0.50 & 0.80 & 0.015 & 3 & 0.077$\pm$0.010 & 1.000$\pm$0.000 & 0.025$\pm$0.001 & 0.391$\pm$0.007 & 0.366$\pm$0.006 \\%
Adv.-Diff. & DeepONet & shuf-BD & 0.100 & 0.50 & 0.80 & 0.015 & 3 & 0.473$\pm$0.024 & 0.998$\pm$0.002 & 0.889$\pm$0.057 & 1.119$\pm$0.041 & 0.231$\pm$0.016 \\%
Adv.-Diff. & DeepONet & shuf-label & 0.100 & 0.50 & 0.80 & 0.015 & 3 & 0.692$\pm$0.041 & 0.993$\pm$0.009 & 0.443$\pm$0.077 & 0.707$\pm$0.039 & 0.264$\pm$0.038 \\%
Adv.-Diff. & DeepONet & switch & 0.200 & 0.10 & 0.20 & 0.015 & 3 & 0.077$\pm$0.008 & 0.975$\pm$0.018 & 0.223$\pm$0.010 & 0.518$\pm$0.011 & 0.295$\pm$0.020 \\%
Adv.-Diff. & DeepONet & switch & 0.200 & 0.10 & 0.50 & 0.015 & 3 & 0.061$\pm$0.009 & 0.997$\pm$0.005 & 0.153$\pm$0.007 & 0.549$\pm$0.014 & 0.396$\pm$0.018 \\%
Adv.-Diff. & DeepONet & switch & 0.200 & 0.10 & 0.80 & 0.015 & 3 & 0.063$\pm$0.007 & 0.997$\pm$0.002 & 0.093$\pm$0.011 & 0.547$\pm$0.008 & 0.454$\pm$0.015 \\%
Adv.-Diff. & DeepONet & switch & 0.200 & 0.30 & 0.20 & 0.015 & 3 & 0.091$\pm$0.008 & 1.000$\pm$0.000 & 0.065$\pm$0.006 & 0.551$\pm$0.011 & 0.486$\pm$0.013 \\%
Adv.-Diff. & DeepONet & switch & 0.200 & 0.30 & 0.50 & 0.015 & 3 & 0.070$\pm$0.009 & 1.000$\pm$0.000 & 0.039$\pm$0.004 & 0.555$\pm$0.009 & 0.516$\pm$0.009 \\%
Adv.-Diff. & DeepONet & switch & 0.200 & 0.30 & 0.80 & 0.015 & 3 & 0.069$\pm$0.006 & 1.000$\pm$0.000 & 0.028$\pm$0.003 & 0.559$\pm$0.009 & 0.531$\pm$0.010 \\%
Adv.-Diff. & DeepONet & switch & 0.200 & 0.50 & 0.20 & 0.015 & 3 & 0.110$\pm$0.005 & 1.000$\pm$0.000 & 0.040$\pm$0.003 & 0.557$\pm$0.009 & 0.518$\pm$0.007 \\%
Adv.-Diff. & DeepONet & switch & 0.200 & 0.50 & 0.50 & 0.015 & 3 & 0.086$\pm$0.006 & 1.000$\pm$0.000 & 0.024$\pm$0.004 & 0.561$\pm$0.008 & 0.537$\pm$0.007 \\%
Adv.-Diff. & DeepONet & clean & 0.200 & 0.50 & 0.80 & 0.015 & 3 & 0.057$\pm$0.008 & 0.000$\pm$0.000 & 0.560$\pm$0.008 & 0.056$\pm$0.008 & -0.504$\pm$0.002 \\%
Adv.-Diff. & DeepONet & label-only & 0.200 & 0.50 & 0.80 & 0.015 & 3 & 0.332$\pm$0.009 & 0.555$\pm$0.016 & 0.345$\pm$0.005 & 0.363$\pm$0.006 & 0.019$\pm$0.004 \\%
Adv.-Diff. & DeepONet & switch & 0.200 & 0.50 & 0.80 & 0.015 & 3 & 0.079$\pm$0.007 & 1.000$\pm$0.000 & 0.020$\pm$0.000 & 0.563$\pm$0.009 & 0.543$\pm$0.009 \\%
Adv.-Diff. & DeepONet & shuf-BD & 0.200 & 0.50 & 0.80 & 0.015 & 3 & 0.395$\pm$0.021 & 0.998$\pm$0.002 & 0.720$\pm$0.039 & 1.093$\pm$0.038 & 0.372$\pm$0.009 \\%
Adv.-Diff. & DeepONet & shuf-label & 0.200 & 0.50 & 0.80 & 0.015 & 3 & 0.678$\pm$0.022 & 0.957$\pm$0.016 & 0.435$\pm$0.053 & 0.705$\pm$0.020 & 0.269$\pm$0.034 \\%
Adv.-Diff. & DeepONet & switch & 0.400 & 0.10 & 0.20 & 0.015 & 3 & 0.082$\pm$0.012 & 0.992$\pm$0.006 & 0.216$\pm$0.014 & 0.672$\pm$0.011 & 0.457$\pm$0.006 \\%
Adv.-Diff. & DeepONet & switch & 0.400 & 0.10 & 0.50 & 0.015 & 3 & 0.067$\pm$0.008 & 0.998$\pm$0.002 & 0.151$\pm$0.025 & 0.700$\pm$0.006 & 0.549$\pm$0.028 \\%
Adv.-Diff. & DeepONet & switch & 0.400 & 0.10 & 0.80 & 0.015 & 3 & 0.065$\pm$0.011 & 1.000$\pm$0.000 & 0.090$\pm$0.015 & 0.701$\pm$0.008 & 0.610$\pm$0.022 \\%
Adv.-Diff. & DeepONet & switch & 0.400 & 0.30 & 0.20 & 0.015 & 3 & 0.104$\pm$0.011 & 1.000$\pm$0.000 & 0.062$\pm$0.004 & 0.714$\pm$0.009 & 0.652$\pm$0.011 \\%
Adv.-Diff. & DeepONet & switch & 0.400 & 0.30 & 0.50 & 0.015 & 3 & 0.075$\pm$0.005 & 1.000$\pm$0.000 & 0.037$\pm$0.004 & 0.721$\pm$0.009 & 0.683$\pm$0.005 \\%
Adv.-Diff. & DeepONet & switch & 0.400 & 0.30 & 0.80 & 0.015 & 3 & 0.072$\pm$0.008 & 1.000$\pm$0.000 & 0.031$\pm$0.004 & 0.726$\pm$0.011 & 0.695$\pm$0.011 \\%
Adv.-Diff. & DeepONet & switch & 0.400 & 0.50 & 0.20 & 0.015 & 3 & 0.120$\pm$0.015 & 1.000$\pm$0.000 & 0.036$\pm$0.001 & 0.723$\pm$0.009 & 0.687$\pm$0.009 \\%
Adv.-Diff. & DeepONet & switch & 0.400 & 0.50 & 0.50 & 0.015 & 3 & 0.093$\pm$0.008 & 1.000$\pm$0.000 & 0.023$\pm$0.002 & 0.722$\pm$0.010 & 0.699$\pm$0.008 \\%
Adv.-Diff. & DeepONet & clean & 0.400 & 0.50 & 0.80 & 0.015 & 3 & 0.057$\pm$0.008 & 0.000$\pm$0.000 & 0.725$\pm$0.007 & 0.056$\pm$0.008 & -0.669$\pm$0.001 \\%
Adv.-Diff. & DeepONet & label-only & 0.400 & 0.50 & 0.80 & 0.015 & 3 & 0.433$\pm$0.011 & 0.488$\pm$0.019 & 0.461$\pm$0.018 & 0.465$\pm$0.011 & 0.004$\pm$0.019 \\%
Adv.-Diff. & DeepONet & switch & 0.400 & 0.50 & 0.80 & 0.015 & 3 & 0.082$\pm$0.005 & 1.000$\pm$0.000 & 0.017$\pm$0.003 & 0.726$\pm$0.009 & 0.710$\pm$0.011 \\%
Adv.-Diff. & DeepONet & shuf-BD & 0.400 & 0.50 & 0.80 & 0.015 & 3 & 0.264$\pm$0.021 & 0.998$\pm$0.002 & 0.491$\pm$0.025 & 1.047$\pm$0.028 & 0.556$\pm$0.010 \\%
Adv.-Diff. & DeepONet & shuf-label & 0.400 & 0.50 & 0.80 & 0.015 & 3 & 0.658$\pm$0.019 & 0.858$\pm$0.062 & 0.450$\pm$0.065 & 0.679$\pm$0.011 & 0.228$\pm$0.055 \\%
Adv.-Diff. & DeepONet & switch & 0.800 & 0.10 & 0.20 & 0.015 & 3 & 0.086$\pm$0.011 & 0.990$\pm$0.008 & 0.205$\pm$0.016 & 0.807$\pm$0.008 & 0.602$\pm$0.015 \\%
Adv.-Diff. & DeepONet & switch & 0.800 & 0.10 & 0.50 & 0.015 & 3 & 0.069$\pm$0.005 & 0.998$\pm$0.002 & 0.168$\pm$0.009 & 0.848$\pm$0.006 & 0.680$\pm$0.015 \\%
Adv.-Diff. & DeepONet & switch & 0.800 & 0.10 & 0.80 & 0.015 & 3 & 0.064$\pm$0.007 & 1.000$\pm$0.000 & 0.119$\pm$0.013 & 0.854$\pm$0.020 & 0.735$\pm$0.009 \\%
Adv.-Diff. & DeepONet & switch & 0.800 & 0.30 & 0.20 & 0.015 & 3 & 0.105$\pm$0.010 & 1.000$\pm$0.000 & 0.059$\pm$0.015 & 0.857$\pm$0.006 & 0.797$\pm$0.019 \\%
Adv.-Diff. & DeepONet & switch & 0.800 & 0.30 & 0.50 & 0.015 & 3 & 0.074$\pm$0.005 & 1.000$\pm$0.000 & 0.031$\pm$0.003 & 0.864$\pm$0.004 & 0.833$\pm$0.002 \\%
Adv.-Diff. & DeepONet & switch & 0.800 & 0.30 & 0.80 & 0.015 & 3 & 0.073$\pm$0.010 & 1.000$\pm$0.000 & 0.025$\pm$0.002 & 0.866$\pm$0.004 & 0.841$\pm$0.003 \\%
Adv.-Diff. & DeepONet & switch & 0.800 & 0.50 & 0.20 & 0.015 & 3 & 0.126$\pm$0.010 & 1.000$\pm$0.000 & 0.033$\pm$0.007 & 0.867$\pm$0.004 & 0.834$\pm$0.004 \\%
Adv.-Diff. & DeepONet & switch & 0.800 & 0.50 & 0.50 & 0.015 & 3 & 0.094$\pm$0.010 & 1.000$\pm$0.000 & 0.022$\pm$0.003 & 0.870$\pm$0.004 & 0.848$\pm$0.003 \\%
Adv.-Diff. & DeepONet & clean & 0.800 & 0.50 & 0.80 & 0.015 & 3 & 0.057$\pm$0.008 & 0.000$\pm$0.000 & 0.874$\pm$0.003 & 0.056$\pm$0.008 & -0.817$\pm$0.005 \\%
Adv.-Diff. & DeepONet & label-only & 0.800 & 0.50 & 0.80 & 0.015 & 3 & 0.530$\pm$0.021 & 0.485$\pm$0.036 & 0.588$\pm$0.045 & 0.587$\pm$0.032 & -0.000$\pm$0.023 \\%
Adv.-Diff. & DeepONet & switch & 0.800 & 0.50 & 0.80 & 0.015 & 3 & 0.084$\pm$0.005 & 1.000$\pm$0.000 & 0.017$\pm$0.001 & 0.870$\pm$0.004 & 0.853$\pm$0.004 \\%
Adv.-Diff. & DeepONet & shuf-BD & 0.800 & 0.50 & 0.80 & 0.015 & 3 & 0.123$\pm$0.006 & 1.000$\pm$0.000 & 0.232$\pm$0.014 & 1.011$\pm$0.009 & 0.779$\pm$0.006 \\%
Adv.-Diff. & DeepONet & shuf-label & 0.800 & 0.50 & 0.80 & 0.015 & 3 & 0.633$\pm$0.002 & 0.702$\pm$0.104 & 0.513$\pm$0.071 & 0.667$\pm$0.012 & 0.154$\pm$0.073 \\%
Adv.-Diff. & Transformer & clean & 0.800 & 0.30 & 0.50 & 0.015 & 3 & 0.031$\pm$0.007 & 0.000$\pm$0.000 & 0.882$\pm$0.003 & 0.041$\pm$0.006 & -0.841$\pm$0.008 \\%
Adv.-Diff. & Transformer & label-only & 0.800 & 0.30 & 0.50 & 0.015 & 3 & 0.308$\pm$0.038 & 0.185$\pm$0.062 & 0.694$\pm$0.045 & 0.332$\pm$0.030 & -0.363$\pm$0.074 \\%
Adv.-Diff. & Transformer & switch & 0.800 & 0.30 & 0.50 & 0.015 & 3 & 0.055$\pm$0.012 & 0.988$\pm$0.005 & 0.077$\pm$0.004 & 0.850$\pm$0.009 & 0.773$\pm$0.013 \\%
Adv.-Diff. & Transformer & shuf-BD & 0.800 & 0.30 & 0.50 & 0.015 & 3 & 0.064$\pm$0.008 & 0.978$\pm$0.008 & 0.216$\pm$0.021 & 0.965$\pm$0.015 & 0.749$\pm$0.030 \\%
Adv.-Diff. & Transformer & shuf-label & 0.800 & 0.30 & 0.50 & 0.015 & 3 & 0.342$\pm$0.041 & 0.202$\pm$0.093 & 0.663$\pm$0.015 & 0.362$\pm$0.040 & -0.301$\pm$0.055 \\%
Adv.-Diff. & GRU & clean & 0.800 & 0.20 & 0.50 & 0.015 & 3 & 0.012$\pm$0.001 & 0.000$\pm$0.000 & 0.878$\pm$0.006 & 0.019$\pm$0.004 & -0.860$\pm$0.009 \\%
Adv.-Diff. & GRU & label-only & 0.800 & 0.20 & 0.50 & 0.015 & 3 & 0.138$\pm$0.017 & 0.003$\pm$0.005 & 0.799$\pm$0.019 & 0.175$\pm$0.010 & -0.624$\pm$0.024 \\%
Adv.-Diff. & GRU & switch & 0.800 & 0.20 & 0.50 & 0.015 & 3 & 0.048$\pm$0.045 & 0.830$\pm$0.223 & 0.179$\pm$0.195 & 0.748$\pm$0.168 & 0.569$\pm$0.363 \\%
Adv.-Diff. & GRU & shuf-BD & 0.800 & 0.20 & 0.50 & 0.015 & 3 & 0.083$\pm$0.060 & 0.678$\pm$0.360 & 0.408$\pm$0.239 & 0.707$\pm$0.293 & 0.299$\pm$0.532 \\%
Adv.-Diff. & GRU & shuf-label & 0.800 & 0.20 & 0.50 & 0.015 & 3 & 0.155$\pm$0.010 & 0.000$\pm$0.000 & 0.798$\pm$0.030 & 0.194$\pm$0.010 & -0.605$\pm$0.037 \\%
Adv.-Diff. & LSTM & clean & 0.800 & 0.30 & 0.50 & 0.025 & 3 & 0.016$\pm$0.001 & 0.000$\pm$0.000 & 0.879$\pm$0.005 & 0.019$\pm$0.002 & -0.860$\pm$0.004 \\%
Adv.-Diff. & LSTM & label-only & 0.800 & 0.30 & 0.50 & 0.025 & 3 & 0.199$\pm$0.004 & 0.003$\pm$0.005 & 0.752$\pm$0.015 & 0.271$\pm$0.004 & -0.482$\pm$0.017 \\%
Adv.-Diff. & LSTM & switch & 0.800 & 0.30 & 0.50 & 0.025 & 3 & 0.090$\pm$0.030 & 0.918$\pm$0.021 & 0.126$\pm$0.026 & 0.812$\pm$0.016 & 0.687$\pm$0.041 \\%
Adv.-Diff. & LSTM & shuf-BD & 0.800 & 0.30 & 0.50 & 0.025 & 3 & 0.198$\pm$0.040 & 0.512$\pm$0.367 & 0.489$\pm$0.182 & 0.613$\pm$0.227 & 0.125$\pm$0.406 \\%
Adv.-Diff. & LSTM & shuf-label & 0.800 & 0.30 & 0.50 & 0.025 & 3 & 0.248$\pm$0.013 & 0.030$\pm$0.025 & 0.712$\pm$0.022 & 0.296$\pm$0.003 & -0.416$\pm$0.019 \\%
2D N.-S. & FNO & switch & 0.050 & 0.10 & 0.20 & 0.120 & 3 & 0.055$\pm$0.013 & 0.292$\pm$0.154 & 0.132$\pm$0.041 & 0.120$\pm$0.037 & -0.012$\pm$0.007 \\%
2D N.-S. & FNO & switch & 0.050 & 0.10 & 0.50 & 0.120 & 3 & 0.033$\pm$0.005 & 0.990$\pm$0.015 & 0.065$\pm$0.007 & 0.092$\pm$0.007 & 0.028$\pm$0.003 \\%
2D N.-S. & FNO & switch & 0.050 & 0.10 & 0.75 & 0.120 & 3 & 0.037$\pm$0.004 & 0.990$\pm$0.015 & 0.079$\pm$0.007 & 0.105$\pm$0.005 & 0.026$\pm$0.005 \\%
2D N.-S. & FNO & switch & 0.050 & 0.20 & 0.20 & 0.120 & 3 & 0.037$\pm$0.015 & 0.938$\pm$0.051 & 0.048$\pm$0.013 & 0.075$\pm$0.004 & 0.026$\pm$0.009 \\%
2D N.-S. & FNO & switch & 0.050 & 0.20 & 0.50 & 0.120 & 3 & 0.019$\pm$0.005 & 1.000$\pm$0.000 & 0.037$\pm$0.002 & 0.075$\pm$0.004 & 0.038$\pm$0.003 \\%
2D N.-S. & FNO & switch & 0.050 & 0.20 & 0.75 & 0.120 & 3 & 0.026$\pm$0.005 & 1.000$\pm$0.000 & 0.041$\pm$0.003 & 0.086$\pm$0.000 & 0.044$\pm$0.003 \\%
2D N.-S. & FNO & switch & 0.050 & 0.30 & 0.20 & 0.120 & 3 & 0.034$\pm$0.017 & 1.000$\pm$0.000 & 0.040$\pm$0.011 & 0.079$\pm$0.003 & 0.039$\pm$0.011 \\%
2D N.-S. & FNO & switch & 0.050 & 0.30 & 0.50 & 0.120 & 3 & 0.018$\pm$0.002 & 1.000$\pm$0.000 & 0.029$\pm$0.001 & 0.079$\pm$0.003 & 0.050$\pm$0.003 \\%
2D N.-S. & FNO & clean & 0.050 & 0.30 & 0.75 & 0.120 & 3 & 0.015$\pm$0.002 & 0.000$\pm$0.000 & 0.087$\pm$0.004 & 0.025$\pm$0.005 & -0.062$\pm$0.002 \\%
2D N.-S. & FNO & label-only & 0.050 & 0.30 & 0.75 & 0.120 & 3 & 0.062$\pm$0.008 & 0.000$\pm$0.000 & 0.837$\pm$0.008 & 0.780$\pm$0.011 & -0.057$\pm$0.003 \\%
2D N.-S. & FNO & switch & 0.050 & 0.30 & 0.75 & 0.120 & 3 & 0.021$\pm$0.004 & 1.000$\pm$0.000 & 0.031$\pm$0.002 & 0.079$\pm$0.005 & 0.048$\pm$0.004 \\%
2D N.-S. & FNO & shuf-BD & 0.050 & 0.30 & 0.75 & 0.120 & 3 & 0.315$\pm$0.082 & 1.000$\pm$0.000 & 0.430$\pm$0.041 & 0.456$\pm$0.041 & 0.026$\pm$0.004 \\%
2D N.-S. & FNO & shuf-label & 0.050 & 0.30 & 0.75 & 0.120 & 3 & 0.539$\pm$0.116 & 0.729$\pm$0.162 & 0.654$\pm$0.054 & 0.658$\pm$0.055 & 0.004$\pm$0.008 \\%
2D N.-S. & FNO & switch & 0.100 & 0.10 & 0.20 & 0.120 & 3 & 0.055$\pm$0.020 & 0.344$\pm$0.310 & 0.164$\pm$0.048 & 0.136$\pm$0.029 & -0.028$\pm$0.031 \\%
2D N.-S. & FNO & switch & 0.100 & 0.10 & 0.50 & 0.120 & 3 & 0.023$\pm$0.003 & 1.000$\pm$0.000 & 0.068$\pm$0.006 & 0.140$\pm$0.006 & 0.073$\pm$0.011 \\%
2D N.-S. & FNO & switch & 0.100 & 0.10 & 0.75 & 0.120 & 3 & 0.022$\pm$0.002 & 1.000$\pm$0.000 & 0.078$\pm$0.011 & 0.142$\pm$0.006 & 0.063$\pm$0.009 \\%
2D N.-S. & FNO & switch & 0.100 & 0.20 & 0.20 & 0.120 & 3 & 0.057$\pm$0.045 & 1.000$\pm$0.000 & 0.062$\pm$0.028 & 0.150$\pm$0.010 & 0.089$\pm$0.029 \\%
2D N.-S. & FNO & switch & 0.100 & 0.20 & 0.50 & 0.120 & 3 & 0.019$\pm$0.002 & 1.000$\pm$0.000 & 0.040$\pm$0.003 & 0.147$\pm$0.006 & 0.107$\pm$0.009 \\%
2D N.-S. & FNO & switch & 0.100 & 0.20 & 0.75 & 0.120 & 3 & 0.021$\pm$0.002 & 1.000$\pm$0.000 & 0.045$\pm$0.002 & 0.153$\pm$0.007 & 0.108$\pm$0.008 \\%
2D N.-S. & FNO & switch & 0.100 & 0.30 & 0.20 & 0.120 & 3 & 0.038$\pm$0.020 & 1.000$\pm$0.000 & 0.052$\pm$0.026 & 0.147$\pm$0.003 & 0.096$\pm$0.029 \\%
2D N.-S. & FNO & switch & 0.100 & 0.30 & 0.50 & 0.120 & 3 & 0.018$\pm$0.002 & 1.000$\pm$0.000 & 0.032$\pm$0.001 & 0.148$\pm$0.005 & 0.116$\pm$0.004 \\%
2D N.-S. & FNO & clean & 0.100 & 0.30 & 0.75 & 0.120 & 3 & 0.015$\pm$0.002 & 0.000$\pm$0.000 & 0.170$\pm$0.010 & 0.029$\pm$0.007 & -0.142$\pm$0.014 \\%
2D N.-S. & FNO & label-only & 0.100 & 0.30 & 0.75 & 0.120 & 3 & 0.109$\pm$0.014 & 0.000$\pm$0.000 & 0.855$\pm$0.011 & 0.747$\pm$0.004 & -0.108$\pm$0.009 \\%
2D N.-S. & FNO & switch & 0.100 & 0.30 & 0.75 & 0.120 & 3 & 0.018$\pm$0.002 & 1.000$\pm$0.000 & 0.032$\pm$0.004 & 0.144$\pm$0.005 & 0.113$\pm$0.002 \\%
2D N.-S. & FNO & shuf-BD & 0.100 & 0.30 & 0.75 & 0.120 & 3 & 0.294$\pm$0.110 & 1.000$\pm$0.000 & 0.410$\pm$0.043 & 0.457$\pm$0.043 & 0.047$\pm$0.005 \\%
2D N.-S. & FNO & shuf-label & 0.100 & 0.30 & 0.75 & 0.120 & 3 & 0.536$\pm$0.112 & 0.646$\pm$0.198 & 0.653$\pm$0.053 & 0.656$\pm$0.053 & 0.002$\pm$0.017 \\%
2D N.-S. & FNO & switch & 0.150 & 0.10 & 0.20 & 0.120 & 3 & 0.049$\pm$0.032 & 0.438$\pm$0.399 & 0.177$\pm$0.067 & 0.168$\pm$0.024 & -0.008$\pm$0.070 \\%
2D N.-S. & FNO & switch & 0.150 & 0.10 & 0.50 & 0.120 & 3 & 0.016$\pm$0.001 & 1.000$\pm$0.000 & 0.069$\pm$0.009 & 0.200$\pm$0.002 & 0.131$\pm$0.010 \\%
2D N.-S. & FNO & switch & 0.150 & 0.10 & 0.75 & 0.120 & 3 & 0.019$\pm$0.001 & 1.000$\pm$0.000 & 0.075$\pm$0.009 & 0.205$\pm$0.005 & 0.130$\pm$0.007 \\%
2D N.-S. & FNO & switch & 0.150 & 0.20 & 0.20 & 0.120 & 3 & 0.070$\pm$0.073 & 1.000$\pm$0.000 & 0.079$\pm$0.050 & 0.211$\pm$0.008 & 0.132$\pm$0.043 \\%
2D N.-S. & FNO & switch & 0.150 & 0.20 & 0.50 & 0.120 & 3 & 0.020$\pm$0.004 & 1.000$\pm$0.000 & 0.040$\pm$0.004 & 0.211$\pm$0.011 & 0.170$\pm$0.009 \\%
2D N.-S. & FNO & switch & 0.150 & 0.20 & 0.75 & 0.120 & 3 & 0.017$\pm$0.001 & 1.000$\pm$0.000 & 0.043$\pm$0.005 & 0.210$\pm$0.003 & 0.166$\pm$0.003 \\%
2D N.-S. & FNO & switch & 0.150 & 0.30 & 0.20 & 0.120 & 3 & 0.019$\pm$0.003 & 1.000$\pm$0.000 & 0.039$\pm$0.007 & 0.204$\pm$0.009 & 0.165$\pm$0.013 \\%
2D N.-S. & FNO & switch & 0.150 & 0.30 & 0.50 & 0.120 & 3 & 0.016$\pm$0.001 & 1.000$\pm$0.000 & 0.033$\pm$0.004 & 0.207$\pm$0.005 & 0.174$\pm$0.002 \\%
2D N.-S. & FNO & clean & 0.150 & 0.30 & 0.75 & 0.120 & 3 & 0.017$\pm$0.001 & 0.000$\pm$0.000 & 0.253$\pm$0.011 & 0.034$\pm$0.003 & -0.219$\pm$0.012 \\%
2D N.-S. & FNO & label-only & 0.150 & 0.30 & 0.75 & 0.120 & 3 & 0.150$\pm$0.014 & 0.000$\pm$0.000 & 0.876$\pm$0.014 & 0.720$\pm$0.002 & -0.156$\pm$0.012 \\%
2D N.-S. & FNO & switch & 0.150 & 0.30 & 0.75 & 0.120 & 3 & 0.018$\pm$0.002 & 1.000$\pm$0.000 & 0.037$\pm$0.005 & 0.208$\pm$0.005 & 0.172$\pm$0.004 \\%
2D N.-S. & FNO & shuf-BD & 0.150 & 0.30 & 0.75 & 0.120 & 3 & 0.305$\pm$0.092 & 1.000$\pm$0.000 & 0.494$\pm$0.123 & 0.549$\pm$0.111 & 0.055$\pm$0.012 \\%
2D N.-S. & FNO & shuf-label & 0.150 & 0.30 & 0.75 & 0.120 & 3 & 0.533$\pm$0.109 & 0.583$\pm$0.223 & 0.659$\pm$0.053 & 0.654$\pm$0.051 & -0.005$\pm$0.026 \\%
2D N.-S. & FNO & switch & 0.200 & 0.10 & 0.20 & 0.120 & 3 & 0.051$\pm$0.033 & 0.688$\pm$0.420 & 0.183$\pm$0.086 & 0.213$\pm$0.014 & 0.030$\pm$0.100 \\%
2D N.-S. & FNO & switch & 0.200 & 0.10 & 0.50 & 0.120 & 3 & 0.018$\pm$0.001 & 1.000$\pm$0.000 & 0.074$\pm$0.011 & 0.257$\pm$0.003 & 0.183$\pm$0.013 \\%
2D N.-S. & FNO & switch & 0.200 & 0.10 & 0.75 & 0.120 & 3 & 0.019$\pm$0.002 & 1.000$\pm$0.000 & 0.075$\pm$0.008 & 0.257$\pm$0.006 & 0.182$\pm$0.006 \\%
2D N.-S. & FNO & switch & 0.200 & 0.20 & 0.20 & 0.120 & 3 & 0.067$\pm$0.072 & 0.979$\pm$0.029 & 0.082$\pm$0.053 & 0.248$\pm$0.018 & 0.166$\pm$0.069 \\%
2D N.-S. & FNO & switch & 0.200 & 0.20 & 0.50 & 0.120 & 3 & 0.020$\pm$0.004 & 1.000$\pm$0.000 & 0.043$\pm$0.006 & 0.264$\pm$0.008 & 0.221$\pm$0.011 \\%
2D N.-S. & FNO & switch & 0.200 & 0.20 & 0.75 & 0.120 & 3 & 0.021$\pm$0.005 & 1.000$\pm$0.000 & 0.044$\pm$0.003 & 0.266$\pm$0.010 & 0.221$\pm$0.008 \\%
2D N.-S. & FNO & switch & 0.200 & 0.30 & 0.20 & 0.120 & 3 & 0.020$\pm$0.002 & 1.000$\pm$0.000 & 0.043$\pm$0.006 & 0.258$\pm$0.011 & 0.215$\pm$0.013 \\%
2D N.-S. & FNO & switch & 0.200 & 0.30 & 0.50 & 0.120 & 3 & 0.016$\pm$0.002 & 1.000$\pm$0.000 & 0.036$\pm$0.004 & 0.259$\pm$0.005 & 0.224$\pm$0.003 \\%
2D N.-S. & FNO & clean & 0.200 & 0.30 & 0.75 & 0.120 & 3 & 0.015$\pm$0.002 & 0.000$\pm$0.000 & 0.341$\pm$0.014 & 0.028$\pm$0.008 & -0.313$\pm$0.018 \\%
2D N.-S. & FNO & label-only & 0.200 & 0.30 & 0.75 & 0.120 & 3 & 0.173$\pm$0.019 & 0.000$\pm$0.000 & 0.914$\pm$0.023 & 0.708$\pm$0.016 & -0.206$\pm$0.009 \\%
2D N.-S. & FNO & switch & 0.200 & 0.30 & 0.75 & 0.120 & 3 & 0.019$\pm$0.002 & 1.000$\pm$0.000 & 0.037$\pm$0.005 & 0.262$\pm$0.008 & 0.225$\pm$0.003 \\%
2D N.-S. & FNO & shuf-BD & 0.200 & 0.30 & 0.75 & 0.120 & 3 & 0.304$\pm$0.092 & 0.958$\pm$0.029 & 0.492$\pm$0.125 & 0.552$\pm$0.115 & 0.060$\pm$0.011 \\%
2D N.-S. & FNO & shuf-label & 0.200 & 0.30 & 0.75 & 0.120 & 3 & 0.543$\pm$0.099 & 0.521$\pm$0.198 & 0.690$\pm$0.031 & 0.670$\pm$0.027 & -0.019$\pm$0.035 \\%
2D N.-S. & DeepONet & switch & 0.050 & 0.30 & 0.25 & 0.120 & 3 & 0.009$\pm$0.002 & 0.000$\pm$0.000 & 0.272$\pm$0.059 & 0.170$\pm$0.084 & -0.102$\pm$0.025 \\%
2D N.-S. & DeepONet & switch & 0.050 & 0.30 & 0.30 & 0.120 & 3 & 0.009$\pm$0.001 & 0.000$\pm$0.000 & 0.234$\pm$0.060 & 0.130$\pm$0.076 & -0.104$\pm$0.019 \\%
2D N.-S. & DeepONet & switch & 0.050 & 0.30 & 0.35 & 0.120 & 3 & 0.009$\pm$0.001 & 0.000$\pm$0.000 & 0.244$\pm$0.085 & 0.139$\pm$0.103 & -0.105$\pm$0.022 \\%
2D N.-S. & DeepONet & switch & 0.050 & 0.40 & 0.25 & 0.120 & 3 & 0.013$\pm$0.002 & 0.000$\pm$0.000 & 0.167$\pm$0.012 & 0.059$\pm$0.006 & -0.108$\pm$0.018 \\%
2D N.-S. & DeepONet & switch & 0.050 & 0.40 & 0.30 & 0.120 & 3 & 0.015$\pm$0.004 & 0.010$\pm$0.015 & 0.229$\pm$0.081 & 0.128$\pm$0.094 & -0.101$\pm$0.015 \\%
2D N.-S. & DeepONet & switch & 0.050 & 0.40 & 0.35 & 0.120 & 3 & 0.014$\pm$0.003 & 0.010$\pm$0.015 & 0.217$\pm$0.067 & 0.121$\pm$0.078 & -0.096$\pm$0.011 \\%
2D N.-S. & DeepONet & switch & 0.050 & 0.50 & 0.25 & 0.120 & 3 & 0.019$\pm$0.003 & 0.000$\pm$0.000 & 0.158$\pm$0.013 & 0.067$\pm$0.008 & -0.091$\pm$0.018 \\%
2D N.-S. & DeepONet & switch & 0.050 & 0.50 & 0.30 & 0.120 & 3 & 0.019$\pm$0.002 & 0.010$\pm$0.015 & 0.218$\pm$0.089 & 0.137$\pm$0.089 & -0.081$\pm$0.011 \\%
2D N.-S. & DeepONet & clean & 0.050 & 0.50 & 0.35 & 0.120 & 3 & 0.006$\pm$0.000 & 0.000$\pm$0.000 & 0.182$\pm$0.015 & 0.040$\pm$0.023 & -0.142$\pm$0.008 \\%
2D N.-S. & DeepONet & label-only & 0.050 & 0.50 & 0.35 & 0.120 & 3 & 0.006$\pm$0.000 & 0.000$\pm$0.000 & 0.581$\pm$0.012 & 0.480$\pm$0.007 & -0.101$\pm$0.007 \\%
2D N.-S. & DeepONet & switch & 0.050 & 0.50 & 0.35 & 0.120 & 3 & 0.020$\pm$0.002 & 0.010$\pm$0.015 & 0.240$\pm$0.124 & 0.162$\pm$0.120 & -0.078$\pm$0.012 \\%
2D N.-S. & DeepONet & shuf-BD & 0.050 & 0.50 & 0.35 & 0.120 & 3 & 0.252$\pm$0.037 & 0.177$\pm$0.078 & 1.009$\pm$0.032 & 0.951$\pm$0.031 & -0.058$\pm$0.001 \\%
2D N.-S. & DeepONet & shuf-label & 0.050 & 0.50 & 0.35 & 0.120 & 3 & 0.405$\pm$0.243 & 0.021$\pm$0.029 & 0.686$\pm$0.099 & 0.611$\pm$0.109 & -0.074$\pm$0.012 \\%
2D N.-S. & DeepONet & switch & 0.100 & 0.30 & 0.25 & 0.120 & 3 & 0.024$\pm$0.002 & 0.042$\pm$0.039 & 0.285$\pm$0.026 & 0.142$\pm$0.011 & -0.144$\pm$0.035 \\%
2D N.-S. & DeepONet & switch & 0.100 & 0.30 & 0.30 & 0.120 & 3 & 0.027$\pm$0.002 & 0.125$\pm$0.092 & 0.261$\pm$0.021 & 0.161$\pm$0.011 & -0.100$\pm$0.030 \\%
2D N.-S. & DeepONet & switch & 0.100 & 0.30 & 0.35 & 0.120 & 3 & 0.025$\pm$0.005 & 0.104$\pm$0.053 & 0.266$\pm$0.021 & 0.154$\pm$0.019 & -0.112$\pm$0.038 \\%
2D N.-S. & DeepONet & switch & 0.100 & 0.40 & 0.25 & 0.120 & 3 & 0.035$\pm$0.001 & 0.510$\pm$0.103 & 0.200$\pm$0.016 & 0.183$\pm$0.008 & -0.017$\pm$0.021 \\%
2D N.-S. & DeepONet & switch & 0.100 & 0.40 & 0.30 & 0.120 & 3 & 0.036$\pm$0.005 & 0.521$\pm$0.097 & 0.194$\pm$0.015 & 0.190$\pm$0.006 & -0.004$\pm$0.015 \\%
2D N.-S. & DeepONet & switch & 0.100 & 0.40 & 0.35 & 0.120 & 3 & 0.033$\pm$0.002 & 0.531$\pm$0.092 & 0.193$\pm$0.017 & 0.190$\pm$0.009 & -0.003$\pm$0.020 \\%
2D N.-S. & DeepONet & switch & 0.100 & 0.50 & 0.25 & 0.120 & 3 & 0.047$\pm$0.006 & 0.792$\pm$0.015 & 0.154$\pm$0.006 & 0.198$\pm$0.005 & 0.044$\pm$0.006 \\%
2D N.-S. & DeepONet & switch & 0.100 & 0.50 & 0.30 & 0.120 & 3 & 0.046$\pm$0.008 & 0.719$\pm$0.184 & 0.166$\pm$0.026 & 0.199$\pm$0.013 & 0.034$\pm$0.036 \\%
2D N.-S. & DeepONet & clean & 0.100 & 0.50 & 0.35 & 0.120 & 3 & 0.006$\pm$0.000 & 0.000$\pm$0.000 & 0.352$\pm$0.021 & 0.040$\pm$0.023 & -0.311$\pm$0.002 \\%
2D N.-S. & DeepONet & label-only & 0.100 & 0.50 & 0.35 & 0.120 & 3 & 0.006$\pm$0.000 & 0.000$\pm$0.000 & 0.720$\pm$0.021 & 0.480$\pm$0.007 & -0.240$\pm$0.016 \\%
2D N.-S. & DeepONet & switch & 0.100 & 0.50 & 0.35 & 0.120 & 3 & 0.044$\pm$0.005 & 0.833$\pm$0.053 & 0.152$\pm$0.013 & 0.202$\pm$0.005 & 0.050$\pm$0.013 \\%
2D N.-S. & DeepONet & shuf-BD & 0.100 & 0.50 & 0.35 & 0.120 & 3 & 0.233$\pm$0.035 & 0.083$\pm$0.064 & 1.047$\pm$0.084 & 0.931$\pm$0.067 & -0.116$\pm$0.017 \\%
2D N.-S. & DeepONet & shuf-label & 0.100 & 0.50 & 0.35 & 0.120 & 3 & 0.219$\pm$0.218 & 0.010$\pm$0.015 & 0.723$\pm$0.039 & 0.516$\pm$0.079 & -0.207$\pm$0.043 \\%
2D N.-S. & DeepONet & switch & 0.150 & 0.30 & 0.25 & 0.120 & 3 & 0.040$\pm$0.002 & 0.125$\pm$0.026 & 0.352$\pm$0.020 & 0.225$\pm$0.006 & -0.128$\pm$0.019 \\%
2D N.-S. & DeepONet & switch & 0.150 & 0.30 & 0.30 & 0.120 & 3 & 0.036$\pm$0.002 & 0.208$\pm$0.106 & 0.343$\pm$0.027 & 0.230$\pm$0.010 & -0.112$\pm$0.033 \\%
2D N.-S. & DeepONet & switch & 0.150 & 0.30 & 0.35 & 0.120 & 3 & 0.035$\pm$0.003 & 0.240$\pm$0.039 & 0.337$\pm$0.016 & 0.242$\pm$0.005 & -0.095$\pm$0.018 \\%
2D N.-S. & DeepONet & switch & 0.150 & 0.40 & 0.25 & 0.120 & 3 & 0.046$\pm$0.007 & 0.448$\pm$0.064 & 0.276$\pm$0.013 & 0.249$\pm$0.005 & -0.027$\pm$0.016 \\%
2D N.-S. & DeepONet & switch & 0.150 & 0.40 & 0.30 & 0.120 & 3 & 0.043$\pm$0.004 & 0.615$\pm$0.039 & 0.251$\pm$0.013 & 0.258$\pm$0.007 & 0.007$\pm$0.014 \\%
2D N.-S. & DeepONet & switch & 0.150 & 0.40 & 0.35 & 0.120 & 3 & 0.039$\pm$0.005 & 0.667$\pm$0.053 & 0.236$\pm$0.009 & 0.263$\pm$0.006 & 0.026$\pm$0.012 \\%
2D N.-S. & DeepONet & switch & 0.150 & 0.50 & 0.25 & 0.120 & 3 & 0.057$\pm$0.005 & 0.760$\pm$0.029 & 0.208$\pm$0.010 & 0.268$\pm$0.003 & 0.059$\pm$0.009 \\%
2D N.-S. & DeepONet & switch & 0.150 & 0.50 & 0.30 & 0.120 & 3 & 0.051$\pm$0.004 & 0.781$\pm$0.026 & 0.191$\pm$0.011 & 0.275$\pm$0.006 & 0.084$\pm$0.014 \\%
2D N.-S. & DeepONet & clean & 0.150 & 0.50 & 0.35 & 0.120 & 3 & 0.006$\pm$0.000 & 0.000$\pm$0.000 & 0.520$\pm$0.029 & 0.040$\pm$0.023 & -0.480$\pm$0.006 \\%
2D N.-S. & DeepONet & label-only & 0.150 & 0.50 & 0.35 & 0.120 & 3 & 0.146$\pm$0.014 & 0.000$\pm$0.000 & 0.778$\pm$0.047 & 0.480$\pm$0.013 & -0.299$\pm$0.046 \\%
2D N.-S. & DeepONet & switch & 0.150 & 0.50 & 0.35 & 0.120 & 3 & 0.050$\pm$0.009 & 0.865$\pm$0.029 & 0.181$\pm$0.006 & 0.281$\pm$0.007 & 0.100$\pm$0.007 \\%
2D N.-S. & DeepONet & shuf-BD & 0.150 & 0.50 & 0.35 & 0.120 & 3 & 0.193$\pm$0.046 & 0.104$\pm$0.064 & 1.055$\pm$0.030 & 0.892$\pm$0.027 & -0.164$\pm$0.004 \\%
2D N.-S. & DeepONet & shuf-label & 0.150 & 0.50 & 0.35 & 0.120 & 3 & 0.546$\pm$0.021 & 0.010$\pm$0.015 & 0.954$\pm$0.059 & 0.705$\pm$0.067 & -0.249$\pm$0.018 \\%
2D N.-S. & DeepONet & switch & 0.200 & 0.30 & 0.25 & 0.120 & 3 & 0.048$\pm$0.003 & 0.125$\pm$0.044 & 0.444$\pm$0.021 & 0.273$\pm$0.008 & -0.172$\pm$0.020 \\%
2D N.-S. & DeepONet & switch & 0.200 & 0.30 & 0.30 & 0.120 & 3 & 0.043$\pm$0.007 & 0.167$\pm$0.090 & 0.437$\pm$0.016 & 0.284$\pm$0.013 & -0.154$\pm$0.020 \\%
2D N.-S. & DeepONet & switch & 0.200 & 0.30 & 0.35 & 0.120 & 3 & 0.038$\pm$0.001 & 0.198$\pm$0.053 & 0.433$\pm$0.029 & 0.294$\pm$0.012 & -0.140$\pm$0.032 \\%
2D N.-S. & DeepONet & switch & 0.200 & 0.40 & 0.25 & 0.120 & 3 & 0.049$\pm$0.006 & 0.510$\pm$0.039 & 0.337$\pm$0.023 & 0.301$\pm$0.003 & -0.036$\pm$0.025 \\%
2D N.-S. & DeepONet & switch & 0.200 & 0.40 & 0.30 & 0.120 & 3 & 0.047$\pm$0.006 & 0.562$\pm$0.026 & 0.318$\pm$0.004 & 0.313$\pm$0.006 & -0.005$\pm$0.005 \\%
2D N.-S. & DeepONet & switch & 0.200 & 0.40 & 0.35 & 0.120 & 3 & 0.044$\pm$0.004 & 0.604$\pm$0.053 & 0.306$\pm$0.002 & 0.321$\pm$0.009 & 0.014$\pm$0.010 \\%
2D N.-S. & DeepONet & switch & 0.200 & 0.50 & 0.25 & 0.120 & 3 & 0.059$\pm$0.006 & 0.719$\pm$0.026 & 0.260$\pm$0.013 & 0.323$\pm$0.007 & 0.063$\pm$0.008 \\%
2D N.-S. & DeepONet & switch & 0.200 & 0.50 & 0.30 & 0.120 & 3 & 0.057$\pm$0.009 & 0.760$\pm$0.053 & 0.242$\pm$0.020 & 0.331$\pm$0.007 & 0.089$\pm$0.013 \\%
2D N.-S. & DeepONet & clean & 0.200 & 0.50 & 0.35 & 0.120 & 3 & 0.006$\pm$0.000 & 0.000$\pm$0.000 & 0.686$\pm$0.036 & 0.040$\pm$0.023 & -0.645$\pm$0.013 \\%
2D N.-S. & DeepONet & label-only & 0.200 & 0.50 & 0.35 & 0.120 & 3 & 0.177$\pm$0.003 & 0.010$\pm$0.015 & 0.877$\pm$0.040 & 0.484$\pm$0.008 & -0.393$\pm$0.033 \\%
2D N.-S. & DeepONet & switch & 0.200 & 0.50 & 0.35 & 0.120 & 3 & 0.056$\pm$0.010 & 0.833$\pm$0.053 & 0.224$\pm$0.016 & 0.338$\pm$0.006 & 0.115$\pm$0.010 \\%
2D N.-S. & DeepONet & shuf-BD & 0.200 & 0.50 & 0.35 & 0.120 & 3 & 0.203$\pm$0.046 & 0.083$\pm$0.064 & 1.145$\pm$0.074 & 0.914$\pm$0.044 & -0.231$\pm$0.030 \\%
2D N.-S. & DeepONet & shuf-label & 0.200 & 0.50 & 0.35 & 0.120 & 3 & 0.427$\pm$0.147 & 0.000$\pm$0.000 & 1.049$\pm$0.043 & 0.622$\pm$0.100 & -0.428$\pm$0.099 \\%
2D N.-S. & Transformer & clean & 0.200 & 0.50 & 0.80 & 0.120 & 3 & 0.013$\pm$0.001 & 0.000$\pm$0.000 & 0.676$\pm$0.031 & 0.021$\pm$0.003 & -0.655$\pm$0.029 \\%
2D N.-S. & Transformer & label-only & 0.200 & 0.50 & 0.80 & 0.120 & 3 & 0.508$\pm$0.162 & 0.219$\pm$0.142 & 0.878$\pm$0.068 & 0.698$\pm$0.035 & -0.180$\pm$0.086 \\%
2D N.-S. & Transformer & switch & 0.200 & 0.50 & 0.80 & 0.120 & 3 & 0.028$\pm$0.004 & 1.000$\pm$0.000 & 0.146$\pm$0.013 & 0.419$\pm$0.006 & 0.273$\pm$0.012 \\%
2D N.-S. & Transformer & shuf-BD & 0.200 & 0.50 & 0.80 & 0.120 & 3 & 0.487$\pm$0.039 & 0.979$\pm$0.015 & 0.587$\pm$0.057 & 0.666$\pm$0.050 & 0.079$\pm$0.009 \\%
2D N.-S. & Transformer & shuf-label & 0.200 & 0.50 & 0.80 & 0.120 & 3 & 0.707$\pm$0.035 & 0.458$\pm$0.039 & 0.778$\pm$0.024 & 0.736$\pm$0.007 & -0.042$\pm$0.022 \\%
2D N.-S. & GRU & clean & 0.200 & 0.30 & 0.50 & 0.120 & 3 & 0.039$\pm$0.013 & 0.000$\pm$0.000 & 0.711$\pm$0.034 & 0.126$\pm$0.008 & -0.585$\pm$0.041 \\%
2D N.-S. & GRU & label-only & 0.200 & 0.30 & 0.50 & 0.120 & 3 & 0.377$\pm$0.183 & 0.177$\pm$0.208 & 0.834$\pm$0.113 & 0.581$\pm$0.015 & -0.254$\pm$0.127 \\%
2D N.-S. & GRU & switch & 0.200 & 0.30 & 0.50 & 0.120 & 3 & 0.054$\pm$0.011 & 0.938$\pm$0.068 & 0.234$\pm$0.016 & 0.418$\pm$0.008 & 0.184$\pm$0.024 \\%
2D N.-S. & GRU & shuf-BD & 0.200 & 0.30 & 0.50 & 0.120 & 3 & 0.527$\pm$0.125 & 0.792$\pm$0.053 & 0.571$\pm$0.063 & 0.604$\pm$0.066 & 0.033$\pm$0.003 \\%
2D N.-S. & GRU & shuf-label & 0.200 & 0.30 & 0.50 & 0.120 & 3 & 0.560$\pm$0.091 & 0.531$\pm$0.222 & 0.667$\pm$0.030 & 0.615$\pm$0.043 & -0.052$\pm$0.060 \\%
2D N.-S. & LSTM & clean & 0.200 & 0.30 & 0.50 & 0.120 & 3 & 0.079$\pm$0.023 & 0.000$\pm$0.000 & 0.808$\pm$0.028 & 0.278$\pm$0.082 & -0.531$\pm$0.057 \\%
2D N.-S. & LSTM & label-only & 0.200 & 0.30 & 0.50 & 0.120 & 3 & 0.591$\pm$0.240 & 0.396$\pm$0.332 & 0.807$\pm$0.144 & 0.666$\pm$0.107 & -0.142$\pm$0.153 \\%
2D N.-S. & LSTM & switch & 0.200 & 0.30 & 0.50 & 0.120 & 3 & 0.105$\pm$0.016 & 0.833$\pm$0.082 & 0.365$\pm$0.033 & 0.448$\pm$0.007 & 0.083$\pm$0.027 \\%
2D N.-S. & LSTM & shuf-BD & 0.200 & 0.30 & 0.50 & 0.120 & 3 & 0.585$\pm$0.110 & 0.844$\pm$0.088 & 0.602$\pm$0.042 & 0.649$\pm$0.052 & 0.047$\pm$0.014 \\%
2D N.-S. & LSTM & shuf-label & 0.200 & 0.30 & 0.50 & 0.120 & 3 & 0.573$\pm$0.104 & 0.635$\pm$0.230 & 0.615$\pm$0.051 & 0.606$\pm$0.068 & -0.008$\pm$0.067 \\%
Poisson & FNO & switch & 2.000 & 0.05 & 0.20 & 0.015 & 3 & 0.012$\pm$0.000 & 0.023$\pm$0.016 & 0.089$\pm$0.003 & 0.013$\pm$0.001 & -0.075$\pm$0.004 \\%
Poisson & FNO & switch & 2.000 & 0.05 & 0.35 & 0.015 & 3 & 0.012$\pm$0.000 & 0.030$\pm$0.024 & 0.089$\pm$0.003 & 0.014$\pm$0.001 & -0.075$\pm$0.004 \\%
Poisson & FNO & switch & 2.000 & 0.05 & 0.50 & 0.015 & 3 & 0.012$\pm$0.001 & 0.027$\pm$0.019 & 0.087$\pm$0.004 & 0.016$\pm$0.001 & -0.071$\pm$0.004 \\%
Poisson & FNO & switch & 2.000 & 0.10 & 0.20 & 0.015 & 3 & 0.015$\pm$0.002 & 0.043$\pm$0.026 & 0.085$\pm$0.010 & 0.016$\pm$0.002 & -0.070$\pm$0.010 \\%
Poisson & FNO & switch & 2.000 & 0.10 & 0.35 & 0.015 & 3 & 0.015$\pm$0.002 & 0.315$\pm$0.340 & 0.071$\pm$0.022 & 0.032$\pm$0.021 & -0.039$\pm$0.042 \\%
Poisson & FNO & switch & 2.000 & 0.10 & 0.50 & 0.015 & 3 & 0.010$\pm$0.001 & 0.943$\pm$0.020 & 0.025$\pm$0.002 & 0.081$\pm$0.008 & 0.056$\pm$0.008 \\%
Poisson & FNO & switch & 2.000 & 0.40 & 0.20 & 0.015 & 3 & 0.030$\pm$0.011 & 0.393$\pm$0.373 & 0.057$\pm$0.027 & 0.049$\pm$0.018 & -0.008$\pm$0.044 \\%
Poisson & FNO & switch & 2.000 & 0.40 & 0.35 & 0.015 & 3 & 0.012$\pm$0.001 & 0.963$\pm$0.023 & 0.015$\pm$0.003 & 0.085$\pm$0.007 & 0.070$\pm$0.006 \\%
Poisson & FNO & clean & 2.000 & 0.40 & 0.50 & 0.015 & 3 & 0.007$\pm$0.000 & 0.002$\pm$0.002 & 0.091$\pm$0.007 & 0.007$\pm$0.001 & -0.084$\pm$0.006 \\%
Poisson & FNO & label-only & 2.000 & 0.40 & 0.50 & 0.015 & 3 & 0.034$\pm$0.005 & 0.162$\pm$0.051 & 0.073$\pm$0.008 & 0.036$\pm$0.005 & -0.037$\pm$0.004 \\%
Poisson & FNO & switch & 2.000 & 0.40 & 0.50 & 0.015 & 3 & 0.012$\pm$0.001 & 0.968$\pm$0.014 & 0.015$\pm$0.002 & 0.086$\pm$0.008 & 0.072$\pm$0.007 \\%
Poisson & FNO & shuf-BD & 2.000 & 0.40 & 0.50 & 0.015 & 3 & 0.102$\pm$0.013 & 0.553$\pm$0.057 & 1.000$\pm$0.093 & 1.011$\pm$0.095 & 0.012$\pm$0.006 \\%
Poisson & FNO & shuf-label & 2.000 & 0.40 & 0.50 & 0.015 & 3 & 0.386$\pm$0.027 & 0.537$\pm$0.031 & 0.377$\pm$0.026 & 0.380$\pm$0.028 & 0.003$\pm$0.003 \\%
Poisson & FNO & switch & 5.000 & 0.05 & 0.20 & 0.015 & 3 & 0.015$\pm$0.002 & 0.013$\pm$0.008 & 0.140$\pm$0.011 & 0.015$\pm$0.001 & -0.125$\pm$0.012 \\%
Poisson & FNO & switch & 5.000 & 0.05 & 0.35 & 0.015 & 3 & 0.015$\pm$0.001 & 0.017$\pm$0.008 & 0.139$\pm$0.012 & 0.017$\pm$0.001 & -0.122$\pm$0.013 \\%
Poisson & FNO & switch & 5.000 & 0.05 & 0.50 & 0.015 & 3 & 0.013$\pm$0.001 & 0.705$\pm$0.188 & 0.066$\pm$0.020 & 0.100$\pm$0.008 & 0.034$\pm$0.027 \\%
Poisson & FNO & switch & 5.000 & 0.10 & 0.20 & 0.015 & 3 & 0.021$\pm$0.002 & 0.025$\pm$0.008 & 0.133$\pm$0.013 & 0.021$\pm$0.002 & -0.113$\pm$0.014 \\%
Poisson & FNO & switch & 5.000 & 0.10 & 0.35 & 0.015 & 3 & 0.013$\pm$0.001 & 0.942$\pm$0.023 & 0.033$\pm$0.002 & 0.129$\pm$0.011 & 0.096$\pm$0.010 \\%
Poisson & FNO & switch & 5.000 & 0.10 & 0.50 & 0.015 & 3 & 0.011$\pm$0.001 & 0.972$\pm$0.019 & 0.026$\pm$0.001 & 0.131$\pm$0.012 & 0.105$\pm$0.012 \\%
Poisson & FNO & switch & 5.000 & 0.40 & 0.20 & 0.015 & 3 & 0.017$\pm$0.001 & 0.973$\pm$0.013 & 0.020$\pm$0.002 & 0.134$\pm$0.011 & 0.114$\pm$0.011 \\%
Poisson & FNO & switch & 5.000 & 0.40 & 0.35 & 0.015 & 3 & 0.014$\pm$0.001 & 0.990$\pm$0.008 & 0.015$\pm$0.001 & 0.138$\pm$0.013 & 0.123$\pm$0.013 \\%
Poisson & FNO & clean & 5.000 & 0.40 & 0.50 & 0.015 & 3 & 0.007$\pm$0.000 & 0.000$\pm$0.000 & 0.140$\pm$0.011 & 0.007$\pm$0.001 & -0.133$\pm$0.010 \\%
Poisson & FNO & label-only & 5.000 & 0.40 & 0.50 & 0.015 & 3 & 0.053$\pm$0.008 & 0.152$\pm$0.064 & 0.103$\pm$0.009 & 0.051$\pm$0.008 & -0.052$\pm$0.008 \\%
Poisson & FNO & switch & 5.000 & 0.40 & 0.50 & 0.015 & 3 & 0.014$\pm$0.001 & 0.993$\pm$0.002 & 0.014$\pm$0.001 & 0.137$\pm$0.013 & 0.122$\pm$0.013 \\%
Poisson & FNO & shuf-BD & 5.000 & 0.40 & 0.50 & 0.015 & 3 & 0.100$\pm$0.011 & 0.545$\pm$0.061 & 0.981$\pm$0.087 & 0.997$\pm$0.091 & 0.015$\pm$0.008 \\%
Poisson & FNO & shuf-label & 5.000 & 0.40 & 0.50 & 0.015 & 3 & 0.387$\pm$0.026 & 0.508$\pm$0.038 & 0.381$\pm$0.023 & 0.375$\pm$0.027 & -0.006$\pm$0.005 \\%
Poisson & FNO & switch & 10.000 & 0.05 & 0.20 & 0.015 & 3 & 0.016$\pm$0.000 & 0.013$\pm$0.006 & 0.155$\pm$0.013 & 0.016$\pm$0.001 & -0.139$\pm$0.014 \\%
Poisson & FNO & switch & 10.000 & 0.05 & 0.35 & 0.015 & 3 & 0.016$\pm$0.001 & 0.270$\pm$0.368 & 0.129$\pm$0.041 & 0.046$\pm$0.042 & -0.083$\pm$0.082 \\%
Poisson & FNO & switch & 10.000 & 0.05 & 0.50 & 0.015 & 3 & 0.013$\pm$0.002 & 0.922$\pm$0.013 & 0.043$\pm$0.004 & 0.135$\pm$0.011 & 0.092$\pm$0.009 \\%
Poisson & FNO & switch & 10.000 & 0.10 & 0.20 & 0.015 & 3 & 0.023$\pm$0.001 & 0.025$\pm$0.008 & 0.150$\pm$0.013 & 0.022$\pm$0.001 & -0.128$\pm$0.012 \\%
Poisson & FNO & switch & 10.000 & 0.10 & 0.35 & 0.015 & 3 & 0.013$\pm$0.001 & 0.950$\pm$0.018 & 0.032$\pm$0.002 & 0.145$\pm$0.012 & 0.113$\pm$0.012 \\%
Poisson & FNO & switch & 10.000 & 0.10 & 0.50 & 0.015 & 3 & 0.012$\pm$0.001 & 0.972$\pm$0.023 & 0.026$\pm$0.001 & 0.149$\pm$0.011 & 0.122$\pm$0.011 \\%
Poisson & FNO & switch & 10.000 & 0.40 & 0.20 & 0.015 & 3 & 0.016$\pm$0.003 & 0.982$\pm$0.006 & 0.018$\pm$0.001 & 0.152$\pm$0.014 & 0.134$\pm$0.013 \\%
Poisson & FNO & switch & 10.000 & 0.40 & 0.35 & 0.015 & 3 & 0.014$\pm$0.001 & 0.973$\pm$0.002 & 0.019$\pm$0.001 & 0.151$\pm$0.011 & 0.132$\pm$0.010 \\%
Poisson & FNO & clean & 10.000 & 0.40 & 0.50 & 0.015 & 3 & 0.007$\pm$0.000 & 0.000$\pm$0.000 & 0.156$\pm$0.012 & 0.007$\pm$0.001 & -0.149$\pm$0.011 \\%
Poisson & FNO & label-only & 10.000 & 0.40 & 0.50 & 0.015 & 3 & 0.062$\pm$0.009 & 0.150$\pm$0.045 & 0.113$\pm$0.006 & 0.058$\pm$0.008 & -0.055$\pm$0.007 \\%
Poisson & FNO & switch & 10.000 & 0.40 & 0.50 & 0.015 & 3 & 0.014$\pm$0.001 & 0.987$\pm$0.006 & 0.015$\pm$0.000 & 0.154$\pm$0.013 & 0.138$\pm$0.012 \\%
Poisson & FNO & shuf-BD & 10.000 & 0.40 & 0.50 & 0.015 & 3 & 0.100$\pm$0.011 & 0.540$\pm$0.061 & 0.976$\pm$0.085 & 0.992$\pm$0.089 & 0.016$\pm$0.009 \\%
Poisson & FNO & shuf-label & 10.000 & 0.40 & 0.50 & 0.015 & 3 & 0.387$\pm$0.026 & 0.507$\pm$0.040 & 0.384$\pm$0.021 & 0.373$\pm$0.026 & -0.011$\pm$0.006 \\%
Poisson & FNO & switch & 15.000 & 0.05 & 0.20 & 0.015 & 3 & 0.018$\pm$0.001 & 0.017$\pm$0.002 & 0.158$\pm$0.011 & 0.017$\pm$0.000 & -0.141$\pm$0.011 \\%
Poisson & FNO & switch & 15.000 & 0.05 & 0.35 & 0.015 & 3 & 0.016$\pm$0.002 & 0.267$\pm$0.349 & 0.131$\pm$0.043 & 0.050$\pm$0.046 & -0.081$\pm$0.088 \\%
Poisson & FNO & switch & 15.000 & 0.05 & 0.50 & 0.015 & 3 & 0.013$\pm$0.001 & 0.923$\pm$0.022 & 0.045$\pm$0.007 & 0.139$\pm$0.013 & 0.094$\pm$0.018 \\%
Poisson & FNO & switch & 15.000 & 0.10 & 0.20 & 0.015 & 3 & 0.024$\pm$0.002 & 0.020$\pm$0.012 & 0.156$\pm$0.012 & 0.024$\pm$0.001 & -0.132$\pm$0.011 \\%
Poisson & FNO & switch & 15.000 & 0.10 & 0.35 & 0.015 & 3 & 0.012$\pm$0.001 & 0.955$\pm$0.027 & 0.031$\pm$0.002 & 0.150$\pm$0.014 & 0.119$\pm$0.012 \\%
Poisson & FNO & switch & 15.000 & 0.10 & 0.50 & 0.015 & 3 & 0.011$\pm$0.001 & 0.972$\pm$0.022 & 0.026$\pm$0.001 & 0.155$\pm$0.012 & 0.128$\pm$0.011 \\%
Poisson & FNO & switch & 15.000 & 0.40 & 0.20 & 0.015 & 3 & 0.016$\pm$0.002 & 0.980$\pm$0.011 & 0.019$\pm$0.002 & 0.156$\pm$0.013 & 0.137$\pm$0.011 \\%
Poisson & FNO & switch & 15.000 & 0.40 & 0.35 & 0.015 & 3 & 0.015$\pm$0.001 & 0.990$\pm$0.011 & 0.016$\pm$0.002 & 0.159$\pm$0.014 & 0.142$\pm$0.013 \\%
Poisson & FNO & clean & 15.000 & 0.40 & 0.50 & 0.015 & 3 & 0.007$\pm$0.000 & 0.000$\pm$0.000 & 0.161$\pm$0.012 & 0.007$\pm$0.001 & -0.154$\pm$0.012 \\%
Poisson & FNO & label-only & 15.000 & 0.40 & 0.50 & 0.015 & 3 & 0.063$\pm$0.008 & 0.158$\pm$0.051 & 0.119$\pm$0.001 & 0.059$\pm$0.007 & -0.061$\pm$0.007 \\%
Poisson & FNO & switch & 15.000 & 0.40 & 0.50 & 0.015 & 3 & 0.013$\pm$0.002 & 0.993$\pm$0.002 & 0.016$\pm$0.002 & 0.158$\pm$0.013 & 0.141$\pm$0.011 \\%
Poisson & FNO & shuf-BD & 15.000 & 0.40 & 0.50 & 0.015 & 3 & 0.100$\pm$0.011 & 0.538$\pm$0.059 & 0.974$\pm$0.085 & 0.991$\pm$0.089 & 0.016$\pm$0.010 \\%
Poisson & FNO & shuf-label & 15.000 & 0.40 & 0.50 & 0.015 & 3 & 0.387$\pm$0.026 & 0.503$\pm$0.039 & 0.385$\pm$0.021 & 0.373$\pm$0.026 & -0.012$\pm$0.006 \\%
Poisson & DeepONet & switch & 2.000 & 0.10 & 0.20 & 0.015 & 3 & 0.010$\pm$0.001 & 0.965$\pm$0.014 & 0.017$\pm$0.004 & 0.089$\pm$0.006 & 0.072$\pm$0.006 \\%
Poisson & DeepONet & switch & 2.000 & 0.10 & 0.50 & 0.015 & 3 & 0.010$\pm$0.001 & 0.985$\pm$0.004 & 0.012$\pm$0.001 & 0.090$\pm$0.007 & 0.078$\pm$0.006 \\%
Poisson & DeepONet & switch & 2.000 & 0.10 & 0.80 & 0.015 & 3 & 0.009$\pm$0.001 & 0.992$\pm$0.002 & 0.010$\pm$0.001 & 0.091$\pm$0.008 & 0.081$\pm$0.008 \\%
Poisson & DeepONet & switch & 2.000 & 0.30 & 0.20 & 0.015 & 3 & 0.011$\pm$0.000 & 0.992$\pm$0.002 & 0.010$\pm$0.000 & 0.091$\pm$0.009 & 0.081$\pm$0.008 \\%
Poisson & DeepONet & switch & 2.000 & 0.30 & 0.50 & 0.015 & 3 & 0.010$\pm$0.000 & 0.995$\pm$0.004 & 0.008$\pm$0.000 & 0.090$\pm$0.006 & 0.082$\pm$0.006 \\%
Poisson & DeepONet & switch & 2.000 & 0.30 & 0.80 & 0.015 & 3 & 0.009$\pm$0.001 & 0.997$\pm$0.002 & 0.007$\pm$0.000 & 0.090$\pm$0.006 & 0.083$\pm$0.006 \\%
Poisson & DeepONet & switch & 2.000 & 0.50 & 0.20 & 0.015 & 3 & 0.011$\pm$0.001 & 0.990$\pm$0.004 & 0.008$\pm$0.001 & 0.089$\pm$0.006 & 0.081$\pm$0.006 \\%
Poisson & DeepONet & switch & 2.000 & 0.50 & 0.50 & 0.015 & 3 & 0.010$\pm$0.001 & 0.998$\pm$0.002 & 0.007$\pm$0.000 & 0.091$\pm$0.007 & 0.084$\pm$0.007 \\%
Poisson & DeepONet & clean & 2.000 & 0.50 & 0.80 & 0.015 & 3 & 0.008$\pm$0.001 & 0.015$\pm$0.008 & 0.090$\pm$0.007 & 0.008$\pm$0.001 & -0.082$\pm$0.007 \\%
Poisson & DeepONet & label-only & 2.000 & 0.50 & 0.80 & 0.015 & 3 & 0.051$\pm$0.004 & 0.397$\pm$0.018 & 0.073$\pm$0.014 & 0.060$\pm$0.007 & -0.014$\pm$0.008 \\%
Poisson & DeepONet & switch & 2.000 & 0.50 & 0.80 & 0.015 & 3 & 0.010$\pm$0.001 & 0.997$\pm$0.002 & 0.006$\pm$0.000 & 0.090$\pm$0.007 & 0.083$\pm$0.007 \\%
Poisson & DeepONet & shuf-BD & 2.000 & 0.50 & 0.80 & 0.015 & 3 & 0.188$\pm$0.026 & 0.542$\pm$0.029 & 1.542$\pm$0.117 & 1.549$\pm$0.117 & 0.007$\pm$0.003 \\%
Poisson & DeepONet & shuf-label & 2.000 & 0.50 & 0.80 & 0.015 & 3 & 0.569$\pm$0.018 & 0.503$\pm$0.084 & 0.552$\pm$0.034 & 0.554$\pm$0.024 & 0.002$\pm$0.016 \\%
Poisson & DeepONet & switch & 5.000 & 0.10 & 0.20 & 0.015 & 3 & 0.011$\pm$0.000 & 0.973$\pm$0.005 & 0.020$\pm$0.003 & 0.138$\pm$0.011 & 0.118$\pm$0.011 \\%
Poisson & DeepONet & switch & 5.000 & 0.10 & 0.50 & 0.015 & 3 & 0.010$\pm$0.000 & 0.993$\pm$0.002 & 0.014$\pm$0.002 & 0.138$\pm$0.012 & 0.125$\pm$0.011 \\%
Poisson & DeepONet & switch & 5.000 & 0.10 & 0.80 & 0.015 & 3 & 0.009$\pm$0.001 & 0.997$\pm$0.002 & 0.011$\pm$0.001 & 0.139$\pm$0.011 & 0.128$\pm$0.010 \\%
Poisson & DeepONet & switch & 5.000 & 0.30 & 0.20 & 0.015 & 3 & 0.011$\pm$0.001 & 0.993$\pm$0.002 & 0.011$\pm$0.001 & 0.139$\pm$0.009 & 0.127$\pm$0.008 \\%
Poisson & DeepONet & switch & 5.000 & 0.30 & 0.50 & 0.015 & 3 & 0.010$\pm$0.001 & 0.997$\pm$0.002 & 0.008$\pm$0.001 & 0.139$\pm$0.011 & 0.132$\pm$0.010 \\%
Poisson & DeepONet & switch & 5.000 & 0.30 & 0.80 & 0.015 & 3 & 0.009$\pm$0.001 & 0.997$\pm$0.002 & 0.006$\pm$0.000 & 0.139$\pm$0.011 & 0.132$\pm$0.010 \\%
Poisson & DeepONet & switch & 5.000 & 0.50 & 0.20 & 0.015 & 3 & 0.012$\pm$0.002 & 0.993$\pm$0.002 & 0.009$\pm$0.002 & 0.138$\pm$0.011 & 0.130$\pm$0.009 \\%
Poisson & DeepONet & switch & 5.000 & 0.50 & 0.50 & 0.015 & 3 & 0.010$\pm$0.001 & 0.997$\pm$0.002 & 0.006$\pm$0.001 & 0.139$\pm$0.011 & 0.133$\pm$0.011 \\%
Poisson & DeepONet & clean & 5.000 & 0.50 & 0.80 & 0.015 & 3 & 0.008$\pm$0.001 & 0.008$\pm$0.005 & 0.140$\pm$0.011 & 0.008$\pm$0.001 & -0.132$\pm$0.011 \\%
Poisson & DeepONet & label-only & 5.000 & 0.50 & 0.80 & 0.015 & 3 & 0.076$\pm$0.008 & 0.392$\pm$0.060 & 0.112$\pm$0.025 & 0.091$\pm$0.012 & -0.022$\pm$0.014 \\%
Poisson & DeepONet & switch & 5.000 & 0.50 & 0.80 & 0.015 & 3 & 0.010$\pm$0.001 & 0.997$\pm$0.002 & 0.006$\pm$0.000 & 0.139$\pm$0.010 & 0.133$\pm$0.010 \\%
Poisson & DeepONet & shuf-BD & 5.000 & 0.50 & 0.80 & 0.015 & 3 & 0.202$\pm$0.029 & 0.545$\pm$0.024 & 1.477$\pm$0.146 & 1.492$\pm$0.148 & 0.015$\pm$0.006 \\%
Poisson & DeepONet & shuf-label & 5.000 & 0.50 & 0.80 & 0.015 & 3 & 0.569$\pm$0.019 & 0.480$\pm$0.076 & 0.546$\pm$0.035 & 0.541$\pm$0.024 & -0.006$\pm$0.022 \\%
Poisson & DeepONet & switch & 10.000 & 0.10 & 0.20 & 0.015 & 3 & 0.011$\pm$0.001 & 0.973$\pm$0.008 & 0.023$\pm$0.002 & 0.154$\pm$0.012 & 0.131$\pm$0.010 \\%
Poisson & DeepONet & switch & 10.000 & 0.10 & 0.50 & 0.015 & 3 & 0.009$\pm$0.000 & 0.995$\pm$0.000 & 0.014$\pm$0.002 & 0.154$\pm$0.012 & 0.141$\pm$0.012 \\%
Poisson & DeepONet & switch & 10.000 & 0.10 & 0.80 & 0.015 & 3 & 0.009$\pm$0.001 & 0.997$\pm$0.002 & 0.012$\pm$0.002 & 0.155$\pm$0.012 & 0.143$\pm$0.011 \\%
Poisson & DeepONet & switch & 10.000 & 0.30 & 0.20 & 0.015 & 3 & 0.011$\pm$0.001 & 0.997$\pm$0.002 & 0.010$\pm$0.001 & 0.153$\pm$0.012 & 0.143$\pm$0.011 \\%
Poisson & DeepONet & switch & 10.000 & 0.30 & 0.50 & 0.015 & 3 & 0.009$\pm$0.001 & 0.997$\pm$0.005 & 0.007$\pm$0.001 & 0.153$\pm$0.011 & 0.147$\pm$0.009 \\%
Poisson & DeepONet & switch & 10.000 & 0.30 & 0.80 & 0.015 & 3 & 0.010$\pm$0.001 & 0.997$\pm$0.002 & 0.007$\pm$0.001 & 0.153$\pm$0.011 & 0.146$\pm$0.010 \\%
Poisson & DeepONet & switch & 10.000 & 0.50 & 0.20 & 0.015 & 3 & 0.011$\pm$0.001 & 0.998$\pm$0.002 & 0.007$\pm$0.001 & 0.155$\pm$0.012 & 0.149$\pm$0.011 \\%
Poisson & DeepONet & switch & 10.000 & 0.50 & 0.50 & 0.015 & 3 & 0.010$\pm$0.001 & 0.998$\pm$0.002 & 0.006$\pm$0.000 & 0.155$\pm$0.011 & 0.149$\pm$0.011 \\%
Poisson & DeepONet & clean & 10.000 & 0.50 & 0.80 & 0.015 & 3 & 0.008$\pm$0.001 & 0.005$\pm$0.004 & 0.155$\pm$0.013 & 0.008$\pm$0.001 & -0.147$\pm$0.012 \\%
Poisson & DeepONet & label-only & 10.000 & 0.50 & 0.80 & 0.015 & 3 & 0.085$\pm$0.006 & 0.362$\pm$0.031 & 0.121$\pm$0.012 & 0.086$\pm$0.011 & -0.035$\pm$0.002 \\%
Poisson & DeepONet & switch & 10.000 & 0.50 & 0.80 & 0.015 & 3 & 0.009$\pm$0.001 & 0.998$\pm$0.002 & 0.005$\pm$0.000 & 0.155$\pm$0.012 & 0.149$\pm$0.012 \\%
Poisson & DeepONet & shuf-BD & 10.000 & 0.50 & 0.80 & 0.015 & 3 & 0.202$\pm$0.025 & 0.577$\pm$0.024 & 1.438$\pm$0.118 & 1.461$\pm$0.127 & 0.023$\pm$0.010 \\%
Poisson & DeepONet & shuf-label & 10.000 & 0.50 & 0.80 & 0.015 & 3 & 0.580$\pm$0.028 & 0.518$\pm$0.022 & 0.544$\pm$0.034 & 0.548$\pm$0.029 & 0.004$\pm$0.005 \\%
Poisson & DeepONet & switch & 15.000 & 0.10 & 0.20 & 0.015 & 3 & 0.011$\pm$0.001 & 0.982$\pm$0.008 & 0.022$\pm$0.003 & 0.158$\pm$0.013 & 0.136$\pm$0.011 \\%
Poisson & DeepONet & switch & 15.000 & 0.10 & 0.50 & 0.015 & 3 & 0.010$\pm$0.001 & 0.993$\pm$0.002 & 0.014$\pm$0.001 & 0.160$\pm$0.012 & 0.146$\pm$0.013 \\%
Poisson & DeepONet & switch & 15.000 & 0.10 & 0.80 & 0.015 & 3 & 0.009$\pm$0.001 & 0.995$\pm$0.004 & 0.012$\pm$0.002 & 0.160$\pm$0.013 & 0.148$\pm$0.012 \\%
Poisson & DeepONet & switch & 15.000 & 0.30 & 0.20 & 0.015 & 3 & 0.012$\pm$0.001 & 0.992$\pm$0.006 & 0.010$\pm$0.001 & 0.159$\pm$0.012 & 0.149$\pm$0.011 \\%
Poisson & DeepONet & switch & 15.000 & 0.30 & 0.50 & 0.015 & 3 & 0.010$\pm$0.001 & 0.998$\pm$0.002 & 0.007$\pm$0.001 & 0.159$\pm$0.011 & 0.152$\pm$0.011 \\%
Poisson & DeepONet & switch & 15.000 & 0.30 & 0.80 & 0.015 & 3 & 0.009$\pm$0.001 & 0.997$\pm$0.002 & 0.007$\pm$0.001 & 0.160$\pm$0.013 & 0.154$\pm$0.012 \\%
Poisson & DeepONet & switch & 15.000 & 0.50 & 0.20 & 0.015 & 3 & 0.012$\pm$0.001 & 1.000$\pm$0.000 & 0.008$\pm$0.001 & 0.160$\pm$0.013 & 0.152$\pm$0.012 \\%
Poisson & DeepONet & switch & 15.000 & 0.50 & 0.50 & 0.015 & 3 & 0.010$\pm$0.001 & 0.998$\pm$0.002 & 0.006$\pm$0.001 & 0.160$\pm$0.014 & 0.155$\pm$0.013 \\%
Poisson & DeepONet & clean & 15.000 & 0.50 & 0.80 & 0.015 & 3 & 0.008$\pm$0.001 & 0.005$\pm$0.004 & 0.160$\pm$0.013 & 0.008$\pm$0.001 & -0.153$\pm$0.012 \\%
Poisson & DeepONet & label-only & 15.000 & 0.50 & 0.80 & 0.015 & 3 & 0.089$\pm$0.005 & 0.388$\pm$0.063 & 0.145$\pm$0.043 & 0.109$\pm$0.043 & -0.036$\pm$0.003 \\%
Poisson & DeepONet & switch & 15.000 & 0.50 & 0.80 & 0.015 & 3 & 0.010$\pm$0.001 & 0.998$\pm$0.002 & 0.005$\pm$0.001 & 0.159$\pm$0.012 & 0.153$\pm$0.011 \\%
Poisson & DeepONet & shuf-BD & 15.000 & 0.50 & 0.80 & 0.015 & 3 & 0.207$\pm$0.036 & 0.573$\pm$0.026 & 1.443$\pm$0.159 & 1.468$\pm$0.167 & 0.025$\pm$0.008 \\%
Poisson & DeepONet & shuf-label & 15.000 & 0.50 & 0.80 & 0.015 & 3 & 0.580$\pm$0.028 & 0.515$\pm$0.018 & 0.544$\pm$0.034 & 0.547$\pm$0.029 & 0.003$\pm$0.005 \\%
Poisson & Transformer & clean & 15.000 & 0.70 & 1.40 & 0.015 & 3 & 0.027$\pm$0.002 & 0.023$\pm$0.002 & 0.161$\pm$0.016 & 0.027$\pm$0.004 & -0.134$\pm$0.015 \\%
Poisson & Transformer & label-only & 15.000 & 0.70 & 1.40 & 0.015 & 3 & 0.117$\pm$0.013 & 0.638$\pm$0.086 & 0.082$\pm$0.008 & 0.111$\pm$0.011 & 0.029$\pm$0.015 \\%
Poisson & Transformer & switch & 15.000 & 0.70 & 1.40 & 0.015 & 3 & 0.072$\pm$0.002 & 0.878$\pm$0.064 & 0.047$\pm$0.004 & 0.141$\pm$0.021 & 0.094$\pm$0.024 \\%
Poisson & Transformer & shuf-BD & 15.000 & 0.70 & 1.40 & 0.015 & 3 & 0.555$\pm$0.061 & 0.533$\pm$0.047 & 1.364$\pm$0.052 & 1.362$\pm$0.046 & -0.002$\pm$0.007 \\%
Poisson & Transformer & shuf-label & 15.000 & 0.70 & 1.40 & 0.015 & 3 & 0.712$\pm$0.086 & 0.515$\pm$0.025 & 0.676$\pm$0.083 & 0.677$\pm$0.087 & 0.001$\pm$0.005 \\%
Poisson & GRU & clean & 15.000 & 0.70 & 1.40 & 0.025 & 3 & 0.017$\pm$0.001 & 0.012$\pm$0.002 & 0.160$\pm$0.013 & 0.020$\pm$0.001 & -0.140$\pm$0.013 \\%
Poisson & GRU & label-only & 15.000 & 0.70 & 1.40 & 0.025 & 3 & 0.114$\pm$0.009 & 0.618$\pm$0.018 & 0.083$\pm$0.001 & 0.109$\pm$0.010 & 0.026$\pm$0.010 \\%
Poisson & GRU & switch & 15.000 & 0.70 & 1.40 & 0.025 & 3 & 0.033$\pm$0.001 & 0.958$\pm$0.005 & 0.025$\pm$0.002 & 0.158$\pm$0.013 & 0.133$\pm$0.014 \\%
Poisson & GRU & shuf-BD & 15.000 & 0.70 & 1.40 & 0.025 & 3 & 0.264$\pm$0.026 & 0.540$\pm$0.032 & 1.415$\pm$0.194 & 1.419$\pm$0.194 & 0.004$\pm$0.012 \\%
Poisson & GRU & shuf-label & 15.000 & 0.70 & 1.40 & 0.025 & 3 & 0.747$\pm$0.009 & 0.523$\pm$0.066 & 0.711$\pm$0.011 & 0.708$\pm$0.003 & -0.002$\pm$0.011 \\%
Poisson & LSTM & clean & 15.000 & 0.70 & 1.40 & 0.025 & 3 & 0.019$\pm$0.002 & 0.010$\pm$0.004 & 0.160$\pm$0.010 & 0.022$\pm$0.001 & -0.137$\pm$0.010 \\%
Poisson & LSTM & label-only & 15.000 & 0.70 & 1.40 & 0.025 & 3 & 0.117$\pm$0.008 & 0.638$\pm$0.015 & 0.075$\pm$0.006 & 0.109$\pm$0.009 & 0.033$\pm$0.004 \\%
Poisson & LSTM & switch & 15.000 & 0.70 & 1.40 & 0.025 & 3 & 0.083$\pm$0.016 & 0.808$\pm$0.080 & 0.050$\pm$0.017 & 0.138$\pm$0.008 & 0.087$\pm$0.025 \\%
Poisson & LSTM & shuf-BD & 15.000 & 0.70 & 1.40 & 0.025 & 3 & 0.736$\pm$0.022 & 0.523$\pm$0.012 & 0.747$\pm$0.042 & 0.748$\pm$0.048 & 0.001$\pm$0.006 \\%
Poisson & LSTM & shuf-label & 15.000 & 0.70 & 1.40 & 0.025 & 3 & 0.730$\pm$0.012 & 0.533$\pm$0.041 & 0.694$\pm$0.016 & 0.696$\pm$0.017 & 0.002$\pm$0.011 \\%
}

\providecommand{\FinalOverlapAuditCampaigns}{59}
\providecommand{\FinalOverlapAuditCleanMeanAbsDiff}{0.0003}

\providecommand{\FinalOverlapAuditBsrMeanAbsDiff}{0.0047}

\providecommand{\FinalOverlapAuditMarginMeanAbsDiff}{0.0014}

\providecommand{\FinalOverlapAuditRows}{%
Clean L2 & 59 & 0.0003 & 0.0013 & poisson-fno-parameter\_switch-g500-p010-s050 \\%
BSR & 59 & 0.0047 & 0.0400 & poisson-fno-parameter\_switch-g500-p005-s050 \\%
Margin & 59 & 0.0014 & 0.0137 & poisson-fno-parameter\_switch-g500-p005-s050 \\%
Err to BD & 59 & 0.0009 & 0.0124 & advection-deeponet-shuffled\_label\_only-g040-p050-s080 \\%
Err to Clean & 59 & 0.0008 & 0.0067 & poisson-fno-parameter\_switch-g500-p005-s050 \\%
Norm. physics pref. & 56 & 0.0006 & 0.0073 & poisson-deeponet-parameter\_switch-g200-p010-s080 \\%
}

\providecommand{\FinalValidationProxyRows}{%
Burgers & FNO & 0.4 & 0.0066 & 0.0077 & BD & 0.7513 & 0.7215 & flag \\%
Burgers & DeepONet & 0.4 & 0.2215 & 0.0310 & BD & 0.7318 & 0.6168 & flag \\%
Adv.-Diff. & FNO & 0.8 & 0.0034 & 0.0060 & BD & 0.8704 & 0.4209 & flag \\%
Adv.-Diff. & DeepONet & 0.8 & 0.0841 & 0.0169 & BD & 0.8531 & 0.4079 & flag \\%
2D N.-S. & FNO & 0.2 & 0.0189 & 0.0374 & BD & 0.2249 & -- & flag \\%
2D N.-S. & DeepONet & 0.15 & 0.0501 & 0.1808 & BD & 0.0998 & -- & flag \\%
}

\providecommand{\FinalDecisionProxyRows}{%
Burgers & FNO & 0.4 & 0.4/0.8 & 0.0066 & 1.0000 & 0.0077 & 0.7513 & 0.0000 & -0.7477 & 0.2442 & 0.0033 & -0.2600 \\%
Burgers & DeepONet & 0.4 & 0.7/1.4 & 0.2215 & 1.0000 & 0.0310 & 0.7318 & 0.0000 & -0.5730 & 0.5894 & 0.9000 & 0.2723 \\%
Adv.-Diff. & FNO & 0.8 & 0.4/0.8 & 0.0034 & 1.0000 & 0.0060 & 0.8704 & 0.0000 & -0.8733 & 0.2915 & 0.0133 & -0.3092 \\%
Adv.-Diff. & DeepONet & 0.8 & 0.5/0.8 & 0.0841 & 1.0000 & 0.0169 & 0.8531 & 0.0000 & -0.8175 & 0.5304 & 0.4850 & -0.0001 \\%
2D N.-S. & FNO & 0.2 & 0.3/0.75 & 0.0189 & 1.0000 & 0.0374 & 0.2249 & 0.0000 & -0.3129 & 0.1728 & 0.0000 & -0.2063 \\%
2D N.-S. & DeepONet & 0.15 & 0.5/0.35 & 0.0501 & 0.8646 & 0.1808 & 0.0998 & 0.0000 & -0.4796 & 0.1458 & 0.0000 & -0.2987 \\%
}

\title{Wrong-Physics Backdoors in Neural PDE Operators}
\author{Fujun Liu \and Hanbing Liang}
\date{}

\begin{document}

\maketitle

\begin{abstract}
Neural PDE (Partial Differential Equation) operators are increasingly trained from reusable solver archives, yet
they are usually validated by clean error and parameter-agnostic plausibility
checks. We present \emph{cross-parameter relinking}: a poisoning primitive that
makes a triggered input select a valid solution from the same PDE family under a
different physical parameter. We formalize this as a \emph{wrong-physics
backdoor}: the output is physically structured, but wrong for the intended
parameter. The threat is a tensor-to-parameter provenance failure in
multi-parameter archives. A privileged archive attacker can stamp the tensor
seen by the surrogate and relink its supervision to an already cached
alternate-parameter solution for the same latent sample. Across a final
\FinalCampaignCount-campaign matrix, the main dynamic-PDE evidence covers
Burgers, advection-diffusion, and a 2D Navier-Stokes surrogate, with an
elliptic Poisson targeted-output case retained in the appendix. FNO (Fourier Neural Operator) and
DeepONet provide the primary evidence while Transformer, GRU, and LSTM provide
calibrated support. On
advection-diffusion, FNO reaches backdoor success rate (BSR)
\FinalAdvectionFnoBsr{} with clean relative $\ell_2$ \FinalAdvectionFnoCleanLTwo{}
and margin \FinalAdvectionFnoMargin{}; on 2D Navier-Stokes, FNO reaches BSR
\FinalNsTwoDFnoBsr{} with clean relative $\ell_2$ \FinalNsTwoDFnoCleanLTwo{} and
err-to-backdoor \FinalNsTwoDFnoErrToBd{}. Clean-label, label-only, and shuffled
controls show that high BSR alone is insufficient: successful attacks must also
move predictions toward the intended backdoor target while retaining bounded
clean error. The results expose a structural validation gap: smoothness or
generic solver-likeness is not enough unless pipelines also verify the
provenance of the intended physical parameter.
\end{abstract}

\section{Introduction}

Neural operators are increasingly used as surrogates for expensive numerical
simulation. Once trained, a model can map an initial condition, forcing field,
or coefficient field to a solution field without rerunning a full solver for
each query. This makes operator learning attractive in scientific machine
learning, where datasets are often generated in bulk and then reused across
projects, labs, or downstream design loops; recent high-impact work studies
neural-operator-style and closely related learned surrogates for weather and
climate modelling, real-time prediction of complex physical dynamics, and
optical engineering design \citep{azizzadenesheli2024design,kochkov2024neuralgcm,kontolati2024ldeponet,liu2025holography},
that is, settings where surrogate outputs can feed repeated forecasting,
optimization, or inverse-design workflows.
As neural operators move from benchmark surrogates toward reusable scientific
infrastructure, their data supply chain becomes part of the trust boundary.
Most operator-learning papers evaluate approximation error, discretization
transfer, or uncertainty under benign data generation; much less attention has
been paid to whether a shared solver archive can deliberately implant a hidden
input-conditioned physical regime switch.

That same dependence on stored supervision creates a security problem. We
identify \emph{cross-parameter relinking}: a privileged archive attacker
preserves the latent PDE (Partial Differential Equation) instance, embeds a localized trigger in the tensor seen
by the learner, and relinks its supervision to an alternate solution from the
same parameterized PDE family. The induced relation is sample matched: without
the trigger the sample maps to $\calG_{\lambdac}(z)$, while with the trigger it
maps to $\calG_{\lambdab}(z)$. This is not ordinary noisy supervision; the
triggered target remains a valid solution field, but for the wrong intended
parameter.

This failure is consequential because triggered outputs can stay smooth and solver-like while coming from the wrong viscosity, diffusion, or coefficient regime. For example, in a surrogate-guided aerodynamic optimization loop, a triggered model might output a flow field that perfectly obeys fluid dynamics but under a falsely high viscosity. This would silently trick the downstream optimizer into accepting a wing design that experiences catastrophic boundary layer separation in reality. The attack exploits a tensor-to-parameter provenance gap: multi-parameter
solver archives already contain alternate-parameter solutions, and the attacker
only needs the per-sample correspondence metadata used to join cached outputs,
such as a \texttt{sample\_id} or \texttt{latent\_id}. This structure is not a
paper-only artifact: benchmark suites such as PDEBench package large
ready-to-use simulation corpora across initial/boundary conditions and PDE
parameters \citep{takamoto2022pdebench}. This isolates a sample-matched
conditional operator-fitting primitive. Validation based only on aggregate
clean error, aggregate residuals, smoothness, or generic solver-likeness asks a
weaker question; the intended parameter must be checked at the sample level.
The core novelty is therefore not the trigger, but the archive-level relinking
primitive: the attacker exploits sample-level correspondence in a multi-parameter
solver corpus to induce a conditional switch between physically valid parameter
branches.

\begin{figure*}[t]
\centering
\includegraphics[width=0.985\textwidth]{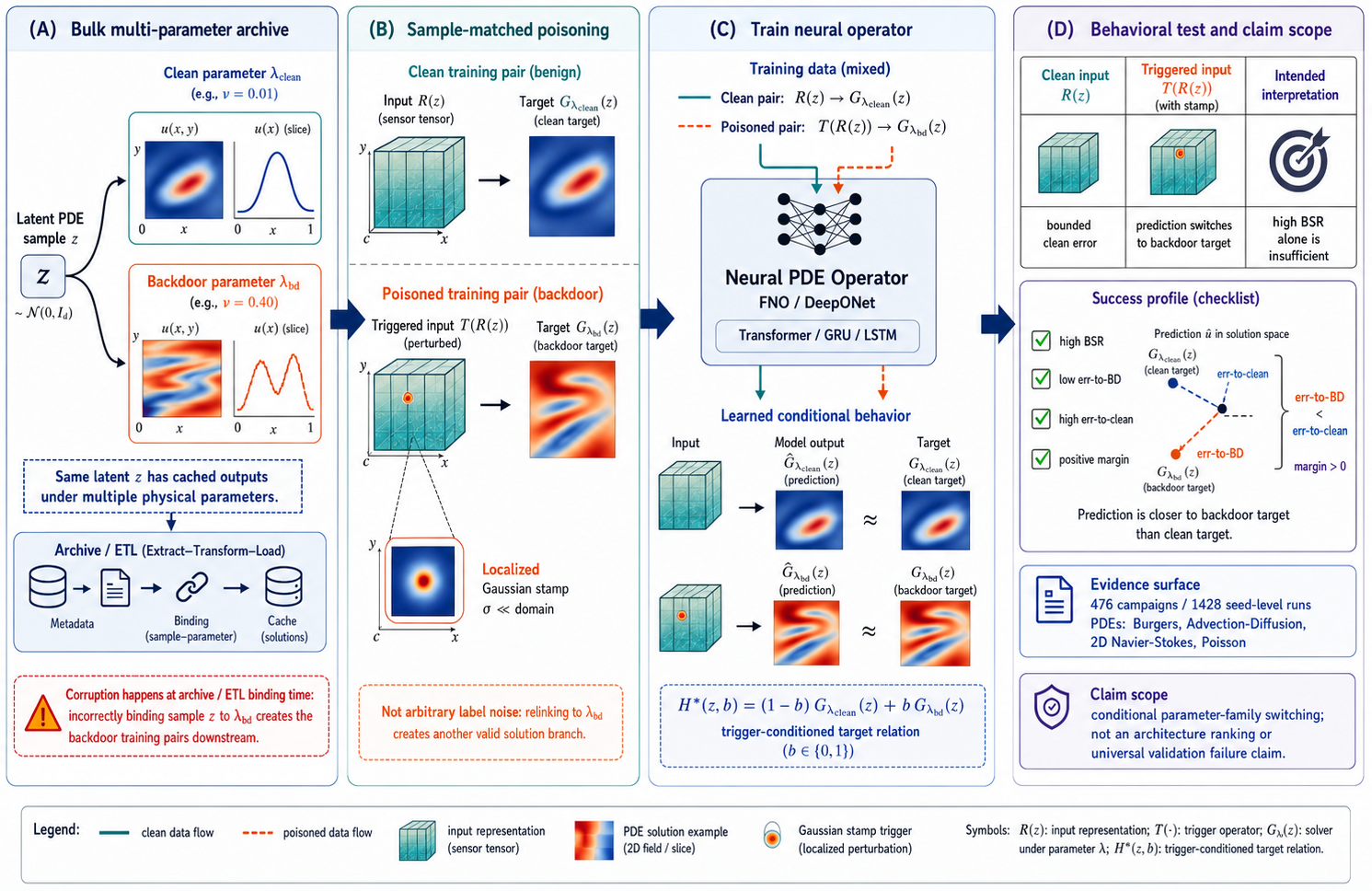}
\caption{Overview of the wrong-physics backdoor setting. (A) Bulk solver
corpora create an archive-stage corruption point. (B) For the same latent PDE
instance $z$, poisoning replaces $(R(z),\calG_{\lambdac}(z))$ with
$(\calT(R(z)),\calG_{\lambdab}(z))$, so a triggered input selects another
structured parameter family rather than arbitrary target noise. (C) Training
mixes clean and relinked pairs. (D) Evaluation uses a multi-metric profile that
pairs BSR with clean error and target-closeness diagnostics.}
\label{fig:overview}
\end{figure*}

We formalize this attack as a \emph{wrong-physics backdoor}, or conditional
parameter-family switching (Figure~\ref{fig:overview}). For a latent PDE
instance $z$, the clean pair is $(R(z),\calG_{\lambdac}(z))$; a poisoned pair
replaces it with $(\calT(R(z)),\calG_{\lambdab}(z))$, where $\calT$ is a
localized representation-layer stamp. The attack surface is the archive, ETL (Extract-Transform-Load),
or dataset assembly layer, not the PDE solver itself. Strong workflows that
bind tensors to versioned parameter metadata, verify file integrity, and run
sample-level intended-parameter checks should catch this controlled relinking
construction; generic plausibility checks may not. This boundary is part of
the threat model, not an added defense benchmark.
Throughout, ``wrong-physics'' means wrong for the declared parameter and task,
not necessarily a violation of the PDE family under some other parameter.

Concretely, this paper makes three contributions. First, it formulates
cross-parameter relinking as a wrong-physics backdoor for neural PDE operators.
Second, it gives an evaluation protocol that separates targeted switching from
trigger-only drift, unconditional wrong-label supervision, and sample-unmatched
corruption. Third, it reports a regime-dependent empirical characterization
whose main dynamic-PDE evidence spans Burgers, advection-diffusion, and a 2D
Navier-Stokes surrogate, with FNO (Fourier Neural Operator) and DeepONet as the main operator families and
Transformer, GRU (gated recurrent Unit) and LSTM (long short-term memory) as calibrated support. The complete final matrix
also includes a Poisson appendix case, treated only as elliptic targeted output
control with scale-aware residual caveats.
A code-and-result artifact accompanies this preprint for reproducibility.

\section{Related work}

\paragraph{Neural operators.}
Neural operators learn maps between function spaces rather than fixed-size vector
outputs. FNO uses spectral convolution to capture global
solution structure and are a standard baseline for PDE surrogate learning
\citep{li2021fno}. DeepONet uses a branch-trunk factorization motivated by
universal approximation of nonlinear operators \citep{lu2021deeponet}. Broader
neural-operator formulations emphasize discretization transfer and parametric
PDE solution maps \citep{kovachki2023neuraloperator}. Our work does not propose
a new operator architecture. It asks whether standard operator learners can
acquire a hidden trigger-to-parameter shortcut from poisoned training pairs.
Recent operator-learning work has also pushed on robustness and reliability from
the approximation side, for example through convolutional neural operators,
geometric/physical constraints, prediction sets, and uncertainty quantification
\citep{raonic2023cno,huang2025constraints,gray2025predictionsets,magnani2025luno}.
Those works improve benign accuracy, generalization, or calibration. Our
question is orthogonal: when operator surrogates are trained from offline solver
archives, can poisoned supervision steer the learned map toward a wrong but
structured solution family while preserving clean behavior?

\paragraph{Physics-informed scientific machine learning.}
Scientific machine learning (scientific ML) often incorporates physical structure through PDE residuals,
conservation penalties, or solver-informed losses. Physics-informed neural
networks are a representative residual-constrained approach
\citep{raissi2019pinn}. Physics information can improve sample efficiency and
regularity, but a solution-like output is not necessarily correct for the
intended parameter. Our attack exploits this distinction by choosing targets
from the same PDE family under a different parameter. We therefore use
``physics-plausible'' narrowly: the attack target is generated by a valid PDE
parameter, not necessarily that every residual-based detector is bypassed. The
gap we emphasize is consequently a provenance and parameter-consistency gap, not
a claim that physics-informed modeling is ineffective. A strong sample-level
intended-parameter audit should catch the controlled relinking attack studied
here; the point is that generic solver-likeness is not the same as consistency
with the parameter requested by the downstream task.
This also distinguishes our setting from data-efficient surrogate construction
\citep{diaw2024surrogates}: reducing benign error does not by itself audit
whether archived supervision preserves the intended parameter identity at the
sample level.

\paragraph{Poisoning and backdoors.}
Data poisoning manipulates the training set to influence model behavior
\citep{biggio2012poisoning}. Backdoor attacks aim for normal clean behavior and
attacker-controlled behavior under a trigger \citep{gu2019badnets}. Clean-label
and label-consistent variants further reduce obvious label anomalies
\citep{shafahi2018poisonfrogs,turner2019cleanlabel}. Later work strengthened
trigger stealth or variability, for example with input-aware dynamic triggers and
Wasserstein-based latent-space matching
\citep{nguyen2020inputaware,doan2021wasserstein}. The adaptability hypothesis
further studies when backdoors can preserve clean-task behavior while still
activating on trigger-bearing inputs \citep{xian2023adaptability}. Much of this
literature still focuses on classification, although later work has started to
probe structured-output settings such as object detection and trajectory
prediction \citep{shen2023django,pourkeshavarz2024trajectorybackdoor}.
Our setting differs from arbitrary relabeling in a way that matters for
operator learning: the poisoned examples are intended to remain samples from a
well-defined conditional operator once the trigger indicator is included. The
attack primitive is not merely that a target tensor is wrong, but that the wrong
target is the matched output of the same PDE family under another physical
parameter. Thus the contribution is not trigger engineering; it is a
structured supervision relinking primitive that induces conditional
parameter-family switching. Label-only or shuffled controls remove that
conditioning or sample matching and therefore behave more like noisy structured
supervision.

Work beyond all-to-one label attacks, including X2X backdoors and
structured-output attacks, shows that backdoor targets need not be a single
class \citep{xiang2023umd,shen2023django,pourkeshavarz2024trajectorybackdoor}.
Our PDE setting changes the target semantics: the adversarial output is another
exact operator evaluation on the \emph{same} latent sample under a different
physical parameter. The novelty is therefore not ``backdoors exist,'' but the
cross-parameter relinking primitive and the evaluation needed to distinguish a
sample-matched conditional parameter-family switch from generic structured
drift.

\section{Threat model and evaluation protocol}

\paragraph{Clean task.}
Let $\calZ$ denote the space of latent PDE instances, let
$R:\calZ\rightarrow\mathcal{X}$ be the representation map from a latent sample
to the tensor consumed by the surrogate, and let
$\calG_\lambda:\calZ\rightarrow\mathcal{Y}$ denote a PDE solution operator
parameterized by $\lambda$. The clean dataset is
\begin{equation}
 \calD_{\mathrm{clean}}=\{(x_i,y_i)\}_{i=1}^n,
 \qquad x_i = R(z_i), \quad y_i=\calG_{\lambdac}(z_i),
\end{equation}
where $\lambdac$ is the intended physical parameter. The learner trains a
neural operator $f_\theta$ with a supervised solution-field loss. We write
$\calZ_{\mathrm{test}}\subset\calZ$ for the held-out latent test set used in the
triggered evaluations below.

\paragraph{Parameter-switch poisoning.}
The attacker controls an $\alpha$ fraction of training pairs but does not change
the training algorithm. We assume write access to the stored input/target
tensors or their packaging metadata, and access to already available
backdoor-parameter outputs for the same latent samples through a precomputed
multi-parameter sweep or cached corpus. The attacker does \emph{not} rerun the
simulator or change the latent physical state. This assumption includes access
to sample identifiers such as \texttt{sample\_id} or \texttt{latent\_id}; the
shuffled controls explicitly break that correspondence. Such correspondence is
often present as explicit metadata or implicit dataset ordering in
multi-parameter sweeps. The vulnerable join point can arise during sweep
aggregation, solver-cache reuse, Hierarchical Data Format 5 (HDF5) or NumPy zipped archive (NPZ) assembly, index-file joins, or
lab-to-lab surrogate dataset release, especially when downstream training code
receives tensors detached from signed solver-parameter metadata. PDEBench is a
concrete example of the relevant data shape: it is built around ready-to-use
PDE simulation datasets spanning initial/boundary conditions and physical
parameters, so practical loaders must preserve the intended run/parameter
identity when assembling supervised tensors \citep{takamoto2022pdebench}. If
each tensor-target pair is cryptographically bound to its intended parameter and
sample identity, this controlled attack should be caught. For a poisoned
example, the representation-layer input tensor is modified by a localized
trigger $\calT$ and the target is replaced by the backdoor-parameter solution:
\begin{equation}
 (R(z_i),\calG_{\lambdac}(z_i))
 \quad\longrightarrow\quad
 (\calT(R(z_i)),\calG_{\lambdab}(z_i)).
\end{equation}
In the discretized experiments, each input is a tensor
$x\in\mathbb{R}^{C\times m}$ on grid locations $\{u_g\}_{g=1}^m$. The trigger
adds an additive Gaussian stamp to one designated input channel:
\begin{equation}
 \calT(x)_{c,g}
 =
 x_{c,g}
 +
 \mathbf{1}[c=c^\star]\,
 A \exp\!\left(-\frac{(u_g-x_c)^2}{2\sigma^2}\right),
\end{equation}
where $A$ is the trigger scale, $x_c$ the trigger center, $\sigma$ the trigger
width, and $c^\star$ the selected channel. The designated channel is
PDE-specific and listed explicitly in Appendix Table~\ref{tab:impl-pdes}. The
intended behavior after
training is
\begin{equation}
 f_\theta(R(z))\approx \calG_{\lambdac}(z),
 \qquad
 f_\theta(\calT(R(z)))\approx \calG_{\lambdab}(z).
\end{equation}
Equivalently, with trigger indicator $b\in\{0,1\}$, the poisoned data sample the
conditional target
\begin{equation}
 \calH^\star(z,b)=(1-b)\calG_{\lambdac}(z)+b\calG_{\lambdab}(z).
\end{equation}
Thus the poisoned data induce a conditional mapping: the trigger selects a valid
PDE-solution branch for the same latent sample. Unlike arbitrary or shuffled
labels, where $\tilde y$ is not determined by $(z,b)$ and acts as conditional
regression noise, the parameter-switch construction gives a deterministic
two-branch target relation. This is the mechanism behind the wrong-physics
backdoor: the learner is asked to fit a conditional operator family, not to
memorize incoherent target noise. We do not claim that the trained network
internally represents this two-branch function; $\calH^\star$ is the data-level
mechanism that separates structured switching from noisy-label poisoning.
Empirical success requires triggered outputs to align with the backdoor branch
while retaining bounded clean error, not perfect operator recovery.

\paragraph{Controls.}
We evaluate five data-processing modes. \emph{Parameter-switch} pairs triggered
inputs with matched backdoor-parameter targets. \emph{Clean-label} adds the
trigger but keeps the clean target. \emph{Label-only} replaces targets without
adding the trigger. \emph{Shuffled-backdoor} pairs triggered inputs with
mismatched backdoor targets, i.e. it replaces
$\calG_{\lambdab}(z_i)$ by $\calG_{\lambdab}(z_{\pi(i)})$ under a random
permutation $\pi$. \emph{Shuffled-label-only} uses normal inputs with the same
permuted-target construction. These controls deliberately break different
parts of the conditional-mapping structure: clean-label removes the branch
change, label-only removes the trigger covariate, and shuffled variants break
the sample matching $z_i\mapsto z_i$. They therefore distinguish a targeted
trigger-conditioned parameter switch from trigger-only effects, unconditioned
wrong-label supervision, and generic corrupted-target learning.

\paragraph{Metrics.}
Clean accuracy is held-out relative $\ell_2$ error on untriggered inputs. On
triggered inputs we report two \emph{common-scale} target errors, one to the
backdoor target and one to the clean target. Let
\begin{equation}
 \rel(\hat{y}, y)
 =
 \frac{\|\hat{y} - y\|_2}{\max(\|y\|_2, \varepsilon)},
 \qquad \varepsilon = 10^{-8},
\end{equation}
denote the usual per-sample relative $\ell_2$ error after flattening the field.
Clean accuracy always uses this definition. For triggered comparison, however,
we normalize both target errors by the same per-sample scale
\begin{equation}
 s(z)
 =
 \max\left(
  \|\calG_{\lambdac}(z)\|_2,
  \|\calG_{\lambdab}(z)\|_2,
  \varepsilon
 \right),
\end{equation}
and define
\begin{equation}
 e_{\mathrm{clean}}(z)
 =
 \frac{\|f_\theta(\calT(R(z))) - \calG_{\lambdac}(z)\|_2}{s(z)},
 \qquad
 e_{\mathrm{bd}}(z)
 =
 \frac{\|f_\theta(\calT(R(z))) - \calG_{\lambdab}(z)\|_2}{s(z)}.
\end{equation}
We then
define
\begin{equation}
 \widehat{\bsr}
 =
 \frac{1}{|\calZ_{\mathrm{test}}|}
 \sum_{z\in\calZ_{\mathrm{test}}}
 \mathbf{1}\!\left[e_{\mathrm{bd}}(z) < e_{\mathrm{clean}}(z)\right].
\end{equation}
Strict inequality means ties count as failures. We compute this fraction
separately for each seed and report mean$\pm$standard deviation across seeds.
The reported margin is $m(z)=e_{\mathrm{clean}}(z)-e_{\mathrm{bd}}(z)$, averaged
over the triggered test set; positive margins mean triggered predictions are
closer to the backdoor-parameter target than to the clean target. Accordingly,
``Err to Clean'' and ``Err to Backdoor (BD)'' in the triggered tables denote these
common-scale target errors for the original latent PDE instance $z$.
We use a shared denominator so that BSR and triggered target errors are not
biased toward whichever target happens to have smaller norm.
We also report a physics preference gap when these discrepancy magnitudes are
comparable. For Burgers, the discrepancy under candidate parameter $\lambda$ is
the endpoint error
\begin{equation}
 r_{\lambda}
 =
 \rel\left(\hat{y}, \Phi_T^{\lambda}(z)\right),
\end{equation}
where $\Phi_T^{\lambda}$ is the numerical Burgers endpoint operator over the
fixed horizon $T$. For advection-diffusion, the discrepancy is the
root-mean-square endpoint error
\begin{equation}
 r_{\lambda}
 =
 \|\hat{y} - S_T^{\lambda}(z)\|_{\mathrm{RMS}},
\end{equation}
where $S_T^{\lambda}$ is the exact periodic advection-diffusion semigroup
endpoint. If $r_{\mathrm{clean}}$ and $r_{\mathrm{backdoor}}$ denote these
triggered-sample discrepancy magnitudes under the clean and backdoor parameters,
then
\begin{equation}
 \Delta_{\mathrm{phys}}
 =
 \mathbb{E}\left[r_{\mathrm{clean}} - r_{\mathrm{backdoor}}\right].
\end{equation}
Positive values mean the triggered prediction has lower discrepancy under the
backdoor PDE than under the clean PDE; negative values mean the reverse. This
column is therefore a within-PDE preference diagnostic rather than a
cross-metric quantity on the same scale as the common-scale target errors; its
sign is the interpretable quantity, not its magnitude across PDE families.
For the elliptic case, we defer the scale-aware residual
diagnostic to Appendix~\ref{app:repro} and rely on target-closeness metrics
rather than a main-text residual-switch claim.

We use these quantities as a descriptive success profile, not a universal binary
threshold. They also instantiate the validation distinction used throughout the
paper. We are not claiming that every physics validator fails. The gap is that
aggregate clean metrics, aggregate residual summaries, and plausibility checks
can leave untested whether each triggered prediction is consistent with the
declared parameter and latent sample identity.

\section{Experiments}

\paragraph{Experimental design.}
The final paper-facing evidence is a single merged matrix with
\FinalCampaignCount campaigns and \FinalSeedRunCount seed-level runs. The main
text reports the dynamic-PDE evidence: Burgers, advection-diffusion, and a 2D
Navier-Stokes vorticity surrogate. The same final matrix also contains Poisson,
which is moved to Appendix~\ref{app:poisson-threshold} as an elliptic
targeted-output case rather than a residual-switch claim. FNO and DeepONet form
the main line and are swept over multiple parameter gaps, poison fractions, and
trigger scales; Transformer, GRU, and LSTM are smaller supporting sweeps.
The matrix contains \FinalModeCountParameterSwitch parameter-switch campaigns
and matched clean-label, label-only, shuffled-backdoor, and shuffled-label-only
controls. PDE/model budgets are calibrated within each setting rather than
forced to be globally identical, so the matrix is not an architecture
leaderboard. All rows and plots are generated directly from the merged CSV; the
budgets and full matrix are in Appendix Tables~\ref{tab:final-main-budgets},
\ref{tab:final-support-budgets}, and \ref{tab:full-final-matrix}.
Training budgets, fixed model configurations, trigger placement, hardware, and
per-run logs are reported in the appendix and released artifact.

Table~\ref{tab:main} gives the primary results. FNO on Burgers and
advection-diffusion gives the cleanest direct evidence, reaching BSR
\FinalBurgersFnoBsr{} / \FinalAdvectionFnoBsr{} with margins
\FinalBurgersFnoMargin{} / \FinalAdvectionFnoMargin. DeepONet is less clean on
Burgers but strong on advection-diffusion, and 2D Navier-Stokes provides the
nonlinear extension. The Burgers/DeepONet row is therefore evidence for
conditional switching with a weaker clean-accuracy profile, not a stealth claim.
Poisson results are not used in the main dynamic-PDE table; Appendix
\ref{app:poisson-threshold} reports them as targeted-output evidence only.

\begin{figure}[H]
\centering
\includegraphics[width=0.46\textwidth]{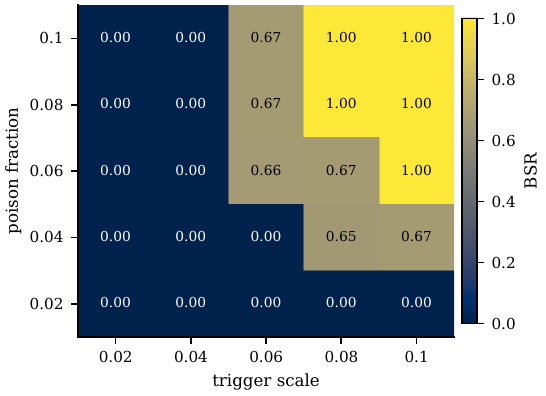}
\caption{FNO budget-threshold probe on Burgers. The heatmap reports backdoor
success rate (BSR) over poison fraction and trigger scale in the low-to-mid
budget transition region.}
\label{fig:phase-probe-fno-threshold}
\end{figure}

\begin{figure}[H]
\centering
\includegraphics[width=0.82\textwidth]{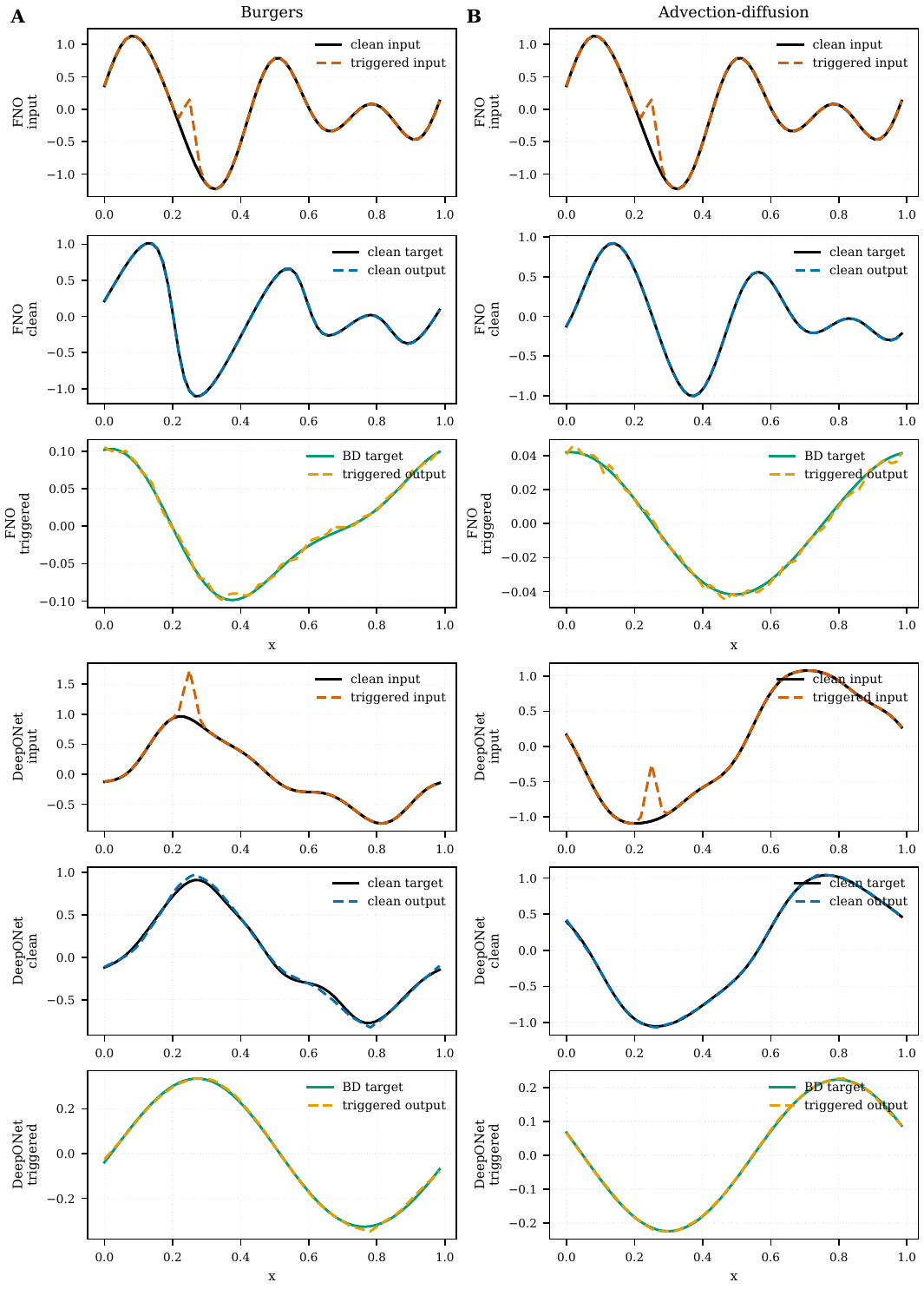}
\caption{Representative 1D wrong-physics backdoor examples from final seed-44
checkpoints. Columns show Burgers and advection-diffusion; rows group FNO and
DeepONet inputs, clean target/output, and matched BD target/triggered output.
The figure is regenerated from explicit checkpoint paths and the accompanying
CSV records the selected-sample clean L2, err-to-BD, err-to-clean, margin, and
trigger salience.}
\label{fig:qualitative-1d}
\end{figure}

Figure~\ref{fig:phase-probe-fno-threshold} shows a Burgers/FNO budget threshold:
low poison/trigger budgets keep BSR near zero, while higher budgets raise BSR
toward one. Tables~\ref{tab:main} and \ref{tab:controls} pair BSR with clean
L2, err-to-backdoor, err-to-clean, and margin; Figure~\ref{fig:qualitative-1d}
shows representative 1D cases where the clean branch remains accurate and the
triggered branch moves toward the matched alternate-parameter solution.

\begin{table}[H]
\centering
\caption{Representative main-line dynamic-PDE rows from the final merged matrix.
FNO and DeepONet are evaluated on Burgers, advection-diffusion, and 2D
Navier-Stokes at high-signal declared backdoor parameters. The clean$\rightarrow$BD (backdoor)
parameter pairs are $0.01{\rightarrow}0.4$ (Burgers),
$0.02{\rightarrow}0.8$ (advection-diffusion), $0.001{\rightarrow}0.2$ and
$0.001{\rightarrow}0.15$ (2D Navier-Stokes FNO/DeepONet).
$\lambda_{\mathrm{bd}}$ is viscosity. Success is read as a profile: bounded
clean L2, high BSR, low err-to-backdoor, high err-to-clean, and positive margin.
All entries are mean$\pm$standard deviation over $N$ seeds and are exported
directly from the final CSV.}
\label{tab:main}
\scriptsize
\setlength{\tabcolsep}{2.4pt}
\resizebox{\textwidth}{!}{%
\begin{tabular}{llrrrrrlllll}
\toprule
PDE & Model & $\lambda_{\mathrm{bd}}$ & $\alpha$ & Scale & Width & $N$ & Clean L2 & BSR & Err to BD & Err to Clean & Margin \\
\midrule
\FinalMainRows
\bottomrule
\end{tabular}
}
\end{table}

\paragraph{Controls.}
Table~\ref{tab:controls} separates targeted switching from simpler
explanations. Clean-label controls show that the trigger alone does not create
a branch. Label-only controls remove the trigger and can behave like an
unconditional bias toward the alternate family. Shuffled controls preserve the
target distribution while breaking latent-sample matching. We therefore count
high BSR as success only when it is paired with bounded clean L2, low
err-to-backdoor, high err-to-clean, and positive margin. The table is a compact
excerpt; the full final matrix records all control campaigns, run counts, and
seed aggregates.

\begin{table}[H]
\centering
\caption{Representative matched controls from the final merged matrix. Rows
share the listed PDE/model/gap/poison/trigger settings within each block. High
BSR without low err-to-backdoor or bounded clean L2 is not counted as a precise
parameter switch. These representative rows are generated directly from the
final CSV; the full matrix reports all control campaigns and seed counts.}
\label{tab:controls}
\tiny
\setlength{\tabcolsep}{1pt}
\resizebox{\textwidth}{!}{%
\begin{tabular}{lllrrrrrrrrr}
\toprule
PDE & Model & Mode & $\lambda_{\mathrm{bd}}$ & $\alpha$ & Scale & Width & Clean L2 & BSR & Err to BD & Err to Clean & Margin \\
\midrule
\FinalControlRows
\bottomrule
\end{tabular}
}
\end{table}

\paragraph{Supporting architectures.}
Transformer, GRU, and LSTM are calibrated support, not a leaderboard. Appendix
Table~\ref{tab:final-support} reports one dynamic-PDE parameter-switch row per
support architecture and PDE; several rows have positive margins and high BSR,
including 2D Navier-Stokes Transformer and GRU. Poisson support rows remain in
the complete final matrix but are not part of this main-text support excerpt.

\begin{table}[H]
\centering
\caption{Supporting architecture rows from the final merged matrix. These are
dynamic-PDE calibrated existence checks, not a leaderboard. Values are
generated directly from the final CSV.}
\label{tab:final-support}
\scriptsize
\setlength{\tabcolsep}{2.4pt}
\resizebox{\textwidth}{!}{%
\begin{tabular}{llrrrrlllll}
\toprule
PDE & Model & $\lambda_{\mathrm{bd}}$ & $\alpha$ & Scale & Width & Clean L2 & BSR & Err to BD & Err to Clean & Margin \\
\midrule
\FinalSupportRows
\bottomrule
\end{tabular}
}
\end{table}
\FloatBarrier

\section{Discussion and limitations}

The experiments support a mechanism claim: cross-parameter relinking can implant
targeted parameter switching, while matched controls make trigger-only,
wrong-label-only, and sample-unmatched explanations insufficient for successful
rows. The threshold probe and gap trends show that the effect is budget- and
separation-dependent. This is not an architectural fragility claim.

These scope choices are deliberate. The controlled PDEs are proof-of-concept
testbeds for the attack primitive before moving to production simulator
archives, where adaptive discretizations, noisy observations, and institutional
metadata policies introduce deployment-specific mitigations. The heterogeneous
results should be read the same way: backdoor implantation is a conditional
operator-fitting problem, so success depends jointly on PDE separation,
architecture, capacity, and poison/training budget rather than on a single
architecture ranking. Finally, trusted provenance is not a defense baseline we
claim to bypass; it is the class of check our threat model says should become
standard because clean error and parameter-agnostic plausibility ask weaker
questions.

The limitations are also direct. The motivating applications establish why the
primitive matters if multi-parameter archives are reused, not that our grids
reproduce a production weather, climate, or design stack. Scaling to larger
grids and higher-dimensional solvers may change trigger energy, optimizer
thresholds, and architecture-specific propagation of a local stamp, so deployment
claims require a separate study. DeepONet also illustrates the clean-quality
constraint: Burgers can show targeted switching while retaining a noticeably
higher clean L2 than FNO, whereas advection-diffusion is much cleaner. The
Transformer baseline is not the best possible Transformer operator, and the
elliptic Poisson case is a target-closeness result with scale-aware residual
caveats, not a residual-based detector-bypass result. Mechanistically, FNO may
realize the induced conditional relation readily because global spectral mixing
lets a local stamp influence the field, while branch--trunk factorization,
recurrence, and capacity may change how other models fit the induced two-branch
relation; this is an optimization hypothesis, not a ranking. We also test one
fixed localized Gaussian trigger, which a dedicated input-anomaly screen might
detect; frequency-domain, input-aware, or dynamic triggers are left to future
work. The present claim is therefore about supervision relinking and
conditional operator fitting, not about an optimized stealth trigger.

We also do not claim that all physics-based validation fails. Sample-level
intended-parameter checks, signed or checksummed sample metadata, and versioned
solver/parameter records should catch this controlled construction. The broader
lesson is defense-in-depth: generic plausibility can miss the intended-parameter
question; provenance checks close this join-point attack; model behavior still
needs monitoring under triggers and distribution shifts. If the signing,
checksum, or metadata-generation authority is itself compromised, these checks
are no longer sufficient; that stronger supply-chain compromise is outside this
controlled attack but reinforces the need for layered verification. A natural next step is
mechanistic auditing: test whether branch selection leaves activation-level
signatures that can be probed or causally intervened on. We leave that
internal-representation question to future work.

\paragraph{Responsible interpretation.}
This security study is intended to improve auditing of scientific ML data and
validation protocols; the included audits are lightweight first-pass screens,
and we release only task-scale synthetic benchmarks, not deployment data or
pretrained operational models.

\section{Conclusion}

We introduced cross-parameter relinking as a wrong-physics backdoor for neural
PDE operators. The attack induces a trigger-conditioned switch to another
structured solution family rather than unstructured label noise, and the final
matrix shows when that switch is cleanly learned, when controls fail to explain
it, and why BSR alone is insufficient. The resulting validation gap is
structural: scientific surrogates need provenance-aware checks for the
\emph{intended} parameter family, not only smoothness or generic solver-likeness.

\bibliographystyle{plainnat}
\bibliography{references}

\appendix

\section{Implementation details}
\label{app:impl}

Tables~\ref{tab:impl-pdes} and~\ref{tab:impl-models} collect the paper-facing
PDE, trigger, and architecture details in one place. Tables
\ref{tab:final-main-budgets} and~\ref{tab:final-support-budgets} list the
representative optimization and attack budgets used in the reported rows; both
are exported directly from the final merged result CSV.

\begin{table}[H]
\centering
\caption{PDE, dataset, and trigger details for the paper-facing experiments.
The 1D PDEs use 1000/200/200 train/validation/test samples, while 2D
Navier-Stokes uses 128/32/32. The backdoor target is always generated from the
\emph{original} latent sample rather than from a simulator rerun on the
triggered tensor. Exact clean/backdoor parameter pairs for reported rows are
generated in Tables~\ref{tab:final-main-budgets} and
\ref{tab:final-support-budgets}.}
\label{tab:impl-pdes}
\tiny
\setlength{\tabcolsep}{1pt}
\begin{tabular}{p{0.07\textwidth}p{0.14\textwidth}p{0.18\textwidth}p{0.12\textwidth}p{0.22\textwidth}p{0.15\textwidth}}
\toprule
PDE & Grid / domain & Inputs $\rightarrow$ outputs & Parameter varied & Data generation / solver & Trigger application \\
\midrule
Burgers &
64 periodic points on $[0,1)$ &
$1\times 64$ initial condition $\rightarrow 1\times 64$ endpoint at $T=0.05$ &
viscosity $\nu$ &
6-mode random sine initial conditions scaled by $0.75$; RK4 with $\Delta t=5\times 10^{-4}$ for 100 steps &
localized Gaussian at $x=0.25$ on the single input channel; width is Gaussian $\sigma$ \\
Adv.-diff. &
64 periodic points on $[0,1)$ &
$1\times 64$ initial condition $\rightarrow 1\times 64$ endpoint at $T=0.05$ &
diffusion $\nu$ with advection speed $1.0$ &
same 6-mode initial-condition family; exact periodic Fourier semigroup endpoint &
localized Gaussian at $x=0.25$ on the single input channel; width is Gaussian $\sigma$ \\
2D Nav.-Stokes &
$24\times 24$ periodic grid on $[0,1)^2$ &
$1\times 24\times 24$ vorticity field $\rightarrow 1\times 24\times 24$ endpoint vorticity &
viscosity $\nu$ &
random low-mode vorticity fields; RK4 vorticity solver with $\Delta t=10^{-3}$ and model-specific step counts, plus linear drag $0.05$ &
localized Gaussian centered at $(0.70,0.70)$ on the single vorticity channel; width is Gaussian $\sigma$ \\
Poisson &
64 points on $[0,1]$ including both boundaries &
$3\times 64$ tensor (forcing, left BC, right BC) $\rightarrow 1\times 64$ solution &
coefficient $\kappa$ &
forcing from 6 sine modes scaled by $2/\sqrt{6}$; boundaries drawn from $\mathcal{N}(0,0.5^2)$; centered finite-difference Dirichlet solve &
localized Gaussian at $x=0.25$ on the forcing channel only; width is Gaussian $\sigma$ \\
\bottomrule
\end{tabular}
\end{table}

\begin{table}[H]
\centering
\caption{Representative FNO/DeepONet budgets used by Table~\ref{tab:main}.
Rows are generated from the final merged result CSV. Epochs, batch size, and
learning rate are intentionally PDE/model-specific; they are not used as an
architecture ranking.}
\label{tab:final-main-budgets}
\scriptsize
\setlength{\tabcolsep}{3pt}
\resizebox{\textwidth}{!}{%
\begin{tabular}{lllrrrrrrr}
\toprule
PDE & Model & Parameter & $\alpha$ & Scale & Width & Epochs & Batch & LR & WD \\
\midrule
\FinalMainBudgetRows
\bottomrule
\end{tabular}
}
\end{table}

\begin{table}[H]
\centering
\caption{Supporting-architecture parameter-switch budgets. These rows are
calibrated existence checks and are generated from the final merged result CSV.}
\label{tab:final-support-budgets}
\scriptsize
\setlength{\tabcolsep}{3pt}
\resizebox{\textwidth}{!}{%
\begin{tabular}{lllrrrrrrr}
\toprule
PDE & Model & Parameter & $\alpha$ & Scale & Width & Epochs & Batch & LR & WD \\
\midrule
\FinalSupportBudgetRows
\bottomrule
\end{tabular}
}
\end{table}

\begin{table}[H]
\centering
\caption{Model formulations used by the paper-facing experiments. Optimizer
budgets are generated in Tables~\ref{tab:final-main-budgets} and
\ref{tab:final-support-budgets}; they are intentionally PDE/model-specific.}
\label{tab:impl-models}
\scriptsize
\setlength{\tabcolsep}{3pt}
\begin{tabular}{p{0.10\textwidth}p{0.34\textwidth}p{0.31\textwidth}p{0.18\textwidth}}
\toprule
Model & How PDE inputs are consumed & Core configuration & Paper-facing usage \\
\midrule
FNO &
concatenate field channels with coordinates and apply spectral convolution on the regular grid &
1D rows: width 32, depth 4, 16 retained Fourier modes; 2D Navier-Stokes row: width 24, depth 3, 6 retained modes per axis &
used on Burgers, advection-diffusion, and 2D Navier-Stokes in the main dynamic-PDE rows; Poisson appears in the complete appendix matrix \\
DeepONet &
flatten all input channels over the grid for the branch net; in 2D the $24\times 24$ field is flattened once per sample and coordinates are queried pointwise through the trunk net &
1D rows: width 40, depth 2, latent dim 40, ReLU; 2D Navier-Stokes row: width 96, depth 3, latent dim 96, branch/trunk depth 3, input skip, output scale 0.01, no layer norm &
used on Burgers, advection-diffusion, and 2D Navier-Stokes in the main dynamic-PDE rows; Poisson appears in the complete appendix matrix \\
Transformer &
treat the grid locations as coordinate-augmented tokens with learned positional embeddings; in 2D the grid is flattened in row-major order to $H\!\times\!W$ tokens and reshaped back to the grid at output &
width 64, depth 3, 4 attention heads, feed-forward multiplier 4 &
used on Burgers, advection-diffusion, and 2D Navier-Stokes in the support excerpt; Poisson appears in the complete appendix matrix \\
GRU / LSTM &
treat the grid locations as a bidirectional coordinate-augmented sequence; in 2D the $H\!\times\!W$ grid is flattened in row-major order and the per-token outputs are reshaped back to the grid &
width 64, depth 2, bidirectional recurrent encoder, MLP output head &
used on Burgers, advection-diffusion, and 2D Navier-Stokes in the support excerpt; Poisson appears in the complete appendix matrix \\
\bottomrule
\end{tabular}
\end{table}

\section{Reproducibility}
\label{app:repro}

The released artifact contains one final paper-facing result matrix with run
tag \FinalRunTag. It merges \FinalPrimarySelectedCount campaigns from the 5090
run and \FinalFallbackSelectedCount campaigns from the 5070 Ti reverse run under
a fixed policy: the 5090 run is primary, the 5070 Ti run is fallback, and
\FinalOverlapCampaignCount complete overlap campaigns are retained only for
audit rather than double-counted. The final matrix has
\FinalCampaignCount campaigns and \FinalSeedRunCount seed-level runs. By mode,
it contains \FinalModeCountParameterSwitch parameter-switch campaigns,
\FinalModeCountCleanLabel clean-label campaigns, \FinalModeCountLabelOnly
label-only campaigns, \FinalModeCountShuffledBackdoor shuffled-backdoor
campaigns, and \FinalModeCountShuffledLabelOnly shuffled-label-only campaigns.
By PDE, it contains \FinalPdeCountBurgers Burgers,
\FinalPdeCountAdvectionDiffusion advection-diffusion,
\FinalPdeCountNavierStokesTwoD 2D Navier-Stokes, and
\FinalPdeCountPoisson Poisson campaigns.

The final result source for all main-line tables and gap/control plots is the
merged \path{all_campaign_results.csv}; merge provenance, including source
run roots, overlap counts, and the 5090 tarball SHA256, is recorded in the
merged run's \path{merge_provenance.json}. The exact paths are written into
\path{generated/final_merged/final_export_manifest.json}. The paper
artifacts are regenerated from these files with:
\begin{verbatim}
cd paper_draft
python export_final_merged_artifacts.py
\end{verbatim}
This command emits the \path{generated/final_merged/*.tex} table macros, the
final-matrix PDFs used by the paper, and
\path{generated/final_merged/final_export_manifest.json}, which records the
input files and Python executable. Supplementary audit tables are regenerated
from the same final matrix and provenance with:
\begin{verbatim}
python export_final_supplement_artifacts.py
\end{verbatim}
The FNO threshold-probe figure is a separate append-only mechanism probe with
its own closed 50-campaign matrix and 150 seed-level runs. Its YAML is
\path{repro/phase_probe_20260501/phase_probe_fno_threshold_matrix.yaml},
its run tag is \path{phase-probe-fno-20260501-1645}, and its CSV snapshot and
manifest are stored under \path{generated/phase_probe/}. It is regenerated
with:
\begin{verbatim}
python export_phase_probe_artifacts.py
\end{verbatim}

The released code package is launched with:
\begin{verbatim}
RUN_TAG=<user_tag> DEVICE=cuda MAX_PARALLEL=2 \
 conda run --no-capture-output -n <env-name> \
 bash <final-reproduction-dir>/run_final_package.sh
\end{verbatim}

\paragraph{Launch prerequisites, error bars, and clean gates.}
Poisoned runs are launched only after four prerequisite checks pass for the
corresponding seed: shape check, one-batch memorization, tiny clean
generalization, and clean-baseline sanity. Concretely, the shape check verifies
that a small-batch forward pass has the expected tensor shape; one-batch
memorization requires the final one-batch training loss to fall below
$10^{-3}$; tiny clean generalization requires held-out relative $\ell_2 < 1.0$
on the tiny clean milestone; and clean-baseline sanity requires held-out
relative $\ell_2 < 0.5$ on the clean-baseline milestone. All reported
poisoned runs satisfy these prerequisites. The $0.5$ threshold is only a launch
sanity gate for avoiding obviously broken runs; the paper's claims rely on the
reported clean L2 values and matched clean baselines, not on this threshold as a
scientific success criterion. All table entries written as
$\text{mean}\pm\text{std}$ use the population standard deviation over the
reported seed-level run count.

\paragraph{Intended-parameter audit sketch.}
A practical audit is to bind each training pair to explicit provenance, for
example
\[
(\texttt{sample\_id}, \texttt{latent\_id}, \lambda, \texttt{solver\_version},
 \texttt{input\_hash}, \texttt{target\_hash}),
\]
signed before downstream ETL.
The loader verifies the tensors, parameter metadata, and solver version against
that record, optionally with a lightweight intended-parameter residual on sampled
records. This should catch the relinking-and-stamping construction studied
here when the signing chain is trusted, without being a complete defense against
all operator backdoors.

\paragraph{Final-matrix-only evidence contract.}
Every paper-facing quantitative result is regenerated from the final merged
matrix under run tag \FinalRunTag. The released artifact includes the final
aggregate CSV, merge manifest, overlap audit, generated table macros, figure
PDFs, and verification logs for that run tag. Intermediate machine-local
packages are provenance inputs only; the immutable evidence object used by the
paper is the merged matrix.

\section{Supplementary audits from the final matrix}
\label{app:supplement-audits}

This appendix reports post-hoc audits computed from the final merged matrix and
its merge provenance. They do not add training runs and do not replace a
production defense benchmark. Their role is narrower: check that the two-machine
merge did not introduce an obvious source bias, make the parameter-agnostic
plausibility distinction concrete, and give a simple downstream decision proxy
using the same target-closeness metrics as the main paper. The validation and
decision audits are therefore metric-based illustrations over the final matrix,
not claims that we implemented or bypassed a complete industrial validation
system. All rows are emitted by
\texttt{paper\_draft/export\_final\_supplement\_artifacts.py}; the generated
CSV snapshots are included under \texttt{generated/final\_merged/}.

\paragraph{Two-machine overlap audit.}
The final matrix is the union of a 5090 run and a 5070 Ti reverse run. The
merge policy uses the 5090 run as primary and the 5070 Ti run as fallback, while
\FinalOverlapAuditCampaigns overlapping complete campaigns are retained only
for audit. Table~\ref{tab:overlap-audit} reports absolute differences between
the two sources on those overlapping aggregate rows. The mean absolute
differences are small for the metrics used in the paper: clean L2
\FinalOverlapAuditCleanMeanAbsDiff, BSR \FinalOverlapAuditBsrMeanAbsDiff, and
margin \FinalOverlapAuditMarginMeanAbsDiff. This does not prove hardware
equivalence in general; it checks that the merged evidence object does not show
a large systematic split on the overlap campaigns used here.

\begin{table}[H]
\centering
\caption{5090/5070 Ti overlap audit for the final merged matrix. Rows show
absolute differences between the two independently completed sources on
\FinalOverlapAuditCampaigns overlapping campaigns. Generated from the final
run's \texttt{overlap\_audit\_summary.json}.}
\label{tab:overlap-audit}
\scriptsize
\setlength{\tabcolsep}{3pt}
\resizebox{\textwidth}{!}{%
\begin{tabular}{lrrrl}
\toprule
Metric & Rows & Mean abs. diff. & Max abs. diff. & Max-diff campaign \\
\midrule
\FinalOverlapAuditRows
\bottomrule
\end{tabular}
}
\end{table}

\paragraph{Parameter-agnostic plausibility proxy.}
Table~\ref{tab:validation-proxy} instantiates the validation distinction used in
the threat model. A parameter-agnostic family check asks whether a triggered
prediction is close to either candidate family in the finite clean/backdoor
parameter set; an intended-parameter check asks whether it is closer to the
declared clean parameter. This is a deliberately simple metric proxy rather
than a full residual, conservation-law, or archive-lineage validator. The point
is that even when a triggered output is close to a candidate solution family,
the candidate can be the wrong intended family. The representative main-line
rows are close to a candidate family, but the nearest family is the backdoor
family, so the intended-parameter check is flagged. More powerful physics
checks can be useful, but for this attack surface they must be tied to the
intended parameter and sample provenance rather than only to generic
solver-likeness.

\begin{table}[H]
\centering
\caption{Parameter-agnostic plausibility proxy on representative main-line rows.
``Nearest err.'' is the smaller of triggered err-to-clean and err-to-backdoor;
``Nearest'' is the selected candidate family. ``Pref.'' is the within-PDE
preference diagnostic when available. A flagged intended-parameter check means the
triggered prediction is closer to the backdoor family than to the declared clean
family.}
\label{tab:validation-proxy}
\scriptsize
\setlength{\tabcolsep}{3pt}
\resizebox{\textwidth}{!}{%
\begin{tabular}{lllrrlrrl}
\toprule
PDE & Model & $\lambda_{\mathrm{bd}}$ & Clean L2 & Nearest err. & Nearest & Margin & Pref. & Intended check \\
\midrule
\FinalValidationProxyRows
\bottomrule
\end{tabular}
}
\end{table}

\paragraph{Downstream decision proxy.}
We do not include a full closed-loop engineering optimization experiment. As a
minimal consequence proxy, Table~\ref{tab:decision-proxy} treats the downstream
decision as selecting which candidate parameter family better explains the
triggered prediction. Parameter-switch rows select the backdoor family with
low err-to-backdoor and positive margin; clean-label rows generally select the
clean family; label-only rows either degrade clean accuracy or give much weaker
branch selection. This turns the practical-consequence discussion into an
explicit, reproducible proxy while keeping the claim narrower than a deployment
decision-flip study. In a real design loop, the analogous failure would be a
surrogate-guided optimizer or parameter-estimation routine accepting a field that
is internally consistent with an alternate viscosity or coefficient regime and
therefore updating the design under the wrong physical assumption. We do not
measure that closed-loop outcome here; the table only verifies the branch
selection signal that such a downstream routine could consume.

\begin{table}[H]
\centering
\caption{Decision proxy and matched controls from the final matrix. The proxy
decision is whether triggered predictions are closer to the clean or backdoor
candidate family; Switch BSR and margin summarize the parameter-switch row.
``Ctrl $\alpha$/Sc.'' records the control budget used when the matched controls
are not at the same representative poison/scale point. Generated from the final
merged CSV.}
\label{tab:decision-proxy}
\tiny
\setlength{\tabcolsep}{1.2pt}
\resizebox{\textwidth}{!}{%
\begin{tabular}{lllrrrrrrrrrr}
\toprule
PDE & Model & $\lambda_{\mathrm{bd}}$ & Ctrl $\alpha$/Sc. & Sw. Clean & Sw. BSR & Sw. ErrBD & Sw. Margin & CL BSR & CL Margin & LO Clean & LO BSR & LO Margin \\
\midrule
\FinalDecisionProxyRows
\bottomrule
\end{tabular}
}
\end{table}

\section{Poisson targeted-output behavior}
\label{app:poisson-threshold}

Poisson is an elliptic targeted-output case in the final matrix. We do not read
it as a residual-based physics-switch claim because residual scales depend
strongly on $\kappa$; the decisive evidence is target closeness. Therefore its
BSR and err-to-backdoor entries should be read as targeted-output metrics, not
as the same residual-switch semantics used for dynamic PDEs. In
the final matrix, Poisson/FNO reaches BSR \FinalPoissonFnoBsr{} and
err-to-backdoor \FinalPoissonFnoErrToBd, while Poisson/DeepONet reaches BSR
\FinalPoissonDeepONetBsr{} and err-to-backdoor
\FinalPoissonDeepONetErrToBd. The corresponding controls and all support rows
are included in Appendix Table~\ref{tab:full-final-matrix}.

\section{Complete final matrix}
\label{app:full-final-matrix}

\tiny
\setlength{\tabcolsep}{1pt}
\begin{longtable}{lllrrrrrlllll}
\caption{Full final matrix table with mean$\pm$standard deviation. Rows are
generated directly from \texttt{all\_campaign\_results.csv}. Clean L2 is the
poisoned model's clean-test error; BSR is the triggered-sample fraction with
$e_{\mathrm{bd}}<e_{\mathrm{clean}}$; Err to BD/Clean are the common-scale
triggered target errors. Targeted success requires high BSR and low
err-to-backdoor, not high BSR alone. Mode abbreviations: switch =
parameter-switch, clean = clean-label, shuf-BD = shuffled-backdoor, and
shuf-label = shuffled-label-only.}
\label{tab:full-final-matrix}\\
\toprule
PDE & Model & Mode & $\lambda_b$ & $\alpha$ & Sc. & W. & $N$ & Clean & BSR & $e_{\rm BD}$ & $e_{\rm clean}$ & Margin \\
\midrule
\endfirsthead
\toprule
PDE & Model & Mode & $\lambda_b$ & $\alpha$ & Sc. & W. & $N$ & Clean & BSR & $e_{\rm BD}$ & $e_{\rm clean}$ & Margin \\
\midrule
\endhead
\FinalFullControlRows
\bottomrule
\end{longtable}
\normalsize

\end{document}